\documentclass{article} % For LaTeX2e
\usepackage{iclr2022_conference,times}

\usepackage{amsmath,amsfonts,bm}

\def\eqref#1{equation~\ref{#1}}
\def\1{\bm{1}}

\DeclareMathAlphabet{\mathsfit}{\encodingdefault}{\sfdefault}{m}{sl}
\SetMathAlphabet{\mathsfit}{bold}{\encodingdefault}{\sfdefault}{bx}{n}

\newcommand{\pdata}{p_{\rm{data}}}
\newcommand{\pmodel}{p_{\rm{model}}}

\DeclareMathOperator*{\argmax}{arg\,max}
\DeclareMathOperator*{\argmin}{arg\,min}

\usepackage[colorlinks=true, linkcolor=black, citecolor=gray]{hyperref}  % edited hyperref import to make citations grey
\usepackage{url}

\usepackage{booktabs}       % professional-quality tables
\usepackage{amsfonts,amsmath,amssymb,amsthm}
\usepackage{todonotes}
\usepackage{subcaption}
\usepackage{mathtools}
\usepackage{ifthen}  % for variable-length appendix
\usepackage{wrapfig}
\usepackage{multirow}
\usepackage{makecell}
\newtheorem{theorem}{Theorem}[section] % numbered theorems
\newcommand{\optimal}[1]{\ensuremath{{#1}^*}}
\usepackage{enumitem}% http://ctan.org/pkg/enumitem
\setlist[itemize]{noitemsep, topsep=0pt}
\usepackage[capitalise]{cleveref}

\makeatletter
\newcommand*{\rom}[1]{\expandafter\@slowromancap\romannumeral #1@}
\makeatother

\title{Conditional Image Generation by \\ Conditioning Variational Auto-Encoders}
\author{%
  William Harvey, Saeid Naderiparizi \& Frank Wood\thanks{Frank Wood is also affiliated with the Montréal Institute for Learning Algorithms (Mila) and Inverted AI.}
  \\
  Department of Computer Science\\
  University of British Columbia\\
  Vancouver, Canada \\
  \texttt{\{wsgh,saeidnp,fwood\}@cs.ubc.ca} \\
}

\iclrfinalcopy % Uncomment for camera-ready version, but NOT for submission.
\begin{document}

\maketitle

\newcommand{\I}{\mathbf{x}}
\newcommand{\z}{\mathbf{z}}
\newcommand{\tildeI}{\mathbf{\tilde{x}}}
\newcommand{\partq}{c}
\newcommand{\partphi}{\psi}
\newcommand{\Partphi}{\Psi}
\newcommand{\partI}{\mathbf{y}}
\newcommand{\unp}{\optimal{p}}
\newcommand{\pcomp}{p_{\text{cond}}}
\newcommand{\EX}{\mathbb{E}}
\newcommand{\mutinf}{I^*_{\tildeI{},\partI{}|\z{}}}
\newcommand{\mutinfsub}{REPLACE} % {I^*_{(\I{}\setminus\partI{}),\partI{}|z}}
\newcommand{\coord}{l}

\DeclarePairedDelimiterX{\infdivx}[2]{(}{)}{%
  #1\;\delimsize\|\;#2%
}
\newcommand{\kl}{\text{KL}\infdivx}
\newcommand{\changed}[2]{\textcolor{red}{[Old: #1]} \textcolor{green}{[New: #2]}}

\begin{abstract}
  We present a conditional variational auto-encoder (VAE) which, to avoid the
  substantial cost of training from scratch, uses an architecture and training
  objective capable of leveraging a foundation model in the form of a pretrained
  unconditional VAE. To train the conditional VAE, we only need to train an
  artifact to perform amortized inference over the unconditional VAE's latent
  variables given a conditioning input. We demonstrate our approach on tasks
  including image inpainting, for which it outperforms state-of-the-art
  GAN-based approaches at faithfully representing the inherent uncertainty. We
  conclude by describing a possible application of our inpainting model, in
  which it is used to perform Bayesian experimental design for the purpose of
  guiding a sensor.
  %
  % We conclude with preliminary investigations into how this quality could be
  % beneficial when Bayesian optimal experimental design is used to guide a
  % sensor.
\end{abstract}

\begin{figure*}[b]
  \centering
  \includegraphics[width=\textwidth]{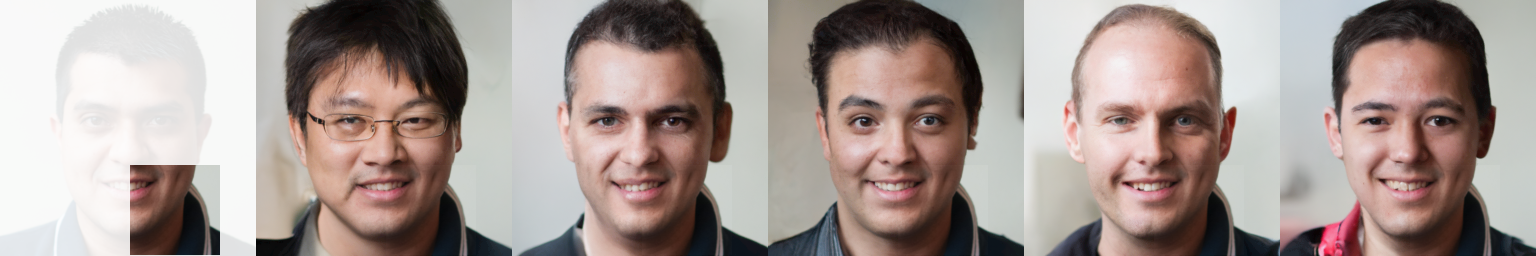}
  \includegraphics[width=\textwidth]{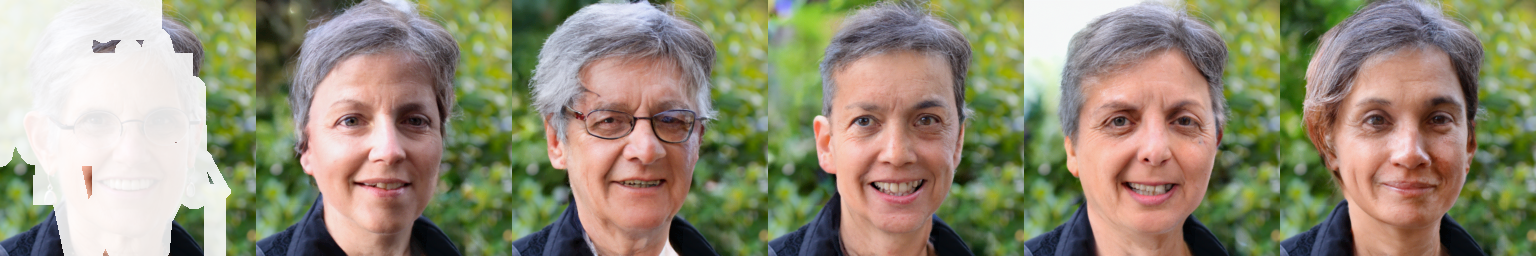}
  \caption{\textbf{Left column:} Images with most pixels masked out.
    \textbf{Rest:} Completions from our method.
  }
  \label{fig:headline}
\end{figure*}

\iffalse
TODO edits to paper:
- rename O_for
- more exanation of / introduction of O_for (eq 7)
- get rid of IPA-R completely
\fi

\section{Introduction}
A major challenge with applying variational auto-encoders (VAEs) to
high-dimensional data is the typically slow training times. For example,
training a state-of-the-art VAE~\citep{vahdat2020nvae,child2020very} on the
$256\times256$ FFHQ dataset~\citep{karras2019style} takes on the order of 1
GPU-year, but a state-of-the-art generative adversarial network
(GAN)~\citep{lin2021anycost,karras2020analyzing} can be trained on the same
dataset in a matter of GPU-weeks. One hypothesis for the cause of this disparity
is that, whereas the ``mass-covering'' training objective for a VAE forces it to
assign probability mass over the entirety of the data distribution, a GAN can
``cut corners'' by dropping modes~\citep{arora2017gans,arora2017generalization}.

We focus on the problem of \textit{conditional} generative modelling: given an
input (e.g.~a partially blanked-out image), we wish to map to a distribution
over outputs (e.g.~plausible completions of the image). Both conditional
GANs~\citep{zheng2019pluralistic,zhao2021large} and conditional
VAEs~\citep{sohn2015learning,ivanov2018variational} are applicable to this
problem, with the same disparity in training times that we described for their
unconditional counterparts. We present an approach based on the conditional VAE
framework but, to mitigate the associated slow training times, we design the
architecture so that we can incorporate pretrained unconditional VAEs. We show
that re-using publicly available pretrained models in this way can lead to
training times and sample quality competitive with GANs, while avoiding mode
dropping.

While requiring an existing pretrained model is a limitation, we note that:
\textbf{(\rom{1})} The unconditional VAE need not have been (pre-)trained on the same
dataset as the conditional model; we show unconditional models trained on
ImageNet are suitable for later use with various photo datasets.
\textbf{(\rom{2})} A single unconditional VAE can be used for later training of
conditional VAEs on any desired conditional generation tasks (e.g.~the same
image model may be later used for image completion or image colourisation).
\textbf{(\rom{3})} There is an increasing trend in the machine learning community towards
sharing large, expensively trained models~\citep{wolf2020transformers},
sometimes referred to as foundation models~\citep{bommasani2021opportunities}.
Most of the unconditional VAEs in our experiments use publicly-available
pretrained weights released by \citet{child2020very}. By presenting a use case
for foundation models in image modelling, we hope to encourage even more sharing
of pretrained weights in this domain.

% \todo[inline]{consider if below paragraph is necessary} We use an unconditional
% VAE as a prior over data, $p(\I{})$. We also require a likelihood
% $p(\partI{}|\I{})$: in the case of image completion, samples from this
% likelihood are copies of $\I{}$ with the values of some stochastically selected
% pixels obscured. Given such a prior and likelihood, Bayes' rule defines a
% posterior distribution over completions for any incomplete image $\partI{}$:
% \begin{equation}
%   \label{eq:bayes-rule}
%   p(\I{} | \partI{}) = \frac{p(\I{})p(\partI{}|\I{})}{\int p(\I{})p(\partI{}|\I{}) \mathrm{d}\I{}} .
% \end{equation}
% We use amortized inference in the VAE's latent space to model this distribution
% given any $\partI{}$.

% Our framework is applicable to any conditional generative modelling task but our
% experiments focus on the image domain, and
%
We demonstrate our approach on several conditional generation tasks in the image
domain but focus in particular on stochastic image completion: the problem of
inferring the posterior distribution over images given the observation of a
subset of pixel values.
For some applications such as photo-editing the implicit distribution defined by
GANs is good enough. We argue that our approach has substantial advantages when
image completion is used as part of a larger pipeline, and discuss one possible
instance of this in \cref{sec:boed}: Bayesian optimal experimental design (BOED)
for guiding a sensor or hard attention
mechanism~\citep{ma2018eddi,harvey2019near,rangrej2021achieving}. In this case,
missing modes of the posterior over images is likely to lead to bad decisions.
We show that our objective corresponds to the mass-covering KL divergence and so
covers the posterior well.
% , outperforming GANs which typically learn a
% distribution with a fraction of the data distribution's
% support~\citep{arora2017gans,arora2017generalization}.
%
This is supported empirically by results indicating that, not only is the visual
quality of our image completions (see \cref{fig:headline}) close to the
state-of-the-art~\citep{zhao2021large}, but our coverage of the ``true''
posterior over image completions is superior to that of any of our baselines.

% the visual quality of image completions
% produced by our method (see \cref{fig:headline}) is close to the
% state-of-the-art~\citep{zhao2021large}, and we show results indicating that our
% coverage of the ``true'' posterior over image completions is superior to that of
% any of our baselines.

% We additionally discuss applications of our approach to out-of-distribution
% detection since, as a by-product of learning a distribution over image
% completions, we learn a distribution over the input incomplete images.

\textbf{Contributions} We develop a method to cheaply convert pretrained
unconditional VAEs into conditional VAEs. The resulting training times and
sample quality are competitive with GANs, while the models avoid the
mode-dropping behaviour associated with GANs. Finally, we showcase a possible
application in Bayesian optimal experimental design that benefits from these
capabilities.

\section{Variational Auto-encoders}
We describe VAEs in terms of three components. \textbf{(\rom{1})} A decoder with
parameters $\theta\in\Theta$ maps from latent variables $\z{}$ to a distribution
over data $\I{}$, which we call $\pmodel{}(\I | \z{}; \theta)$. \textbf{(\rom{2})}
There is a prior over latent variables, $\pmodel{}(\z{};\theta)$. This may have
learnable parameters, which we consider to be part of $\theta$. Together, the
prior and decoder define a joint distribution, $\pmodel{}(\z{}, \I{}; \theta)$.
Finally, \textbf{(\rom{3})} an encoder with parameters $\phi\in\Phi$ maps from
data to an approximate posterior distribution over latent variables, $q(\z{}|\I{};
\phi) \approx \pmodel{}(\z{}|\I{}; \theta)$. Ideally, $\theta$ would be learned to
maximise the log likelihood $\log\pmodel{}(\I{}; \theta) = \log \int
\pmodel{}(\z{}, \I{}; \theta) \mathrm{d}\z{}$, averaged over training examples. Since
this is intractable, $\theta$ and $\phi$ are instead trained jointly to maximise
an average of the evidence lower-bound (ELBO) over each training example $\I{}
\sim \pdata{}(\cdot)$:
\begin{align}
  \EX_{\pdata{}(\I{})} \left[ \text{ELBO}(\theta, \phi, \I{}) \right] &= \EX_{\pdata(\I{})} \EX_{q(\z{}|\I{}; \phi)} \left[ \log\frac{\pmodel{}(\z{}; \theta)\pmodel{}(\I{}|\z{}; \theta)}{q(\z{}|\I{}; \phi)} \right] \label{eq:eelbo}\\
                                                                      &= -\mathcal{H}\left[ \pdata(\I) \right] - \kl[\big] { \pdata{}(\I{})q(\z{}|\I{}; \phi) }{ \pmodel{}(\z{}, \I; \theta) }. \label{eq:elbo-kl-joints}
\end{align}
The data distribution's entropy, $\mathcal{H}\left[ \pdata(\I) \right]$, is
typically a finite constant, and this is guaranteed in our experiments where
$\I{}$ is an image with discrete pixel values. Maximising the above objective
will therefore drive $\pmodel{}(\z{}, \I{}; \theta)$ towards
$\pdata{}(\I{})q(\z{}|\I{}; \phi)$, and so the marginal $\pmodel{}(\I{}; \theta)$
towards $\pdata{}(\I{})$. The KL divergence shown leads to mass-covering
behaviour from $\pmodel{}(\z{}, \I{}; \theta)$~\citep{bishop2006pattern} so
$\pmodel{}(\I{}; \theta)$ should assign probability broadly over the data
distribution $\pdata{}(\I{})$.
% This is important since, for the framing of
% image completion as inference to be well-posed, any observed subsets of image
% pixels should have non-zero probability under $\pmodel{}$.
For notational simplicity in the rest of the paper, parameters $\theta$ and
$\phi$ are not written when clear from the context.

\begin{figure*}[t]
  \centering
  \begin{subfigure}[b]{.32\textwidth}
    \centering
    \includegraphics[height=3.8cm]{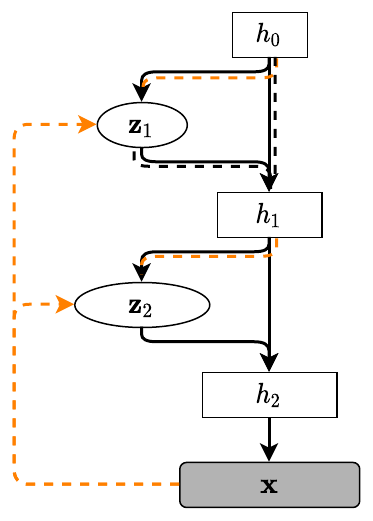}
    \caption{Estimating ELBO.}
    \label{fig:hierarchical-vae}
  \end{subfigure}
  \begin{subfigure}[b]{.32\textwidth}
    \centering
    \includegraphics[height=3.8cm]{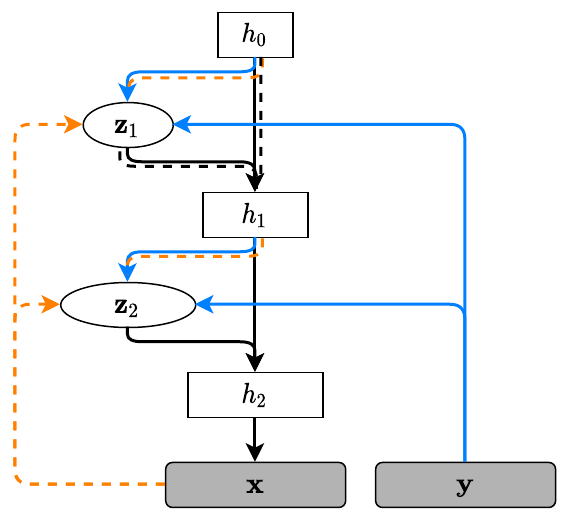}
    \caption{Estimating $\mathcal{O}_\mathrm{for}$.}
    \label{fig:forward-arch}
  \end{subfigure}
  \begin{subfigure}[b]{.32\textwidth}
    \centering
    \includegraphics[height=3.8cm]{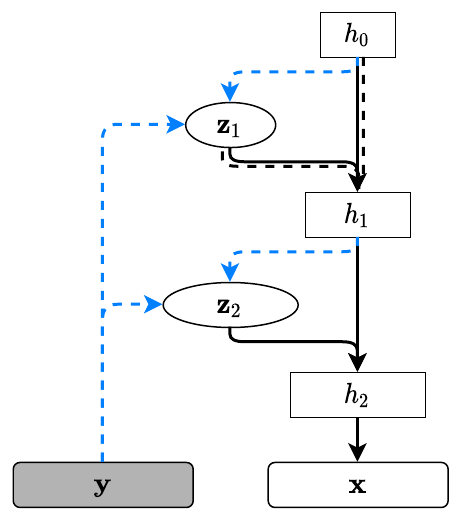}
    \caption{Sampling $\I{}\sim\pcomp{}(\cdot|\partI{})$.}
    \label{fig:reverse-arch}
  \end{subfigure}
  \vspace{-1mm}
  \caption{A hierarchical VAE architecture with $L=2$ layers of latent
    variables. Part (a) illustrates the computation of the ELBO for an
    unconditional VAE; part (b) illustrates the computation of our training
    objective $\mathcal{O}_\mathrm{for}$; and part (c) illustrates the drawing
    of conditional samples. The encoder is shown in orange; the prior and the
    decoder (which maintains a deterministic hidden state $h_{l}$) are both
    shown in black; and the partial encoder is shown in blue. The computation
    graph needed to sample $\z{}$ in each case is drawn with dashed lines, and
    the remainder of the computation graph is drawn with solid lines.}
  \label{fig:conditional-architectures}
  \vspace{-2mm}
\end{figure*}

Hierarchical
VAEs~\citep{gregor2015draw,kingma2016improving,sonderby2016ladder,klushyn2019learning}
partition the latent variables $\z{}$ in a way which has been found to improve the
fidelity of the learned $\pmodel{}(\I{})$, especially for the image
domain~\citep{vahdat2020nvae,child2020very}. In particular, they define $\z{}$ to
consist of $L$ disjoint groups, $\z{}_1,\ldots,\z{}_L$. The prior for each $\z{}_l$ can
depend on the previous groups through the factorisation
\begin{equation}
  \label{eq:hierarchical-prior}
  \pmodel{}(\z{}) = \prod_{l=1}^L \pmodel{}(\z{}_l|\z{}_{<l}).
\end{equation}
where $\z{}_{<l}$ is the null set for $l=1$ and $\{\z{}_1,\ldots,\z{}_{l-1}\}$
otherwise. \cref{fig:hierarchical-vae} shows the hierarchical VAE architecture
we base this work on, in which the dependency of the prior for each $\z{}_l$ on
$\z{}_{<l}$ is maintained via the decoder's hidden state $h_l$. The distribution
produced by the encoder for each $\z{}_l$ also depends on the previous hidden
state $h_{l-1}$ and so factorises as
$q(\z{}|\I{}) = \prod_{l=1}^L q(\z{}_l|\z{}_{<l}, \I{})$. We will parameterise
$\pmodel{}(\z{}_l|\z{}_{<l})$ and $q(\z{}_l|\z{}_{<l}, \I{})$ as diagonal
Gaussian distributions, as is common for hierarchical
VAEs~\citep{sonderby2016ladder,vahdat2020nvae,child2020very}.

\section{Amortized Inference in a Pretrained Artifact}
To convert an unconditional VAE architecture to a conditional architecture, we
introduce a \textit{partial encoder} with parameters $\partphi{}\in\Partphi{}$.
%
% This is fed a conditioning input $\partI{}$. For example, in the case of image
% completion, $\partI{}$ could be an image with some pixels masked out. The
% partial encoder then defines an approximate posterior over the latent variables,
% $\partq{}(\z{}|\partI{}; \partphi)$.
%
Given a conditioning input $\partI{}$ (which could be, e.g., an image with some
pixels masked out in the case of inpainting), the partial encoder defines an
approximate posterior over the latent variables,
$\partq{}(\z{}|\partI{}; \partphi)$.
Since the conditional generation task is
to approximate $\pdata{}(\I{}|\partI{})$, we use the partial encoder to define
\begin{equation}
  \label{eq:marginal-image-posterior}
  \pcomp{}(\I{}|\partI{}; \theta, \partphi) := \int \pmodel{}(\I{}|\z{}; \theta)\partq{}(\z{}|\partI{}; \partphi) \mathrm{d}\z{}
\end{equation}
with learnable parameters $\theta$ and $\partphi$. We can sample from
$\pcomp{}(\I{}|\partI{})$ by sampling $\z{}\sim \partq{}(\cdot|\partI{})$ and then
$\I{}\sim \pmodel{}(\cdot|\z{})$ as shown in \cref{fig:reverse-arch}. This defines
a conditional VAE architecture which, unique amongst related work with
high-dimensional $\I{}$ and
$\partI{}$~\citep{sohn2015learning,zheng2019pluralistic,ivanov2018variational,wan2021high},
has a decoder $\pmodel{}(\I{}|\z{}; \theta)$ with no dependence on $\partI{}$. This
decoder can therefore use an architecture identical to that of an unconditional
VAE and also, as we will show later, re-use unconditional VAE weights.

We now introduce some notation before describing our method further. Let the
distribution of paired data be $\pdata{}(\I{}, \partI{})$, and recall that
training an unconditional VAE matches two joint distributions: the distribution
of unconditional samples, $\pmodel{}(\z{}, \I{})$; and the distribution
resulting from sampling data $\I{}$ and then encoding it, $\pdata{}(\I{})q(\z{}|\I{})$.
% encoding samples from the dataset,
We define
extensions of each to include $\partI{}$:
\begin{align}
  \pmodel{}(\z{}, \I{}, \partI{}; \theta) &= \pmodel{}(\z{}; \theta)\pmodel{}(\I{}|\z{}; \theta)\pdata{}(\partI{}|\I{}), \label{eq:pmodel} \\
  r(\z{}, \I{}, \partI{}; \phi) &= \pdata{}(\I{}, \partI{})q(\z{}|\I{}; \phi), \label{eq:r}
\end{align}
where $\pdata{}(\partI{}|\I{})$ is a (potentially intractable) conditional
distribution under $\pdata{}(\I{}, \partI{})$.
Note that $\pmodel{}(\z{}, \I{}, \partI{}; \theta)$ and $r(\z{}, \I{}, \partI{};
\phi)$ are exactly the two distributions matched by the unconditional VAE
objective in \cref{eq:elbo-kl-joints} with an additional factor of
$\pdata{}(\partI{}|\I{})$.
Therefore, if the unconditional VAE represented by $\theta$ and $\phi$ is well
trained, $\pmodel{}(\z{}, \I{}, \partI{}; \theta)$ and
$r(\z{}, \I{}, \partI{}; \phi)$ will be close. From now on, we will use
$\pmodel{}$ and $r$ to refer to any marginals and conditionals of the above
joint distributions, with the specific marginal or conditional clear from
context.

\subsection{Training objective}

Our training objective, previously used for training conditional
VAEs~\citep{sohn2015learning,ivanov2018variational} and neural
processes~\citep{garnelo2018neural}, is
\begin{align}
  \label{eq:forward-elbo}
  \mathcal{O}_\mathrm{for}(\theta, \phi, \partphi{}) &= \EX_{\pdata{}(\I{}, \partI{})} \EX_{q(\z{}|\I{})} \left[ \log \frac{\pmodel{}(\tildeI{}|\z{})\partq{}(\z{}|\partI{})}{q(\z{}|\I{})} \right] &\leq \EX_{\pdata{}(\I{}, \partI{})} \left[ \log \pcomp{}(\tildeI{}|\partI{}) \right]
\end{align}
where $\tilde{\I{}}$ is the part of $\I{}$ we wish to predict. In general we can
set $\tildeI{}:=\I{}$ but for inpainting define $\tildeI{}$ to consist of only
the dimensions of $\I{}$ not observed in $\partI{}$, abusing notation by
ignoring the implication that $\tildeI{}$ is formally a function of $\partI{}$
as well as $\I{}$. Our architectures have pixel-wise independent
likelihoods so $\pmodel{}(\tildeI{}|\z{})$ is tractable in either case. \Cref{eq:forward-elbo}
lower-bounds $\log \pcomp{}(\tildeI{}|\partI{}) $ similarly to how the ELBO of
an unconditional VAE lower-bounds $\log \pmodel{}(\I{})$. The major difference
is that the prior, $\pmodel{}(\z{})$, is replaced by $\partq{}(\z{}|\partI{})$.
This is reflected in \cref{fig:forward-arch} where each $\z{}_l$ is conditioned
on $\partI{}$ via the partial encoder.
Similar to standard estimators for an unconditional hierarchical VAE's ELBO,
reduced-variance estimates of \cref{eq:forward-elbo} can be obtained by
computing KL divergences between $q(\z{}_l|\z{}_{<l},\I{})$ and
$\partq{}(\z{}_l|\z{}_{<l},\partI{})$ analytically for each layer $l$ (see
\cref{supp:kl-estimates} for details).

We are particularly interested in the properties of the learned partial encoder.
Recall the joint distribution $r(\z{}, \I{}, \partI{}; \phi) = \pdata{}(\I{},
\partI{})q(\z{}|\I{}; \phi)$. Then $r(\z{}|\partI{}; \phi)$ is the intractable
posterior given by marginalising out $\I{}$ and conditioning on $\partI{}$. We
find that fitting $\partphi{}$ to maximise $\mathcal{O}_\mathrm{for}(\theta,
\phi, \partphi{})$ is equivalent to minimising the mass-covering KL divergence
from $r(\z{}|\partI{}; \phi)$ to $\partq{}(\z{}|\partI{}; \partphi)$. We formalise
this statement in the following theorem, which is proven in
\cref{proof:forward-kl}.
\begin{theorem} \label{theorem:forward-kl} For any set $\Partphi$ of
  permissible values of $\partphi{}$, and for any $\theta\in\Theta$ and
  $\phi\in\Phi$,
  \begin{equation} \label{eq:forward-theorem}
    \argmax_{\partphi{} \in \Partphi} \mathcal{O}_\mathrm{for}(\theta, \phi, \partphi{}) = \argmin_{\partphi{} \in \Partphi} \EX_{\pdata{}(\partI{})} \left[ \kl[\big]{ r(\z{}|\partI{}; \phi) }{ \partq{}(\z{}|\partI{}; \partphi) } \right].
  \end{equation}
\end{theorem}
Due to the mass-covering properties of this ``forward'' KL
divergence~\citep{bishop2006pattern}, this theorem indicates that 
%
% Learning $\partq{}(\z{}|\partI{}; \partphi)$ to minimise this ``forward'' KL
% divergence leads to mass-covering behaviour~\citep{bishop2006pattern}, and so
%
the learned $\partq{}(\z{}|\partI{}; \partphi)$ should have good coverage of all
modes of $r(\z{}|\partI{}; \phi)$. Intuitively, the resulting diverse samples of
latent variables $\z{}\sim \partq{}(\cdot|\partI{}; \partphi)$ should lead to
diverse samples of $\tildeI{} \sim \pmodel{}(\cdot|\z{})$ which cover the
``true'' posterior $\pdata{}(\tildeI{}|\partI{})$. We formalise this in
\cref{proof:forward-kl} by showing that maximising $\mathcal{O}_\mathrm{for}$
also minimises an upper-bound on a KL divergence in $\tildeI{}$-space.

% , and subsequently samples of $\tildeI{}$
% given $z$, are therefore likely to be diverse with good coverage of the ``true''
% posterior $\pdata{}(\tildeI{}|\partI{})$.

\subsection{Faster training with a pretrained VAE}
To justify using weights trained as part of an unconditional VAE we present
\cref{theorem:joint-training}
\begin{theorem} \label{theorem:joint-training} Assume that we have a
  sufficiently expressive encoder and decoder that there exist parameters
  $\optimal{\theta}\in\Theta$ and $\optimal{\phi}\in\Phi$ which make the
  unconditional VAE objective (\cref{eq:eelbo}) equal to its upper bound of
  $-\mathcal{H}\left[ \pdata(\I) \right]$. Assume also that $\tildeI{}$ is
  defined such that there is a mapping from $(\tildeI{},\partI{})$ to $\I{}$ and
  that the mutual information ${\mutinf{} := \EX_{\pmodel{}(\tildeI{}, \partI{},
      \z{}; \theta^*)} \left[ \log \frac{\pmodel{}(\tildeI{},\partI{}|\z{}; \theta^*)
      }{ \pmodel{}(\tildeI{}|\z{}; \theta^*)\pmodel{}(\partI{}|\z{}; \theta^*) }
    \right]}$ is zero (see discussion below). Then, given a sufficiently expressive
  partial encoder,
  \begin{equation} \label{eq:joint-training}
    \max_{\partphi{}} \mathcal{O}_\mathrm{for}(\optimal{\theta}, \optimal{\phi}, \partphi{}) = \max_{\theta, \phi, \partphi{}} \mathcal{O}_\mathrm{for}(\theta, \phi, \partphi{}).
  \end{equation}
\end{theorem}
%
% \textcolor{red}{ See \cref{proof:joint-training} for a proof and further explanation of the assumption that $\mutinf{}=0$.}
%
This is proven in \cref{proof:joint-training} and implies that we can use values
of $\theta$ and $\phi$ learned using the unconditional VAE objective. Then to
train a conditional generative model we need only optimise $\partphi{}$. This
leads to faster convergence, as well as faster training iterations since we only
need to compute gradients for, and perform update steps on, the partial
encoder's parameters $\partphi{}$. For all of our experiments in
\cref{sec:experiments} we use pretrained models released by
\citet{child2020very}, leveraging between 2 GPU-weeks and 1 GPU-year of
unconditional VAE training for each dataset. We name our method IPA (Inference
in a Pretrained Artifact).

\textcolor{black}{\Cref{theorem:joint-training} relies on the assumption that
  the mutual information $\mutinf{}$ is zero; as we argue in
  \cref{proof:joint-training}, this is true for inpainting and also
  ``approximately'' holds if $\partI{}$ consists of high-level features. When
  lower level features are conditioned on, e.g.~for super-resolution, there may
  be a significant gap between the left- and right- hand sides of
  \cref{eq:joint-training}. } \Cref{theorem:joint-training} also applies only if
the unconditional VAE parameters are learned on the same dataset as the
conditional VAE is trained on; otherwise there will be a mismatch between the
form of $\pdata{}$ used in \cref{eq:eelbo} to fit $\theta^*$ and $\phi^*$, and
the form of $\pdata{}$ implicit in the $\mathcal{O}_\mathrm{for}$ objective.
However we find empirically that we can use unconditional VAE parameters trained
on ImageNet~\citep{deng2009imagenet} with IPA on several other photo datasets.

\begin{table*}
  \small
  \caption{Image completion results. Best performance is shown in \textbf{bold},
    and second best is \underline{underlined}. In the last row, $t$ denotes the
    ``temperature'' parameter \citep{child2020very}.}
  \label{tab:results-completion}
  \centering
  \begin{tabular}{lccccccccc}
    \toprule
    \multicolumn{1}{r}{} & \multicolumn{3}{c}{CIFAR-10}  & \multicolumn{3}{c}{ImageNet-64}  & \multicolumn{3}{c}{FFHQ-256} \\
    \cmidrule(r){2-4} \cmidrule(r){5-7} \cmidrule(r){8-10} % \cmidrule(r){6-7} \cmidrule(r){8-9}
    Method        & \quad FID$\downarrow$ \quad     & P-IDS$\uparrow$  & LPIPS-GT$\downarrow$ & \quad FID$\downarrow$ \quad    & P-IDS$\uparrow$  & LPIPS-GT$\downarrow$& \quad FID$\downarrow$ \quad    & P-IDS$\uparrow$  & LPIPS-GT$\downarrow$ \\
    \midrule
    ANP                        & 30.03               & ~~5.86              & .0447                         & -                 & -                 & -                & 39.95              & ~~0.93              & .256 \\
    CE                         & 21.92               & ~~4.77              & .0628                         & -                 & -                 & -                & 39.02              & ~~0.66              & .267 \\
    RFR                        & 44.35               & ~~2.76              & .0883                         & -                 & -                   & -              & 72.50              & ~~0.46              & .271 \\
    PIC                        & 14.73               & ~~5.95              & .0332                         & 40.0              & 0.24                & .170           & 11.60              & ~~2.76              & .169 \\
    CoModGAN                   & ~~\underline{9.65}  & 11.59               & .0326                         & 20.2              & ~~7.09              & .160           & ~~\textbf{2.33}    & \textbf{13.57}      & .143 \\
    IPA-R                      & 19.21               & ~~8.56              & .0330                         & 28.8              & 6.46                & .166           & ~~8.82             & ~~4.56              & .142 \\
    IPA (ours)                 & 10.50               & \underline{13.24}   & \textbf{.0262}                & \underline{18.9}  & ~~\underline{9.20}  & \underline{.138}  & ~~3.93             & ~~7.79              & \underline{.123} \\
    IPA$_{t=0.85}$ (ours)      & ~~\textbf{8.61}     & \textbf{14.19}      & \underline{.0263}             & \textbf{15.1}     & \textbf{11.26}    & \textbf{.133}     & ~~\underline{3.29} & ~~\underline{8.50}  & \textbf{.117} \\
    \bottomrule
  \end{tabular}
  \vspace{-1em}
\end{table*}

\section{Experiments} \label{sec:experiments}

\paragraph{Comparison to image completion baselines}

We create an IPA image completion model based on the VD-VAE unconditional
architecture~\citep{child2020very}, and evaluate it for image completion on
three datasets: CIFAR-10~\citep{krizhevsky2009learning},
ImageNet-64~\citep{deng2009imagenet}, and FFHQ-256~\citep{karras2019style}. We
compare against four baselines: Co-Modulated Generative Adversarial Networks
(CoModGAN)~\citep{zhao2021large}; Pluralistic Image Completion
(PIC)~\citep{zheng2019pluralistic}; Context Encoders
(CE)~\citep{pathak2016context}; and Attentive Neural Processes
(ANP)~\citep{kim2019attentive}. We also considered two further methods: we show
samples from VQ-VAE~\citep{peng2021generating} (but not quantitative results
which were too slow to compute because it takes roughly one minute per test
image); and we report results for Recurrent Feature Reasoning for Image
Inpainting (RFR)~\citep{li2020recurrent} with the caveat that it is slow to run
on images with many missing pixels and so, although it used a similar
computational budget to the other models, its training did not converge.

Given pretrained unconditional VAE parameters, IPA is faster to train than the
best-performing baseline, CoModGAN. IPA takes 115 GPU-hours to train on
CIFAR-10, and under 7 GPU-weeks on FFHQ-256. The CoModGAN models are trained for
270 GPU-hours and 8 GPU-weeks respectively. We provide more training details in
\cref{supp:exp-details}.

We report the FID~\citep{heusel2017gans} and P-IDS~\citep{zhao2021large} metrics
between a set of sampled completions from each method and a reference set.
Broadly speaking, these measure the sample quality.
To investigate how well the samples capture all modes of
$\pdata{}(\I{}|\partI{})$, we also report the LPIPS-GT. We compute this using
LPIPS~\citep{zhang2018unreasonable}, a measure of distance between two images.
Specifically, we compute the average over test pairs $(\I{}, \partI{})$ of
$\min_{k=1}^K(\text{LPIPS}(\I{}^{(k)} , \I))$, with each $\I{}^{(k)} \sim
\pcomp{}(\cdot | \partI)$. As $K \rightarrow \infty$, the LPIPS-GT should tend
to zero if the ground truth is always within the support of
$\pcomp(\I{}|\partI{})$. If not, the LPIPS-GT will remain high, penalising
methods which miss modes of the posterior. We use $K=100$. We confirm in
\cref{supp:additional-results} that the LPIPS-GT correlates with diversity
metrics used by related work~\citep{zhu2017toward,li2020multimodal}.

For the image completion tasks, we sample from $\pdata{}(\I{}, \partI{})$ by
first sampling an image $\I{}$ from the dataset, and then sampling an
image-sized binary mask $m$ from the freeform mask distribution used by
\citet{zhao2021large}, which is itself based on \citet{yu2018generative}. We
then set $\partI{} = \texttt{concatenate}(\I{} \odot m, m)$. Here, $\odot$ is a
pixel-wise multiplication operation which removes information from the missing
pixels. The concatenation is performed along the channel dimension and makes it
possible to distinguish between unobserved pixels and zero-valued pixels.

\begin{figure*}[t]
  %\vspace{-.1cm}
  \centering
  \includegraphics[width=\textwidth]{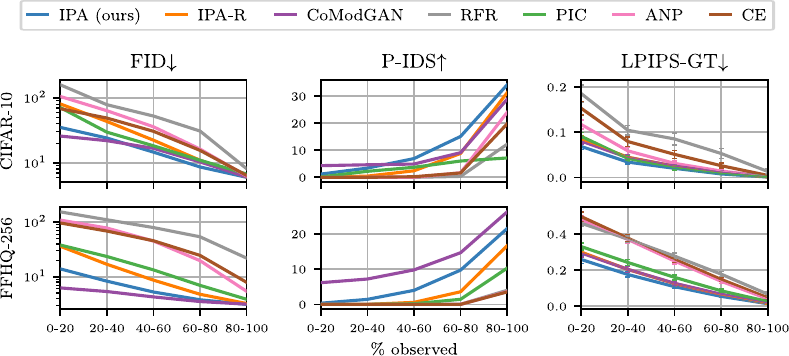}
  %\vspace{-.6cm}
  \caption{Test metrics on CIFAR-10 and FFHQ-256, plotted as a function of
    the mask distribution. Error bars on LPIPS-GT show the standard error of our
    estimate for a single trained network.}
  \label{fig:metrics}
  \vspace{-.5cm}
\end{figure*}

For evaluation, since the number of observed pixels in freeform masks varies
considerably, we follow \citet{zhao2021large} and partition the mask
distribution by conditioning the procedure to return a mask with the proportion
of pixels observed within some range (0-20\%, 20-40\%, and so on) and report
metrics for each range separately in \cref{fig:metrics} (or
\cref{supp:additional-results} for ImageNet-64). To summarise the
overall performance in \cref{tab:results-completion}, we sample masks from a
uniformly-weighted mixture distribution over these five partitions. In terms of
the LPIPS-GT scores in \cref{tab:results-completion}, IPA outperforms the best
baselines by roughly 20\%. \Cref{fig:metrics} shows that there is an improvement
for any proportion of observed pixels. This suggests that IPA produces reliably
diverse samples with good coverage of $\pdata{}(\I{}|\partI{})$. In contrast, we
believe that the GAN-based approaches occasionally miss modes of
$\pdata{}(\I{}|\partI{})$ and can therefore fail to capture the ground-truth.
This hypothesis is supported by samples from CoModGAN we display in
\cref{supp:comodgan-failure}. In terms of sample fidelity, as measured by both
FID and P-IDS, IPA outperforms all baselines on CIFAR-10 when over $40\%$ of the
image is observed, and on ImageNet-64 when over $20\%$ is observed. IPA comes
second to CoModGAN when less is observed and on FFHQ-256.

\begin{figure*}[t]
  \vspace{-.1cm}
  \centering
  \includegraphics[scale=1]{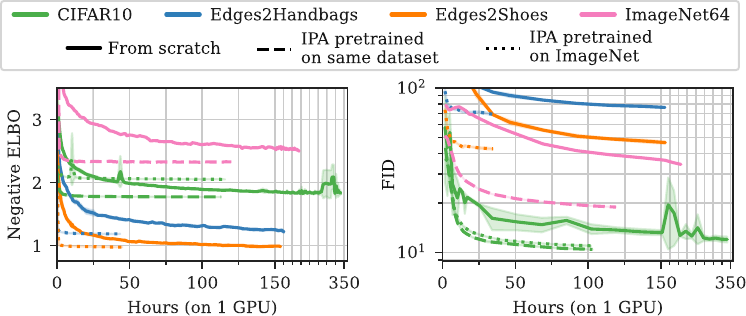}
  \caption{ELBO (\cref{eq:forward-elbo} computed with $\tildeI{} := \I{}$) and
    FID during training using IPA with pretraining on the same dataset, IPA with
    pretraining on ImageNet, and when trained from scratch. Error bars show
    standard deviations computed with 3 runs. IPA makes training faster and
    lower-variance.}
  \label{fig:training}
  \vspace{-.3cm}
\end{figure*}

\paragraph{Edges-to-photos}
We provide an additional demonstration of IPA on the Edges2Shoes and
Edges2Handbags datasets~\citep{isola2016image}, where the task is to generate an
image conditioned on the output of an edge detector applied to that image. We
downsample the datasets to $64\times64$ so that we can use unconditional VAEs
pretrained on ImageNet~\citep{deng2009imagenet} at this resolution by
\citet{child2020very}. We show in \cref{fig:training} that IPA is useful for
these tasks, and provide further discussion below. The images generated are
diverse and photorealistic, as shown in \cref{supp:image-samples}.

\paragraph{Effectiveness of pretraining}
We now seek to determine how important the pretrained unconditional VAE weights
are to IPA. We compare IPA with conditional VAEs which use IPA's architecture
but are trained from scratch, which we will refer to as ``from-scratch''
baselines. That is, $\theta$ and $\phi$ are randomly initialised and trained to
maximise \cref{eq:forward-elbo} (with $\tildeI{} := \I{}$) along with
$\partphi$.
With an infinite training budget, the end-to-end training of the from-scratch
baselines is likely to lead them to outperform any IPA models. Nevertheless it
is apparent from \cref{fig:training} that, in the more realistic situation of a
finite training budget, using IPA can be beneficial. This is the case even for
training budgets of up to a few GPU-weeks on the relatively small CIFAR-10
dataset. In fact, even with only a couple of days of training, IPA on CIFAR-10
(with CIFAR-10 pretraining) achieves better FID and ELBO scores than the
from-scratch baseline trained for several weeks. For ImageNet-64, IPA performs
better after 4 hours than the from-scratch baseline does after a week.
For Edges2Handbags and Edges2Shoes, training with IPA for 2 days yields
performance similar to or better than training with the from-scratch baseline
for 1 week, as measured by the ELBO. This is despite IPA on these datasets using
a trained ImageNet-64 model rather than a model pretrained on those specific
datasets, supporting our suggestion that the dataset used for pretraining need
not exactly match what IPA is then trained on.
Measured by the FID score, IPA's performance is even more appealing: wherever
ELBOs are similar between IPA and the from-scratch baselines, IPA achieves a
significantly better FID score.
We see that IPA pretrained on ImageNet is less effective for CIFAR-10
than it is for the edges-to-photos datasets, but it improves on the from-scratch
baseline in terms of ELBO for the first 30 hours of training, and in terms of
FID until the from-scratch baseline is trained for several weeks.

\paragraph{An alternative training objective}
In \cref{tab:results-completion} and \cref{fig:metrics}, we report results for
IPA-R, a variation of IPA with a different training objective corresponding to a
mode-seeking KL divergence. IPA almost always outperforms IPA-R, but we
nonetheless provide a full description of IPA-R in \cref{supp:ipa-r}.

\section{Inpainting for Bayesian Experimental Design} \label{sec:boed}

In this section, we explore a potential application for stochastic image completion that
requires a faithful representation of the posterior $\pdata{}(\I{}|\partI{})$.
In particular, we consider whether it is possible to automatically target a
chest x-ray at areas most likely to reveal abnormalities. This could avoid the
need to scan the entire chest and so bring benefits including reducing the
patient's radiation exposure.
While doing so is not possible with standard x-ray machines (which do not take
multiple scans consecutively), and would need to be extensively validated before
use in a clinical setting, we believe this is a worthwhile avenue to explore.
Specifically, our imagined system performs a series of x-ray scans, each
targeted at only a small portion of the area of interest. We can select the
coordinates $\coord_t = (x_t, y_t)$ of the location to scan at each step $t$, and
this selection can be informed by what was observed in the previous scans. The
task we consider is how to select $\coord{}_t$ to be maximally informative. In
particular, assume we wish to infer a variable $v$ representing, e.g., whether
the patient has a particular illness. Bayesian optimal experimental design
(BOED)~\citep{chaloner1995bayesian} provides a framework to select a value of
$\coord{}_t$ that is maximally informative about $v$.
It involves taking a Bayesian perspective on the problem of estimating $v$. We
have one posterior distribution over $v$ after taking scans at
$\coord{}_1,\ldots,\coord{}_{t-1}$ and another (typically lower entropy) distribution after
conditioning on a scan at $\coord{}_t$ as well.
The \textit{expected information gain}, or EIG, quantifies the utility of the
choice of $\coord{}_t$ as the expected difference in entropy between these two
distributions. Using BOED involves estimating the EIG and selecting the scan
location, $\coord{}_t$, to minimise it.

\begin{figure}[t]
  \centering
  \hspace{-.75cm}
  \begin{minipage}{.4\textwidth}
    \centering
    \includegraphics[scale=.77]{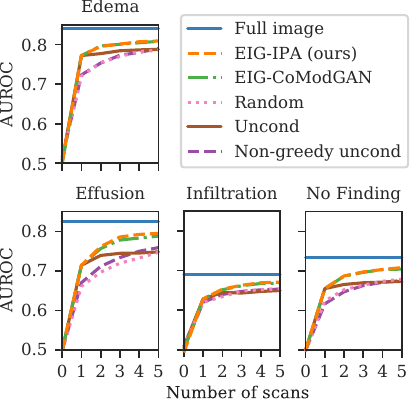}
  \end{minipage}
  ~
  \begin{minipage}{0.54\textwidth}
    \centering
    \includegraphics[scale=.77]{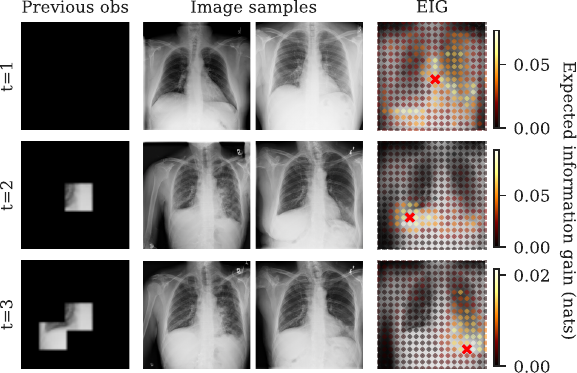}
  \end{minipage}
  \caption{\textbf{Left:} Classification AUROC scores after $1,\ldots,5$ scans
    chosen with each method. Scores for the ``EIG-'' methods more quickly approach the
    upper bound achieved by processing the full image. \textbf{Right:}
    Visualisation of BOED used to select three scan locations for diagnosing
    `Effusion'. The left column shows the observations made prior to each time
    step. We then show samples from IPA (or the dataset when $t=1$). The
    rightmost column shows the EIG overlaid on the pixel-space average of
    sampled images, with the optimal $\coord{}_t$ marked by a red cross.}
  \label{fig:boed}
  \vspace{-.4cm}
\end{figure}

We use an estimator for the EIG similar to that of \citet{harvey2019near}. It
requires two components:
\textbf{(\rom{1})} A neural network trained to classify $v$ given a series of
scans at locations $\coord{}_1,\ldots,\coord{}_t$. This outputs a classification distribution
which we denote $g(v|f_{\coord{}_1,\ldots,\coord{}_t}(\I{}))$, where $f_{\coord{}_1,\ldots,\coord{}_t}$ is a
function mapping from an image to the values of the pixels observed by scans at
$\coord{}_1,\ldots,\coord{}_t$. We use this classification distribution as an approximation of
the posterior over $v$, whose entropy we attempt to minimise by performing BOED.
\textbf{(\rom{2})} A method for sampling image completions conditioned on some
observed pixel values $f_{\coord{}_1,\ldots,\coord{}_{t-1}}(\I{})$. \citet{harvey2019near}
used a “stochastic image completion” module which contributed significant
complexity to their method. We entirely replace this with IPA.

Let the pixel values observed so far be $\partI{}_{\coord{}_1,\ldots,\coord{}_{t-1}} =
f_{\coord{}_1,\ldots,\coord{}_{t-1}}(\I{})$ for a latent image $\I{}$. Given these, we
estimate the EIG of location $\coord{}_t$ as
\begin{align}
  \label{eq:new-eig}
  \text{EIG}(\coord{}_t;& \partI{}_{\coord{}_1,\ldots,\coord{}_{t-1}}) \approx \overbrace{\mathcal{H} \left[ \frac{1}{N} \sum_{n=1}^N g(\cdot|f_{\coord{}_1,\ldots,\coord{}_t}(\I{}^{(n)})) \right]}^{\text{entropy after $t-1$ scans}} - \overbrace{\frac{1}{N} \sum_{n=1}^N  \mathcal{H} \left[ g(\cdot|f_{\coord{}_1,\ldots,\coord{}_t}(\I{}^{(n)})) \right]}^{\text{expected entropy after $t$ scans}},
\end{align}
where $\I{}^{(1)},\ldots,\I{}^{(N)}$ are sampled image completions from IPA
given $\partI{}_{\coord{}_1,\ldots,\coord{}_{t-1}}$. In \cref{sec:supp-boed} we report
hyperparameters, provide further details of our EIG estimator, and compare it to
the estimators used in related work. To select $\coord{}_t$, we simply estimate
$\text{EIG}(\coord{}_t; \partI{}_{\coord{}_1,\ldots,\coord{}_{t-1}})$ for many different values of
$\coord{}_t$ and select the value which maximises it. This process of selecting $\coord{}_t$
and then taking a scan is repeated for each $t=1,\ldots,T$.

We experiment on the NIH Chest X-ray 14 dataset~\citep{wang2017chestx} at
$256\times256$ resolution.
%
%
% , in which each image is labelled with the presence or
% absence of 14 illnesses. We set $v$ to be a binary label indicating the
% presence/absence of a particular illness. A simple extension, with appropriate
% training data, could be to define $v$ to be the severity of an illness, and
% therefore also infer the severity.
%
We simulate a scanner which returns a $64 \times 64$ pixel patch from this
image, and the task is to diagnose the binary presence or absence of an illness.
We run separate experiments diagnosing each of edema, effusion, infiltration and
``no finding'' (an additional label meaning there are no diagnosed illnesses).
With appropriate data, this framework could be extended to also infer the
severity of a given illness.
We envisage BOED being used to select scan locations for an x-ray without
necessarily performing an automated diagnosis. However, to quantify the
informativeness of the chosen locations, \cref{fig:boed} shows the results of
using $g$ to perform a diagnosis, or classification, based on the chosen scan
locations.
Since the conditional distribution $g$ (used to estimate the EIG) depends on
which illness we are classifying, the choice of scan locations is different in
each case.
We compare against a baseline where the image completion is performed by
CoModGAN (our best-performing image completion baseline) rather than IPA, as
well as numerous baselines which choose scan locations without image
completion; see \cref{sec:supp-boed} for details.

Our method (denoted EIG-IPA) narrowly but consistently outperforms EIG-CoModGAN.
%
% We hypothesise that this is due to CoModGAN's tendency to sometimes produce
% almost no diversity in its completions, even when the observed pixels are
% uninformative.
%
We hypothesise that this is due to the aforementioned tendency of CoModGAN to
sometimes collapse to a single mode of the posterior, and exhibit an example of
this behaviour on the x-ray dataset in \cref{supp:image-samples}. In the BOED
context, such ``overconfident'' image completion could lead to salient scan
locations being ignored. Nonetheless, both EIG-IPA and EIG-CoModGAN
significantly outperform the other baselines, giving performance much closer to
the upper bound of a CNN with access to the entire image. Another benefit of the
``EIG-'' approaches is that the choice of scan locations is highly
interpretable; we can see why a particular location was chosen with
visualisations similar to the right of \cref{fig:boed}. This shows the sampled
images $\I{}^{(n)}$ and the estimated EIG for each $\coord{}_t$. In \cref{sec:supp-boed}, we
show that we can further quantify the contribution of each $\I{}^{(n)}$ to the
estimated EIG for each $\coord{}_t$.

\section{Related Work}
\paragraph{Inference in pretrained VAEs}
Several prior studies perform conditional generation using a previously trained
unconditional VAE.
Like us, \citet{rezende2014stochastic,nguyen2016plug,wu2018conditional} do so
through inference in the VAE's latent space. However, they use non-amortized
inference (Gibbs sampling, variational inference, and MCMC respectively),
leading to slow sampling times for any new $\partI{}$.
\citet{duan2019pre} learn variational distributions over $\z{}$ for every possible
value of $\partI{}$, but this is not possible when $\partI{}$ is
high-dimensional or continuous-valued.
\citet{yeh2017semantic} fit the latent variables of a GAN given observations,
but this is neither amortized nor probabilistic.

\paragraph{Conditional VAEs}
Past research on conditional
VAEs~\citep{sohn2015learning,zheng2019pluralistic,ivanov2018variational,wan2021high}
has generally been unable to take advantage of pretrained weights as we have due
to a difference in architectures: unlike almost all prior work, the IPA decoder
does not receive $\partI{}$ as input. The dependence between $\partI{}$ and the
decoder's output must therefore be expressed solely through the conditional
distribution over the latent variables, $\partq{}(\z{}|\partI{})$. This is a crucial
difference because it means that the decoder can have exactly the same
architecture as that of an unconditional VAE. This is key to letting us copy the
pretrained weights of an unconditional VAE to speed up training. The exception
to the above is \citet{ma2018eddi} who, like us, use a conditional VAE decoder
with no dependence on $\partI{}$. Their training objective and use case are
different, however, and they do not consider using pretrained models or use an
architecture which can scale to photorealistic images. Leveraging unconditional
VAEs lets us drastically reduce the computational budget required to train a
conditional VAE. We believe that this paper is the first to demonstrate
photorealistic image completion with conditional VAEs at resolutions as high as
$256\times256$.
Another benefit of the decoder having no dependence on $\partI{}$ is that it
makes impossible the ``posterior collapse'' phenomenon discussed by
\citet{zheng2019pluralistic}, in which a conditional VAE's decoder learns to
ignore $\z{}$ and produce outputs conditioned solely on $\mathbf{y}$.

\paragraph{Image completion}
Early work on image completion, both
before~\citep{bertalmio2000image,bertalmio2001navier,ballester2001filling,levin2003learning,criminisi2003object}
and after~\citep{kohler2014mask,ren2015shepard} deep learning became the
dominant approach, aimed to deterministically fill in missing pixels in images.
Even many methods incorporating generative adversarial networks (GANs), which
were introduced by \citet{goodfellow2014generative} as a tool to learn
distributions, have been found to result in little or no diversity in the
completions produced for a given
input~\citep{song2018spg,yu2018generative,yu2019free,pathak2016context,iizuka2017globally}.
However, some recent methods have managed to obtain diverse completions using
the GAN framework~\citep{zhao2020uctgan,zhao2021large,liu2021pd}.
Another approach is to sample low-resolution images using
VAEs~\citep{zheng2019pluralistic,peng2021generating} or transformer-based
sequence models~\citep{zheng2021tfill,wan2021high}, and then use a GAN for
upsampling. In contrast, we use a VAE to model image completions at the full
resolution. As well as ensuring diverse coverage of the posterior, using such a
likelihood-based model enables applications such as out-of-distribution
detection for inputs $\partI{}$, which we demonstrate in \cref{supp:ood}.
Another related approach is that of \citet{song2020score}, who present a
stochastic differential equation-based image model. This can be used for image
completion but sampling is slow.

\section{Discussion and Conclusion}

We have presented IPA, a method to adapt an unconditional VAE into a conditional
model. Image completions generated with IPA are close to the state-of-the-art in
terms of visual fidelity, and improve on all baselines in terms of their
coverage of the posterior as measured by LPIPS-GT. This high-fidelity coverage
of the posterior makes IPA ideal for use in Bayesian optimal experimental
design, as demonstrated.
%
% In addition, IPA has all the benefits of a
% likelihood-based method, such as the potential to perform out-of-distribution
% detection.
% 
\textcolor{black}{ Our theoretical results suggest that, for the applications
  presented, using the weights of an unconditional VAE is approximately as good
  as training a conditional VAE from scratch. We note however, that there are
  applications for which these results will not hold (e.g.~super-resolution).
  They also provide no guarantees when pretraining is performed on a different
  dataset, although we show empirically that IPA can still be effective in this
  case. }
Future work may look more rigorously at these settings or further improve the
image samples by, e.g., using a partial encoder with more expressive
distributions.
Preliminary experiments using normalizing flows helped improve the ELBO, but
with little impact on the FID scores.

\subsubsection*{Ethics statement}
Our proposed method obtains more complete coverage of a conditional image
distribution than our baselines. Since we so not directly push the
state-of-the-art in terms of realism (and thereby aid, e.g., the creation of
``deepfakes''), we are not aware of any obviously nefarious applications which
will benefit.
A potential positive impact is that, by providing better coverage of the
posterior distribution, our work leads to better representation of minority
groups in certain applications.

For a real-world application of the possible medical imaging procedure outlined
in \cref{sec:boed}, we note that various challenges exist. In addition to the
development of appropriate x-ray imaging hardware, these include more thoroughly
investigating potential sources of failure “in the wild”, and mitigating dataset
biases. Working on the latter in particular is important to ensure that any
decisions made are not worse for groups that are underrepresented in the
dataset. Mitigating this would require more careful dataset curation, and
possibly a new source of data.

\subsubsection*{Acknowledgments}
We acknowledge the support of the Natural Sciences and Engineering Research
Council of Canada (NSERC), the Canada CIFAR AI Chairs Program, and the Intel
Parallel Computing Centers program. Additional support was provided by UBC's
Composites Research Network (CRN), and Data Science Institute (DSI). This
research was enabled in part by technical support and computational resources
provided by WestGrid (www.westgrid.ca), Compute Canada (www.computecanada.ca),
and Advanced Research Computing at the University of British Columbia
(arc.ubc.ca). WH acknowledges support by the University of British Columbia’s
Four Year Doctoral Fellowship (4YF) program.

\bibliography{references}
\bibliographystyle{iclr2022_conference}

\clearpage

\newcounter{longappendix}
\setcounter{longappendix}{1}   % 0 for small appendix with few figures, 1 for
                               % anonymous submission, 2 for big appendix

\ifnum\value{longappendix}>1{
  \section*{Appendix for `Image Completion via Inference in Deep Generative Models'}
  This is the long form of the appendix for `Image Completion via Inference in
  Deep Generative Models' (see \url{https://arxiv.org/abs/2102.12037} for main
  paper).
} \else {} \fi

\appendix

\section{Notation}
We provide \cref{tab:notation} for reference as a concise list of definitions
for each symbol used. More thorough explanations are provided where symbols are
introduced in the text

\begin{table*}
  \caption{Definitions for symbols used.}
  \label{tab:notation}
  \centering
  \begin{tabular}{rp{11cm}}
    \toprule
    Symbol    & Definition   \\
    \midrule
    $\z{}$                                   & VAE's latent variables. \\
    $\I{}$                                & Data we learn a generative model of. \\
    $\partI{}$                            & Data on which our generative model is conditioned. \\
    $\tildeI{}$                           & The part of $\I{}$ which the conditional VAE models, defined as all pixels not in $\partI{}$ for image completion, or simply $\tildeI := \I{}$ for other conditional generation tasks. \\
    $\theta$                              & Parameters for VAE's prior and decoder. \\
    $\phi$                                & Parameters for VAE's encoder. \\
    $\pdata{}(\I{}, \partI{})$            & Distribution from which we assume training/test instances are i.i.d.~samples. \\
    $\pmodel{}(\z{})$                        & Learned prior of VAE. Implicitly parameterised by $\theta$ when not otherwise specified. \\
    $\pmodel{}(\I{}|\z{})$                   & Distribution output by VAE's decoder given $\z{}$. Implicitly parameterised by $\theta$ when not otherwise specified. \\
    $q(\z{}|\I{})$                           & Distribution output by VAE's encoder given $\I{}$. Implicitly parameterised by $\phi$ when not otherwise specified. \\
    $\pmodel{}(\z{}, \I{}, \partI{})$        & Joint distribution defined as $\pmodel{}(\z{})\pmodel{}(\I{}|\z{})\pdata{}(\partI{}|\I{})$. \\
    $r(\z{},\I{}, \partI{})$                 & Joint distribution defined as $\pdata{}(\I{}, \partI{})q(\z{}|\I{})$. \\
    $\partphi$                            & Partial encoder parameters. \\
    $\partq{}(\z{}|\partI{})$                 & Distribution output by partial encoder. Implicitly parameterised by $\partphi$. \\
    $\pcomp{}(\I{}|\partI{})$             & Distribution over from which images are sampled given a conditioning input. Defined as $\int \pmodel{}(\I{}|\z{})\partq{}(\z{}|\partI{}) \mathrm{d}\z{}$. \\
    $\unp{}(\z{}, \I{}, \partI{})$           & Distribution with parameters $\optimal{\theta}$ and $\optimal{\phi}$ that are optimal as defined in \cref{proof:joint-training}. Defined equivalently as $\pmodel{}(\z{}, \I{}, \partI{}; \optimal{\theta})$ or $\pdata{}(\I{}, \partI{}) q(\z{}|\I{}; \optimal{\phi})$.  \\
    \midrule
    \multicolumn{2}{l}{Symbols for Bayesian optimal experimental design:} \\
    $\coord{}_t$                                 & The location scanned at time $t$. \\
    $v$                                   & Variable we wish to infer. \\
    $\partI{}_{\coord{}_1,\ldots,\coord{}_t}$             & Image in which all pixel values are missing except for those observed by scans at locations $\coord{}_1,\ldots,\coord{}_t$. \\
    $f_{\coord{}_1,\ldots,\coord{}_t}$                    & Function mapping from an image $\I{}$ to $\partI{}_{\coord{}_1,\ldots,\coord{}_t}$. \\
    $g(v|\partI{})$                       & Classification distribution for $v$ given partially observed image $\partI{}$. \\
    \bottomrule
  \end{tabular}
  \vspace{-1em}
\end{table*}

\section{Expanded derivations}
\subsection{Expanded derivation of \cref{eq:elbo-kl-joints}}
Here we expand on the steps to show that the expectation of the ELBO in
\cref{eq:eelbo} is equivalent to the lower bound on the negative entropy of
$\pdata{}(\I{})$ in \cref{eq:elbo-kl-joints}.
\begin{align}
  \EX_{\pdata{}(\I{})} \left[ \text{ELBO}(\theta, \phi, \I{}) \right] &= \EX_{\pdata(\I{})} \EX_{q(\z{}|\I{}; \phi)} \left[ \log\frac{\pmodel{}(\z{}; \theta)\pmodel{}(\I{}|\z{}; \theta)}{q(\z{}|\I{}; \phi)} \right] \label{eq:eelbo-repeat}\\
                                                                      &= \EX_{\pdata(\I{})q(\z{}|\I{}; \phi)} \left[ \log\frac{\pmodel{}(\z{}, \I{}; \theta)}{q(\z{}|\I{}; \phi)} \right] \\
                                                                      &= \EX_{\pdata(\I{})q(\z{}|\I{}; \phi)} \left[ \log\pdata{}(\I{}) + \log\frac{\pmodel{}(\z{}, \I{}; \theta)}{\pdata{}(\I{})q(\z{}|\I{}; \phi)} \right] \\
                                                                      &= \EX_{\pdata(\I{})} \left[ \log\pdata(\I{}) \right] + \EX_{\pdata(\I{})q(\z{}|\I{}; \phi)} \left[ \log\frac{\pmodel{}(\z{}, \I{}; \theta)}{\pdata{}(\I{})q(\z{}|\I{}; \phi)} \right] \\
                                                                      &= -\mathcal{H}\left[ \pdata(\I) \right] - \kl[\big] { \pdata{}(\I{})q(\z{}|\I{}; \phi) }{ \pmodel{}(\z{}, \I; \theta) } \label{eq:elbo-kl-joints-repeat}
\end{align}
as written in \cref{eq:elbo-kl-joints}.

\subsection{Expanded derivation of \cref{eq:forward-elbo}} \label{supp:forward-elbo-bound-deriv}
Here we prove \cref{eq:forward-elbo}, which states that IPA's training
objective is a lower bound on $\EX_{\pdata{}(\I{}, \partI{})} \left[ \log
  \pcomp{}(\tildeI{}|\partI{}) \right]$.
\begin{align}
  \mathcal{O}_\mathrm{for}(\theta, \phi, \partphi{}) &= \EX_{\pdata{}(\I{}, \partI{})} \EX_{q(\z{}|\I{})} \left[ \log \frac{\pmodel{}(\tildeI{}|\z{})\partq{}(\z{}|\partI{})}{q(\z{}|\I{})} \right] \\
                                                   &\leq \EX_{\pdata{}(\I{}, \partI{})} \left[ \log \EX_{q(\z{}|\I{})} \left[ \frac{\pmodel{}(\tildeI{}|\z{})\partq{}(\z{}|\partI{})}{q(\z{}|\I{})} \right]  \right] \\
                                                   &= \EX_{\pdata{}(\I{}, \partI{})} \left[ \log \EX_{\partq{}(\z{}|\partI{})} \left[ \pmodel{}(\tildeI{}|\z{}) \right]  \right] \\
                                                   &= \EX_{\pdata{}(\I{}, \partI{})} \left[ \log \pcomp{}(\tildeI{}|\partI{}) \right],
\end{align}
where the final step follows from the definition of $\pcomp{}$ in
\cref{eq:marginal-image-posterior}.

% \section{Proofs} \label{supp:proofs}

\section{Proof and discussion of Theorem~\ref{theorem:forward-kl}} \label{proof:forward-kl}
\subsection{Proof}
\textbf{Statement.}
\textit{
  For any set $\Partphi$ of
  permissible values of $\partphi{}$, and for any $\theta\in\Theta$ and
  $\phi\in\Phi$,
  \begin{equation}
    \argmax_{\partphi{} \in \Partphi} \mathcal{O}_\mathrm{for}(\theta, \phi, \partphi{}) = \argmin_{\partphi{} \in \Partphi} \EX_{\pdata{}(\partI{})} \left[ \kl[\big]{ r(\z{}|\partI{}) }{ \partq{}(\z{}|\partI{}) } \right]. \tag{\ref{eq:forward-theorem}}
  \end{equation}
}

\textbf{Proof.} We can decompose $\mathcal{O}_\mathrm{for}$ as follows.
Starting by multiplying both sides of the fraction by the intractable
conditional distribution $r(\z{}|\partI{})$,
\begin{align}
  \mathcal{O}_\mathrm{for}(\theta, \phi, \partphi{}) &= \EX_{r(\z{},\I{},\partI{})} \left[ \log \frac{\pmodel{}(\tildeI{}|\z{})\partq{}(\z{}|\partI{})r(\z{}|\partI{})}{q(\z{}|\I{})r(\z{}|\partI{})} \right]\\
                                                   &= \EX_{r(\z{},\I{},\partI{})} \left[ \log \frac{\pmodel{}(\tildeI{}|\z{})r(\z{}|\partI{})}{q(\z{}|\I{})} \right] - \EX_{r(\z{},\partI{})} \left[ \log\frac{r(\z{}|\partI{})}{\partq{}(\z{}|\partI{})} \right] \\
                                                   &= C(\theta, \phi) - \EX_{\pdata{}(\partI{})} \left[ \kl[\big]{ r(\z{}|\partI{}) }{ \partq{}(\z{}|\partI{}) } \right]   \label{eq:constant-and-forward-kl}
\end{align}
where $C(\theta, \phi)$ is a grouping of terms which do not depend on the
partial encoder's parameters $\partphi{}$. Any values of $\partphi{}$ which
maximise $\mathcal{O}_\mathrm{for}(\theta, \phi, \partphi{})$ will therefore also
minimise $\EX_{\pdata{}(\partI{})} \left[ \kl[\big]{ r(\z{}|\partI{}) }{
    \partq{}(\z{}|\partI{}) } \right]$ for the same values of $\theta$ and $\phi$,
which proves the theorem.

\subsection{Bound on $\tildeI{}$-space divergence}
The mass-covering KL divergence in \Cref{theorem:forward-kl} indicates that
samples $\z{}\sim \partq{}(\cdot|\partI{})$ will be diverse (when $r(\z{}|\partI{})$ is
diverse/uncertain). One would expect that samples of $\tildeI{}$ conditioned on these
diverse samples of $\z{}$ will also be diverse. Here we make a more formal argument
by showing that minimising $\mathcal{O}_\mathrm{for}$ corresponds to miminising
an upper-bound on an expected mass-covering KL divergence in $\tildeI{}$-space.
Starting from the result in \cref{supp:forward-elbo-bound-deriv},
\begin{align}
  \mathcal{O}_\mathrm{for}(\theta, \phi, \partphi{}) &\leq \EX_{\pdata{}(\I{},\partI{})} \left[ \log \pcomp{}(\tildeI{}|\partI{}) \right] \\
                                                     &= \EX_{\pdata{}(\I{},\partI{})} \left[ \log \pdata{}(\tildeI{}|\partI{}) \right] - \EX_{\pdata{}(\partI{})}\left[ \kl{\pdata{}(\tildeI{}|\partI{})}{\pcomp{}(\tildeI{}|\partI{})} \right].
\end{align}
Rearranging this gives
\begin{align}
  \EX_{\pdata{}(\partI{})}\left[ \kl{\pdata{}(\tildeI{}|\partI{})}{\pcomp{}(\tildeI{}|\partI{})} \right]  \leq \EX_{\pdata{}(\I{},\partI{})} \left[ \log \pdata{}(\tildeI{}|\partI{}) \right] - \mathcal{O}_\mathrm{for}(\theta, \phi, \partphi{}).
\end{align}
Maximising $\mathcal{O}_\mathrm{for}$ therefore minimises an upper-bound on an
expected mass-covering KL divergence in $\tildeI{}$-space, providing a further suggestion that
$\tildeI{}\sim\pcomp{}(\cdot|\partI{})$ will be diverse.

\section{Proof and discussion of Theorem~\ref{theorem:joint-training}}  \label{proof:joint-training}
\subsection{Proof}

\textbf{Statement.} \textit{Assume that we have a sufficiently expressive
  encoder and decoder that there exist parameters $\optimal{\theta}\in\Theta$
  and $\optimal{\phi}\in\Phi$ which make the unconditional VAE objective
  (\cref{eq:eelbo}) equal to its upper bound of $-\mathcal{H}\left[ \pdata(\I)
  \right]$. Assume also that the mutual information ${\mutinf{} :=
  \EX_{\pmodel{}(\tildeI{}, \partI{}, \z{}; \theta^*)} \left[ \log
    \frac{\pmodel{}(\tildeI{},\partI{}|\z{}; \theta^*) }{ \pmodel{}(\tildeI{}|\z{};
      \theta^*)\pmodel{}(\partI{}|\z{}; \theta^*) } \right] =
  \EX_{\pmodel{}(\tildeI{}, \partI{}, \z{}; \theta^*)} \left[ \log
    \frac{\pmodel{}(\tildeI{}|\z{}, \partI{}; \theta^*) }{ \pmodel{}(\tildeI{}|\z{};
      \theta^*) } \right]}$ is zero (meaning that $\tildeI{}$ and $\partI{}$ are
  conditionally independent given $\z{}$) and $\tildeI{}$ is defined such that there
  exists a mapping from $(\tildeI{},\partI{})$ to $\I{}$. Then, given a
  sufficiently expressive partial encoder $\partq{}(\z{}|\partI{}; \partphi)$, }
\begin{equation} \tag{\ref{eq:joint-training}}
  \max_{\partphi{}} \mathcal{O}_\mathrm{for}(\optimal{\theta}, \optimal{\phi}, \partphi{}) = \max_{\theta, \phi, \partphi{}} \mathcal{O}_\mathrm{for}(\theta, \phi, \partphi{}).
\end{equation}

\textbf{Proof.} We have used, and will continue to use, $\pmodel{}(\cdot)$ and
$q(\cdot)$ to refer to joint distributions parameterised by $\theta$ and $\phi$
respectively. Since $\optimal{\theta}$ and $\optimal{\phi}$ are defined as the
maximisers of the unconditional ELBO in \cref{eq:elbo-kl-joints}, and given our
assumption of sufficient expressivity, we have $\pmodel{}(\z{}, \I{};
\optimal{\theta}) = \pdata{}(\I{}) q(\z{}|\I{}; \optimal{\phi})$. We can therefore
introduce notation for the equal joint distributions: $\optimal{p}(\z{}, \I{},
\partI{}) = \pmodel{}(\z{}, \I{}, \partI{}; \optimal{\theta}) = \pdata{}(\I{},
\partI{}) q(\z{}|\I{}; \optimal{\phi})$. We note that we can also write the joint
distribution $\optimal{p}(\z{}, \I{}, \tildeI{}, \partI{}) = \pdata{}(\I{},
\partI{}) q(\z{}|\I{}; \optimal{\phi})p(\tildeI{}|\I{},\partI{})$, where
$p(\tildeI{}|\I{},\partI{})=\delta_{g(\I{},\partI{})}(\tildeI{})$ is the Dirac
distribution mapping $\I{}$ and $\partI{}$ deterministically to
$\tildeI{}=g(\I{},\partI{})$.
Recalling that we define $\tildeI{}$ as either equal to $\I{}$ or as the
dimensions of $\I{}$ not observed in $\partI{}$, such a deterministic $g$ will
always exist.

% We note that defining this
% joint distribution also implicitly defines $\optimal{p}(\z{}, \tildeI{}, \partI{})$
% as a marginal (over any elements of $\I{}$ not in $\tildeI{}$).

We consider the decomposition of $\mathcal{O}_\mathrm{for}$ shown in
\cref{eq:constant-and-forward-kl} and remind the reader that, given a
sufficiently expressive partial encoder, the maximisation over $\partphi$ will
always make the KL divergence zero. That is,
\begin{equation}
  \max_{\partphi{}} \mathcal{O}_\mathrm{for}(\theta, \phi, \partphi{}) = C(\theta, \phi) = \EX_{r(\z{},\I{},\partI{})} \left[ \log \frac{\pmodel{}(\tildeI{}|\z{})r(\z{}|\partI{})}{q(\z{}|\I{})} \right] \label{eq:forward-max-phi}
\end{equation}
for any $\theta$ and $\phi$. Using $\optimal{\theta}$ and $\optimal{\phi}$, and
therefore using $\unp{}$ in place of $\pmodel{}$, $r$ and $q$, we can expand the
left-hand side of \cref{eq:joint-training}.
\begin{align}
  \max_{\partphi{}} \mathcal{O}_\mathrm{for}(\optimal{\theta}, \optimal{\phi}, \partphi{}) &= \EX_{\unp{}(\z{},\I{},\partI{})} \left[ \log \frac{\unp{}(\tildeI{}|\z{})\unp{}(\z{}|\partI{})}{\unp(\z{}|\I{})} \right] \\
                                                                                           &= \EX_{\unp{}(\z{},\I{},\partI{})} \left[ \log \frac{\unp{}(\tildeI{}|\z{},\partI{})\unp{}(\z{}|\partI{})}{\unp(\z{}|\I{})} \right] - \EX_{\unp{}(\z{},\tildeI{},\partI{})} \left[ \log\frac{\unp{}(\tildeI{}|\z{},\partI{})}{\unp{}(\tildeI{}|\z{})} \right] \\
                                                                                           &= \EX_{\unp{}(\z{},\I{},\partI{})} \left[ \log \frac{\unp{}(\tildeI{}|\z{},\partI{})\unp{}(\z{}|\partI{})}{\unp(\z{}|\I{})} \right] - \mutinf{}\\
                                                                                           &= \EX_{\unp{}(\z{},\I{},\partI{})} \left[ \log \frac{\unp{}(\z{},\tildeI{}|\partI{})}{\unp(\z{}|\I{})} \right] - \mutinf{}\\
                                                                                           &= \EX_{\unp{}(\z{},\I{},\partI{})} \left[ \log \frac{\unp{}(\z{}|\tildeI{},\partI{})\unp{}(\tildeI{}|\partI{})}{\unp(\z{}|\I{})} \right] - \mutinf{}. \label{eq:pre-mapping}
\end{align}
% Here, we note that we can make the factorisation $\optimal{p}(\z{}, \I{},
% \tildeI{}, \partI{}) = \pdata{}(\I{}, \partI{}) q(\z{}|\I{};
% \optimal{\phi})\delta(\tildeI{}|\I{},\partI{})$, where
% $\delta(\tildeI{}|\I{},\partI{})$ is the Dirac distribution mapping $\I{}$ and
% $\partI{}$ deterministically to $\tildeI{}$, which we have previously made
% implicit. 
The factorisation $\optimal{p}(\z{}, \I{}, \tildeI{}, \partI{}) = \pdata{}(\I{},
\partI{}) q(\z{}|\I{}; \optimal{\phi})\delta(\tildeI{}|\I{},\partI{})$ implies that
we can express $\unp{}(\z{}|\tildeI{},\partI{})$ with a marginalisation:
$\unp{}(\z{}|\tildeI{},\partI{})=\int \unp{}(\z{}|\I{})
\unp{}(\I{}|\tildeI{},\partI{}) \mathrm{d}\I{}$. Since we have assumed that
there is a deterministic function $f$ mapping $(\tildeI{},\partI{})$ to $\I{}$,
$\unp{}(\I{}|\tildeI{},\partI{})$ is a Dirac distribution and the
marginalisation becomes $\unp{}(\z{}|\tildeI{},\partI{}) = \int \unp{}(\z{}|\I{})
\delta(\I{}|\tildeI{},\partI{}) \mathrm{d}\I{} =
\unp{}(\z{}|\I{}=f(\tildeI{},\partI{}))$. We therefore proceed, rewriting
$\unp{}(\z{}|\tildeI{},\partI{})$ as $\unp{}(\z{}|\I{})$:
\begin{align}
  \max_{\partphi{}} \mathcal{O}_\mathrm{for}(\optimal{\theta}, \optimal{\phi}, \partphi{}) &= \EX_{\unp{}(\z{},\I{},\partI{})} \left[ \log \frac{\unp{}(\z{}|\I{})\unp{}(\tildeI{}|\partI{})}{\unp(\z{}|\I{})} \right] - \mutinf{} \label{eq:post-mapping}\\
                                                                                           &= \EX_{\unp{}(\I{},\partI{})} \left[ \log \unp{} (\tildeI{}|\partI{}) \right] - \mutinf{}\\
                                                                                           &= \EX_{\pdata{}(\I{},\partI{})} \left[ \log \pdata{} (\tildeI{}|\partI{}) \right] - \mutinf{}. \label{eq:max-partphi-bound}
\end{align}
Now that we have this expansion of the left-hand side of
\cref{eq:joint-training}, we can derive a related upper-bound on the right-hand
side as follows. For any $\theta$, any $\phi$, and any $\partphi$, we have
from \cref{supp:forward-elbo-bound-deriv}:
\begin{align}
  \mathcal{O}_\mathrm{for}(\theta, \phi, \partphi{}) &\leq \EX_{\pdata{}(\I{},\partI{})} \left[ \log \pcomp{}(\tildeI{}|\partI{}) \right] \\
                                                     &= \EX_{\pdata{}(\I{},\partI{})} \left[ \log \pdata{}(\tildeI{}|\partI{}) \right] - \EX_{\pdata{}(\partI{})}\left[ \kl{\pdata{}(\tildeI{}|\partI{})}{\pcomp{}(\tildeI{}|\partI{})} \right] \\
                                                     &\leq \EX_{\pdata{}(\I{},\partI{})} \left[ \log \pdata{}(\tildeI{}|\partI{}) \right].
\end{align}
This bound must also hold for the maximum over $\theta$, $\phi$, and
$\partphi$, and so is an upper-bound on the right-hand side of \cref{eq:joint-training}:
\begin{align}
  \max_{\theta, \phi, \partphi{}} \mathcal{O}_\mathrm{for}(\theta, \phi, \partphi{}) &\leq \EX_{\pdata{}(\I{},\partI{})} \left[ \log \pdata{}(\tildeI{}|\partI{}) \right]. \label{eq:max-all-bound}
\end{align}

By relating \cref{eq:max-partphi-bound} and \cref{eq:max-all-bound}, we have:
\begin{align}
  \max_{\theta, \phi, \partphi{}} \mathcal{O}_\mathrm{for}(\theta, \phi, \partphi{}) &\leq  \max_{\partphi{}} \mathcal{O}_\mathrm{for}(\optimal{\theta}, \optimal{\phi}, \partphi{}) + \mutinf{}. \label{eq:bound-1}
\end{align}
We then point out that we can also obtain an inequality in the other direction: since
the right-hand side of \cref{eq:joint-training} is a maximisation of the left-hand side over $\theta$ and
$\phi$, it must be greater than or equal to the left-hand side:
\begin{align}
  \max_{\partphi{}} \mathcal{O}_\mathrm{for}(\optimal{\theta}, \optimal{\phi}, \partphi{}) &\leq  \max_{\theta, \phi, \partphi{}} \mathcal{O}_\mathrm{for}(\theta, \phi, \partphi{})  . \label{eq:bound-2}
\end{align}
Combining \cref{eq:bound-1} and \cref{eq:bound-2}, we have:
\begin{align}
  0 \leq \max_{\theta, \phi, \partphi{}} \mathcal{O}_\mathrm{for}(\theta, \phi, \partphi{}) - \max_{\partphi{}} \mathcal{O}_\mathrm{for}(\optimal{\theta}, \optimal{\phi}, \partphi{}) \leq \mutinf{}. \label{eq:two-bounds}
\end{align}
We now have bounds in both directions on the difference between the right- and
left-hand sides. To conclude the proof, we repeat the assumption that
$\mutinf{}=0$. The upper- and lower- bounds of \cref{eq:two-bounds} are
therefore both zero, so we have the final result that ${\max_{\partphi{}}
\mathcal{O}_\mathrm{for}(\optimal{\theta}, \optimal{\phi}, \partphi{}) =
\max_{\theta, \phi, \partphi{}} \mathcal{O}_\mathrm{for}(\theta, \phi,
\partphi{})}$.

\subsection{Discussion of the assumption that $\mutinf{}=0$}
The above proof of \cref{theorem:joint-training} relies on the assumption that
$\mutinf{} = \EX_{\unp{}(\tildeI{}, \partI{}, \z{})} \left[ \log
  \frac{\unp{}(\tildeI{}|\z{}, \partI{}) }{ \unp{}(\tildeI{}|\z{}) } \right] = 0$.
That is, under the joint distribution $p^*(\z{},\I{},\partI{})$, $\tildeI{}$ and
$\partI{}$ should be conditionally independent given $\z{}$. We now argue that this
is true for image completion, and ``close to'' satisfied \textcolor{black}{when
  $\mathbf{y}$ consists of high-level image features. This means that
  \cref{theorem:joint-training} is informative for inpainting, our
  edges-to-photos experiments and, e.g., class-conditional generation but may
  not even approximately hold for, e.g., super-resolution or image
  colourisation.}

\paragraph{Image completion}
In the VAE architecture we use, the likelihood $\pmodel{}(\I{}|\z{})$ is pixel-wise
independent. In other words, if we write $\I{}$ as a set of pixels
$\{\I{}_1,\ldots,\I{}_N\}$, we can factorise $\pmodel{}(\I{}|\z{})$ as
$\prod_{i=1}^N\pmodel{}(\I{}_i|\z{})$. This means that the conditional mutual
information between any two disjoint subsets of $\I{}$ (conditioned on $\z{}$) will
be zero. Given our definition of $\tildeI{}$ for image completion as the the set
of purely unobserved pixels, we have by construction that $\tildeI{}$ and
$\partI{}$ are disjoint. Therefore, the assumption that $\mutinf{}=0$ is
satisfied for image completion.

\paragraph{When $\mutinf{}$ is non-zero.}
When $\mutinf{}$ is non-zero, we can still use it as an upper-bound on the
difference between the left- and right-hand side of \cref{eq:joint-training}
using \cref{eq:two-bounds}.
Moreover, we posit that $\mutinf{}$ will typically be small when $\partI{}$
represents high-level features (e.g.~the edges in Edges2Photos or a class label)
and we have a pixel-wise independent likelihood. This is because any high-level
structure cannot be modelled by the pixel-wise independent likelihood
$\unp{}(\tildeI{}|\z{})$ and so must be captured by $\z{}$. Intuitively, high-level
features $\partI{}$ are unlikely to be informative about any pixel-level
variations if all image structure spanning multiple pixels ($\z{}$) is already
known. This is equivalent to saying that $\mutinf{}$, the mutual information
between $\tildeI{}$ and $\partI{}$ conditioned on $\z{}$, is likely to be small.

\section{An alternative training objective}
\label{supp:ipa-r}
In this section, we describe an alternative objective previously used by
\citet{ma2018eddi}. We find experimentally that its performance is typically
worse than $\mathcal{O}_\mathrm{for}$, but it has the advantage of only
requiring data $\partI{} \sim \pdata{}(\cdot)$ during training, and not $\I{}$.
This may allow training in settings such as
outpainting~\citep{sabini2018painting}. The objective can be interpreted as a
lower bound on $\log \pmodel{}(\partI{})$:
\begin{align} \label{eq:reverse-obj}
  \mathcal{O}_\text{rev}(\theta, \phi, \partphi{}) &= \EX_{\pdata{}(\partI{})}\EX_{\partq{}(\z{}|\partI{})} \left[ \log\frac{\pmodel{}(\z{},\partI{})}{\partq{}(\z{}|\partI{})} \right] \leq \EX_{\pdata{}(\partI{})} \left[ \log \pmodel(\partI{}) \right].
\end{align}
Computing this objective requires a computation graph similar to that shown in
\cref{fig:reverse-arch}. Note one caveat with this objective is that using it
involves computing and differentiating $\pmodel{}(\z{},\partI{})$, which requires a
tractable likelihood $\pdata{}(\partI{}|\I{})$ (based on the definition of
$\pmodel{}$ in \cref{tab:notation}). This is possible in image completion, but
not in the general case of conditional generation.

The following theorem says that learning $\partphi{}$ to maximise this objective
is equivalent to minimising another KL divergence.
\begin{theorem} \label{theorem:reverse-kl} For any set $\Partphi$ of
  permissible values of $\partphi{}$, and for any $\theta\in\Theta$ and
  $\phi\in\Phi$,
  \begin{equation}
    \argmax_{\partphi{} \in \Partphi} \mathcal{O}_\text{rev}(\theta, \phi, \partphi{}) = \argmin_{\partphi{} \in \Partphi} \EX_{\pdata{}(\partI{})} \left[ \kl[\big] { \partq{}(\z{}|\partI{}; \partphi) }{ \pmodel{}(\z{}|\partI{}; \theta) } \right].
  \end{equation}
\end{theorem}

% \subsection{Proof of Theorem~\ref{theorem:reverse-kl}} \label{proof:reverse-kl}

% \textbf{Statement.} \textit{
%   For any set $\Partphi$ of
%   permissible values of $\partphi{}$, and for any $\theta\in\Theta$ and
%   $\phi\in\Phi$,
%   \begin{equation}
%     \argmax_{\partphi{} \in \Partphi} \mathcal{O}_\text{rev}(\theta, \phi, \partphi{}) = \argmin_{\partphi{} \in \Partphi} \EX_{\pdata{}(\partI{})} \left[ \kl[\big] { \partq{}(\z{}|\partI{}) }{ \pmodel{}(\z{}|\partI{}) } \right].
%   \end{equation}
% }
\textbf{Proof.}
We show this by decomposing the objective:
\begin{align} \label{eq:reverse-obj-supp}
  \mathcal{O}_\text{rev}(\theta, \phi, \partphi{}) &= \EX_{\pdata{}(\partI{})\partq{}(\z{}|\partI{})} \left[ \log\frac{\pmodel{}(\z{},\partI{})}{\partq{}(\z{}|\partI{})} \right] \\
                                                   &= \EX_{\pdata{}(\partI{})\partq{}(\z{}|\partI{})}\Big[ \log \pmodel{}(\partI{}) - \log\frac{\partq{}(\z{}|\partI{})}{\pmodel{}(\z{}|\partI{})} \Big] \\
  &= \EX_{\pdata{}(\partI{})}\Big[ \log \pmodel{}(\partI{}) - \kl[\big] { \partq{}(\z{}|\partI{}) }{ \pmodel{}(\z{}|\partI{}) } \Big] \\
                                                   &= D(\theta) - \EX_{\pdata{}(\partI{})}\Big[ \kl[\big] { \partq{}(\z{}|\partI{}) }{ \pmodel{}(\z{}|\partI{}) } \Big]
\end{align}
where $D(\theta)$ is not dependent on $\partphi$. Therefore, for any $\theta$
and $\phi$, any values of $\partphi\in\Partphi$ which maximise
$\mathcal{O}_\text{rev}(\theta, \phi, \partphi{})$ will also minimise
$\EX_{\pdata{}(\partI{})}\Big[ \kl[\big] { \partq{}(\z{}|\partI{}) }{
  \pmodel{}(\z{}|\partI{}) } \Big]$, proving the theorem.

\Cref{theorem:reverse-kl} implies that maximising $\mathcal{O}_{\text{rev}}$
will minimise a ``reverse'' KL divergence, causing $\partq{}(\z{}|\partI{})$ to
exhibit mode-seeking behaviour. We denote the method of training with this
objective IPA-R (inference in a pretrained artifact with the reverse KL).
As with IPA, we use pretrained $\theta$ and $\phi$ when training IPA-R, which we
justify in the following paragraph. With IPA-R, we also use the pretrained
encoder parameters $\phi$ as an initialisation for $\partphi{}$, which makes
training more stable.

% \paragraph{Using pretrained weights with $\mathcal{O}_\text{rev}$}
To justify the use of a pretrained $\theta$ and $\phi$ with
$\mathcal{O}_\text{rev}$, we show that the objective $\mathcal{O}_\text{rev}$
has a property analogous to that described by \cref{theorem:joint-training}.
\begin{theorem}
  \label{theorem:joint-training-rev}
  Assume we have a sufficiently expressive encoder and decoder such that there
  exists parameters $\optimal{\theta}\in\Theta$ and $\optimal{\phi}\in\Phi$
  which make the unconditional VAE objective (\cref{eq:eelbo}) equal to its
  upper bound of $-\mathcal{H}\left[ \pdata(\I) \right]$. Then, given a
  sufficiently expressive partial encoder $\partq{}(\z{}|\partI{}; \partphi)$,
  \begin{equation*}
    \max_{\partphi{}} \mathcal{O}_\text{rev}(\optimal{\theta}, \optimal{\phi}, \partphi{}) = \max_{\theta, \phi, \partphi{}} \mathcal{O}_\text{rev}(\theta, \phi, \partphi{}).
  \end{equation*}
\end{theorem}

\textbf{Proof.} We will prove \cref{theorem:joint-training-rev} by showing that
both the quantity on the left-hand side and the quantity on the right-hand side
are equal to the negative of the entropy of $\pdata{}(\I{})$. We begin with the
left-hand side,
\begin{align}
  \max_{\partphi{}} \mathcal{O}_\text{rev}(\optimal{\theta}, \optimal{\phi}, \partphi{}) &= \max_{\partphi{}} \EX_{\pdata{}(\partI{})\partq{}(\z{}|\partI{})} \left[ \log\frac{\unp{}(\z{},\partI{})}{\partq{}(\z{}|\partI{})} \right].
\end{align}
This is a variational lower-bound which is made tight by setting $\partphi{}$
such that $\partq{}(\z{}|\partI{}) = \unp{}(\z{}|\partI{})$:
\begin{align}
  \max_{\partphi{}} \mathcal{O}_\text{rev}(\optimal{\theta}, \optimal{\phi}, \partphi{}) &= \EX_{\pdata{}(\partI{})} \left[ \log \unp(\partI{}) \right] = - \mathcal{H} \left[ \pdata{}(\partI{}) \right]. \label{eq:final-lhs-a}
\end{align}
Now we consider the right-hand side.
\begin{align}
  \max_{\theta, \phi, \partphi{}} \mathcal{O}_\text{rev}(\theta, \phi, \partphi{}) &= \max_{\theta, \phi, \partphi{}} \EX_{\pdata{}(\partI{})\partq{}(\z{}|\partI{})} \left[ \log\frac{\pmodel{}(\z{},\partI{})}{\partq{}(\z{}|\partI{})} \right]
\end{align}
This is another variational lower-bound made tight by setting $\partphi{}$
such that $\partq{}(\z{}|\partI{}) = \pmodel{}(\z{}|\partI{})$:
\begin{align}
  \max_{\theta, \phi, \partphi{}} \mathcal{O}_\text{rev}(\theta, \phi, \partphi{}) &= \max_{\theta, \phi} \EX_{\pdata{}(\partI{})} \left[ \log \pmodel{}(\partI{}) \right] \\
                                                                                   &= - \mathcal{H} \left[ \pdata{}(\partI{}) \right].
\end{align}
The right-hand side is therefore equivalent to the left-hand-side as shown in
\cref{eq:final-lhs-a}, proving the theorem.

\section{Estimating KL divergences in practice} \label{supp:kl-estimates}
In this section we describe how we compute unbiased and low-variance estimates
of $\mathcal{O}_\mathrm{for}$ and $\mathcal{O}_\text{rev}$ in practice. For
both, we use similar techniques to those commonly used in unconditional
hierarchical VAEs. First note that each can be written as the sum of a
likelihood and KL divergence term. For the unconditional VAE objective,
$\mathcal{O}_\mathrm{for}$ and $\mathcal{O}_\text{rev}$ respectively:
\begin{align}
  \EX_{\pdata{}(\I{})} \left[ \text{ELBO}(\theta, \phi, \I{}) \right] &= \EX_{\pdata{}(\I{})}\left[ \EX_{q(\z{}|\I{})} \left[ \log\pmodel{}(\I{}|\z{})  \right] - \kl[\big]{q(\z{}|\I{})}{\pmodel{}(\z{})}  \right], \\
  \mathcal{O}_\mathrm{for}(\theta, \phi, \partphi{}) &= \EX_{\pdata{}(\I{}, \partI{})} \left[ \EX_{q(\z{}|\I{})} \left[ \log \pmodel{}(\tildeI{}|\z{}) \right] - \kl{q(\z{}|\I{})}{\partq{}(\z{}|\partI{})} \right], \\
  \mathcal{O}_\text{rev}(\theta, \phi, \partphi{}) &= \EX_{\pdata{}(\partI{})}\EX_{\partq{}(\z{}|\partI{})} \left[ \log\pmodel{}(\partI{}|\z{}) - \kl{\partq{}(\z{}|\partI{})}{\pmodel{}(\z{})} \right].
\end{align}
In a non-hierarchical VAE with Gaussian $\pmodel{}(\z{})$ and $q(\z{}|\I{})$, it is
common to compute the KL divergence term analytically in order to reduce the
variance of the estimate. In hierarchical VAEs where both $\pmodel{}(\z{})$ and
$q(\z{}|\I{})$ consists of multiple Gaussian distributions with non-linear
dependencies, it is not possible to compute the KL divergence analytically.
However, we can still make use of analytic estimates of the KL divergence at
each layer. For the unconditional VAE objective, an estimator can be derived as
follows:
\begin{align}
  \kl[\big]{q(\z{}|\I{})}{\pmodel{}(\z{})} &= \EX_{q(\z{}|\I{})} \left[ \log \frac{q(\z{}|\I{})}{\pmodel{}(\z{})} \right] \\
                                     &= \EX_{q(\z{}|\I{})} \left[ \log \frac{\prod_{l=1}^L q(\z{}_l|\z{}_{<l}, \I{})}{\prod_{l=1}^L \pmodel{}(\z{}_l|\z{}_{<l})} \right] \\
                                     &= \EX_{q(\z{}|\I{})} \left[ \sum_{l=1}^L \log \frac{q(\z{}_l|\z{}_{<l}, \I{})}{\pmodel{}(\z{}_l|\z{}_{<l})} \right] \\
                                     &= \EX_{q(\z{}|\I{})} \left[ \sum_{l=1}^L \log \frac{q(\z{}_l|\z{}_{<l}, \I{})}{\pmodel{}(\z{}_l|\z{}_{<l})} \right] \\
                                     &= \sum_{l=1}^L \EX_{q(\z{}_{\leq l}|\I{})} \left[ \log \frac{q(\z{}_l|\z{}_{<l}, \I{})}{\pmodel{}(\z{}_l|\z{}_{<l})} \right] \\
                                     &= \sum_{l=1}^L \EX_{q(\z{}_{<l}|\I{})} \left[ \kl[\big]{q(\z{}_l|\z{}_{<l}, \I{})}{\pmodel{}(\z{}_l|\z{}_{<l})} \right].
\end{align}
This is used by \citet{vahdat2020nvae,child2020very} to estimate the KL
divergence term for an unconditional hierarchical VAE, by sampling $\z{}\sim
q(\cdot|\I{})$ and then computing the resulting KL divergence for each layer $l$
conditioned on $\z{}_{<l}$. Similar derivations suggest the following estimates for
the KL divergences in $\mathcal{O}_\mathrm{for}$ and $\mathcal{O}_\text{rev}$:
\begin{align}
  \kl[\big]{q(\z{}|\I{})}{\partq{}(\z{}|\partI{})} &= \sum_{l=1}^L \EX_{q(\z{}_{<l}|\I{})} \left[ \kl[\big]{q(\z{}_l|\z{}_{<l}, \I{})}{\partq{}(\z{}_l|\z{}_{<l}, \partI{})} \right], \\
  \kl[\big]{\partq{}(\z{}|\partI{})}{\pmodel{}(\z{})} &= \sum_{l=1}^L \EX_{\partq{}(\z{}_{<l}|\partI{})} \left[ \kl[\big]{\partq{}(\z{}_l|\z{}_{<l}, \partI{})}{\pmodel{}(\z{}_l|\z{}_{<l})} \right].
\end{align}
We compute unbiased estimates of the KL divergence terms by simply sampling $\z{}$
(from $q(\cdot|\I{})$ for $\mathcal{O}_\mathrm{for}$ or
$\partq{}(\cdot|\partI{})$ for $\mathcal{O}_\text{rev}$) and analytically
computing the resulting KL divergence for each layer $l$ conditioned on
$\z{}_{<l}$.

\section{Experimental details} \label{supp:exp-details}

\begin{table}[t]
  \caption{Summary of hyperparameters used for training IPA and our baselines. Reported batch sizes are summed over all GPUs.}
  \label{tab:exp-details}
  \small
  \centering
  \begin{tabular}{p{2cm}p{2.8cm}rrrrrr}
      \toprule
      Dataset & Method        & \makecell[c]{Learning \\ rate} & \makecell[c]{Batch \\ size} & \makecell[c]{Iterations} & \makecell[c]{Trainable \\ parameters} & \makecell[c]{GPUs} & \makecell[c]{Training time \\ (GPU-hours)} \\
      \midrule
      \multirow{9}{*}{CIFAR-10}     & ANP         & $5 \times 10^{-5}$    & 16    & 700k      & 11.5m & V100                            & 33 \\
                                    & CE          & $2 \times 10^{-4}$    & 64    & 352k      & 34m   & 2080 Ti                         & 14 \\
                                    & RFR         & $1 \times 10^{-4}$    & 8     & 675k      & 31m   & V100                            & 450 \\
                                    & PIC         & $1 \times 10^{-5}$    & 20    & 1800k      & 9m    & V100                            & 430 \\
                                    & CoModGAN    & $2 \times 10^{-3}$    & 32    & 781k       & 82m   & {\scriptsize ($2 \times$)} V100 & 268 \\
                                    & IPA-R       & $2 \times 10^{-5}$    & 16    & 250k      & 18m   & 2080 Ti                         & 83 \\
                                    & IPA from scratch  & $2 \times 10^{-4}$  & 9    & 3050k & 63m   & 2080 Ti   & 542    \\
                                    & IPA from ImageNet & $1.5 \times 10^{{-4}}$ & 16 & 700k & 18m   & 2080 Ti   & 144    \\
                                    & IPA         & $2 \times 10^{-4}$    & 30    & 550k      & 18m   & V100                            & 115 \\
      \midrule
      \multirow{5}{*}{ImageNet-64}  & PIC         & $1 \times 10^{-5}$    & 20    & 2100k      & 9m    & V100                            & 430 \\
                                    & CoModGAN    & $2 \times 10^{-3}$    & 32    & 781k       & 99m   & {\scriptsize ($4 \times$)} V100 & 552 \\
                                    & IPA-R       & $5 \times 10^{-4}$    & 1     & 810k      & 58m   & 2080 Ti                         & 121 \\
                                    & IPA from scratch & $5 \times 10^{-5}$ & 1   & 1930k     & 183m  & 2080 Ti                         & 226 \\
                                    & IPA         & $5 \times 10^{-5}$    & 4     & 700k      & 58m   & 2080 Ti                         & 120 \\
      \midrule
      \multirow{7}{*}{FFHQ-256}     & ANP         & $1 \times 10^{-4}$    & 16    & 990k      & 11.5m & V100                            & 43 \\
                                    & CE          & $2 \times 10^{-4}$    & 64    & 492k      & 71.5m & 2080 Ti                         & 46 \\
                                    & RFR         & $1 \times 10^{-4}$    & 8     & 670k      & 31m   & V100                            & 450 \\
                                    & PIC         & $1 \times 10^{-5}$    & 20    & 1900k      & 9m    & V100                            & 430 \\
                                    & CoModGAN    & $2 \times 10^{-3}$    & 32    & 781k       & 108m  & {\scriptsize ($4 \times$)} V100 & 1332 \\
                                    & IPA-R       & $5 \times 10^{-5}$    & 8     & 416k      & 65m   & {\scriptsize ($4 \times$)} V100 & 800 \\
                                    & IPA         & $1.5 \times 10^{-4}$  & 12    & 626k      & 65m   & {\scriptsize ($4 \times$)} V100 & 1132 \\
      \midrule
      \multirow{2}{*}{Edges2Bags}   & IPA from scratch  & $5\times10^{-5}$ & 1   & 760k   & 183m  & 2080 Ti & 175 \\
                                    & IPA from ImageNet & $2\times10^{-4}$   & 4 & 260k   & 58m  & 2080 Ti  & 45 \\
      \midrule
      \multirow{2}{*}{Edges2Shoes}  & IPA from scratch  & $5\times10^{-5}$ & 1   & 750k   & 183m  & 2080 Ti & 174 \\
                                    & IPA from ImageNet & $2\times10^{-4}$ & 4   & 250k   & 58m  & 2080 Ti  & 43 \\
      \midrule
      \multirow{2}{*}{Chest X-ray}  & CoModGAN          & $2\times10^{-3}$ & 32 & 265k & 108m  & {\scriptsize ($4 \times$)} 2080 Ti & 648 \\  % run id 200znw0h
                                    & IPA               & $1.5\times10^{-4}$ & 8  & 180k & 65m & {\scriptsize ($4 \times$)} V100 & 240 \\  % run id oj53943d
      \hline
  \end{tabular}
  \end{table}

\subsection{IPA and IPA-R including from-scratch/from-ImageNet variants}
We release code for training IPA and IPA-R, code for using the trained artifacts
to perform Bayesian experimental design and out-of-distribution detection, and
various pretrained models\footnote{\url{https://github.com/plai-group/ipa}}.

\paragraph{Architectures}
The encoder and decoder architectures we used for CIFAR-10, ImageNet-64, and
FFHQ-256 were the same as those used by \citet{child2020very} for the same
datasets, with 45, 75, and 62 groups of latent variables respectively. The
encoder and decoder had 39 million parameters for CIFAR-10, 125 million for
ImageNet-64, and 115 million for FFHQ-256. We used partial encoders with
structure identical to the encoders, other than an additional input channel to
accept the concatenated mask. The partial encoders contained 18 million, 58
million, and 65 million parameters respectively for CIFAR-10, ImageNet64, and
FFHQ-256. The architecture used for the edges-to-photos experiments was the same
as that used for ImageNet-64. The architecture used for Chest X-ray 14 was
identical to that used for FFHQ-256.

\paragraph{Training details}
Most training hyperparameters were the same as those used by
\citet{child2020very} for unconditional training of the corresponding
architectures. We report the significant differences in \cref{tab:exp-details}
and the following paragraph. Learning rates were selected with sweeps over three
values, and the batch sizes selected were the largest compatible with the GPU's
memory. We use the Weights \& Biases~\citep{wandb} experiment-tracking
infrastructure. The only unconditional VAEs which we trained ourselves were for
ImageNet-32 (used for the ``IPA from ImageNet'' runs on CIFAR-10) and the Chest
X-ray dataset. We trained the ImageNet-32 VAE on a GeForce RTX 2080 Ti for 14
days, using a batch size of 15 and learning rate of $2\times10^{-4}$ (chosen
with a sweep over 2 values). We trained the Chest X-ray VAE on 4 V100 GPUs for
about 5 days, using a batch size of 8, learning rate of $1.5\times10^{{-4}}$
(chosen with a sweep over 3 values) and ``skip
threshold\footnote{\citet{child2020very} improve training stability by
  skipping updates with gradients larger than a set threshold. Using a larger
  threshold, and so skipping fewer updates, was found to be necessary to train
  their unconditional VAE on the Chest X-ray dataset. In all other cases, our
  ``skip threshold'' is the same as that used by \citet{child2020very} for the
  relevant architecture}'' of 15000. While we could have trained a conditional VAE
from scratch for the x-ray experiments, training an unconditional model first
allowed us to speed up later experimentation with IPA.

\subsection{Other baselines}
For all the remaining baselines, we based our implementations on publicly
available (official or unofficial) implementations. A link is provided for each.
All the training procedures were modified to use the same distribution of
partially observed images as for training IPA (see \cref{sec:experiments}).

\paragraph{CoModGAN} We used the official implementation of
\citet{zhao2021large}\footnote{\url{https://github.com/zsyzzsoft/co-mod-gan}}.
See \cref{tab:exp-details} for our hyperparameters, based on those used by
\citet{zhao2021large}.
% For the ImageNet-64 experiment, we used a batch size of 32.
% All other hyperparameters were chosen according to CoModGAN's reported
% hyperparameters. The networks for CIFAR-10, ImageNet-64 and FFHQ-256 had
% 82,097,684, 98,881,049, and 108,041,635 parameters respectively. They were
% trained for 25 million iterations. Training took 134 hours on 2 GPUs for
% CIFAR-10, 138 hours on 4 GPUs for Imagenet-64, and 333 hours on 4 GPUs for
% FFHQ-256. All the GPUs used were Tesla V100s.

\paragraph{PIC} We adapted the official implementation of
\citet{zheng2019pluralistic}\footnote{\url{https://github.com/lyndonzheng/Pluralistic-Inpainting}},
and used their reported hyperparameters where possible (see
\cref{tab:exp-details}). The PIC architecture is designed for $256 \times 256$
images. Therefore, in order to test it on ImageNet-64 and CIFAR-10, we resized
these images to $256 \times 256$ (via bilinear interpolation) before feeding
them into the PIC model. We then down-sampled the inpainted images back to the
original size of $32 \times 32$ for evaluation.

% We used the same architecture, hyperparameters and objective functions as their
% implementation. The networks for all CIFAR-10, FFHQ-256 and ImageNet-64 had
% 9,128,390 parameters. They were trained using Adam
% optimiser~\citep{kingma2015adam} with a batch size of 20, a learning rate of
% $10^{-5}$ and $\beta_1, \beta_2$ parameters of 0 and 0.999 respectively. The
% models were trained for 600, 800 and 34 epochs (corresponding to 9M, 1.8M, and
% 2.2M iterations) respectively for FFHQ-256, CIFAR-10 and ImageNet-64. It
% approximately took 430 hours to train each of these models on a Tesla V100 GPU.

\paragraph{ANP} Our ANP network architecture was based on that of
\citet{kim2019attentive}\footnote{Our implementation of ANP is based on
  \url{https://github.com/EmilienDupont/neural-processes} and
  \url{https://github.com/YannDubs/Neural-Process-Family}}, differing in that we
used hidden and latent dimensions of 512. In the original image inpainting
experiments of \citep{kim2019attentive} the images are $32 \times 32$. Since
ANPs embed each pixel separately and their self-attention and cross-attention
layers attend to all other pixels in the observed and target sets, it is
computationally expensive to scale them to larger images. We therefore
downsampled the FFHQ-256 images to $64 \times 64$, and upsampled the inpainted
image back to $256 \times 256$ via bilinear interpolation. Additionally, at
training time, we randomly dropped half of the observed pixels (a.k.a. context
set in neural process literature) to reduce the computational cost. Finally, the
target set at training time was half of the unobserved pixels. For CIFAR-10, on
the other hand, no modfication was done to the image resolution or observation
masks i.e.~images were fed in at $32 \times 32$ resolution, the context sets
were the set of all observed pixels in masked images $\partI$, and the target
sets consisted of all unobserved pixels.

\paragraph{CE} We use the architecture reported by
\citet{pathak2016context}\footnote{Our implementation of CE is based on
  \url{https://github.com/BoyuanJiang/context_encoder_pytorch}, with some
  modifications according to \url{https://github.com/pathak22/context-encoder}
  to support larger image sizes and non-centered observation masks.}.

\paragraph{RFR} We adapted the official implementation of
\citet{li2020recurrent}\footnote{\url{https://github.com/jingyuanli001/RFR-Inpainting}}.
Similar to PIC, all the input images to this model were resized to
$256 \times 256$. The iterative way of completing images in this baseline makes
it slow to complete images with most of their pixels missing. Therefore, after
around 450 hours of training on a Tesla V100 GPU, they were trained for only 120
and 85 epochs respectively for CIFAR-10 and FFHQ-256. This may partly explain
the poor performance of RFR that we report.

\paragraph{VQ-VAE} We adapted the official implementation of
\citet{peng2021generating}\footnote{https://github.com/USTC-JialunPeng/Diverse-Structure-Inpainting}.
Similar to PIC and RFR, all the input images for this model were resized to
$256 \times 256$. The model has three modules: a VAE, a structure generator and
a texture generator. Each is trained separately:
\begin{itemize}
\item The VAEs for both datasets were trained for 1 million iterations on an RTX
  2080 Ti GPU. It took less than 13 hours for each dataset.
  \item The structure generator for CIFAR-10 was trained for 1 million
        iterations and the FFHQ-256 one was trained for 1.5 million iterations.
        It took 128 hours on a Tesla V100 and 212 hours on an RTX 2080 Ti for
        CIFAR-10 and FFHQ-256 respectively.
  \item The texture generators were trained for 2 million traces for both
        datasets. It took 171 hours on a Tesla V100 and 211 hours on an RTX 2080
        Ti GPU, for CIFAR-10 and FFHQ-256 respectively.
\end{itemize}
This model is very slow to complete images at test time. The authors reported
that it takes 45 seconds to complete a $256 \times 256$ image, and in our
experiments we observed that this time could be up to 3 minutes on an RTX 2080
Ti GPU. This makes evaluating the quantitative metrics we use prohibitively
expensive. We do, however, report qualitative results in
\cref{supp:image-samples}.

\subsection{Metrics} \label{supp:metrics}
\paragraph{FID for completed images}
We use the FID score~\citep{heusel2017gans} to quantify the distance between
$\pdata{}(\I{})$ and $\int \pcomp{}(\I{}|\partI{}) \pdata{}(\partI{})
\mathrm{d}\partI{}$, the distribution resulting from sampling a dataset image,
masking out some pixels, and replacing them by performing inpainting. This
should be zero if $\pcomp{}(\I{}|\partI{}) = \pdata{}(\I{}|\partI{})$. Since
this metric does not explicitly consider multiple completions of the same
observation, we view it as a measure of image quality more than diversity.
% \Cref{fig:metrics} shows FID scores for completions by each method with various
% ratios of observed pixels, and aggregated scores are reported in
% \cref{tab:results-completion}. IPA achieves lower (better) FID scores than most
% baselines but is in turn outperformed by CoModGAN.

\paragraph{P-IDS}
This metric was proposed by \citet{zhao2021large}, who show that is correlates
well with human evaluations. It measures the linear separability of dataset and
inpainted images in a pretrained Inception network's feature space. Higher
values are better, indicating that the data is less separable.
%
% Our evaluations with this metric are mostly consistent with FID scores, except
% on CIFAR-10 with over about 50\% of pixels observed, for which IPA outperforms
% all the baselines.

\paragraph{LPIPS diversity score}
This metric for diversity was proposed by \citet{zhu2017toward}, involving
calculating the LPIPS distance between pairs of images sampled from
$\pcomp{}(\cdot|\partI{})$ for the same $\partI{}$. We report it here for
completeness but it is not a good measure of the ``fit'' between
$\pcomp{}(\I{}|\partI{})$ and $\pdata{}(\I{}|\partI{})$ since it can always be
maximised by making $\pcomp{}(\I{}|\partI{})$ as diverse as possible, without
reference to the ``true'' posterior $\pdata{}(\I{}|\partI{})$.

\paragraph{Faithfulness weighted variance}
The faithfulness weighted variance (FWV) was proposed by
\citet{li2020multimodal}, measuring both how diverse generated images
$\I{}\sim\pcomp{}(\cdot|\partI{})$ are and how well they fit the ground truth.
To compute it we first draw $N$ pairs of $(\I{}_i, \partI{}_i)$ from the test
set. Then for each $\partI{}_i$ we draw $K$ conditional samples from the model
$\hat{\I{}}_{i,j} \sim \pcomp(\cdot | \partI{}_i)$ and let $\bar{\I{}}_i$ be the
pixel-wise average of $\{\hat{\I{}}_{i,j}\}_{j=1}^K$. Finally,
\begin{equation}
  \text{FWV} = \sum_{i=1}^N \sum_{j=1}^K w_{i,j} d_{\text{LPIPS}}(\hat{\I{}}_{i,j}, \bar{\I{}}_i),
  \text{ where }
  w_{i,j} = \exp \left(- \frac{d_{\text{LPIPS}}(\hat{\I{}}_{i,j}, \I{}_i)}{2\sigma^2} \right).
\end{equation}
We normalise by $N$ and report $\frac{\text{FWV}}{N}$. We use $N = K = 100$ and
compute the FWV for various values of $\sigma$. The parameter $\sigma$
determines how closely a sample must match the ground truth to be able to
contribute to the FWV. \textcolor{black}{Using a low $\sigma$ therefore favours
  methods which can most faithfully reconstruct the ground truth, and using a
  high $\sigma$ favours methods which produce diverse samples, regardless of
  their match to the ground truth. We compute this metric for inpainting tasks,
  in which generated images are typically much more diverse than in the
  super-resolution task it was designed for~\citep{li2020multimodal}. We
  therefore report scores orders of magnitude higher than
  \citet{li2020multimodal} and note that it is unclear how meaningful the
  pixel-space average is in our setting. }

% \paragraph{LPIPS-GT}
% \Cref{fig:metrics} and \cref{tab:results-completion} show that IPA outperforms
% all baselines on this metric. We believe this reflects the tendency of GAN-based
% baselines to miss modes of $\pdata{}(\I{}|\partI{})$, while IPA more faithfully
% represents uncertainty. Our qualitative results in the appendix lend evidence to
% this theory by showing that, for some $\partI{}$, all samples from CoModGAN are
% unexpectedly semantically similar.

\section{Additional results} \label{supp:additional-results}

\begin{figure}
  \includegraphics[width=\textwidth]{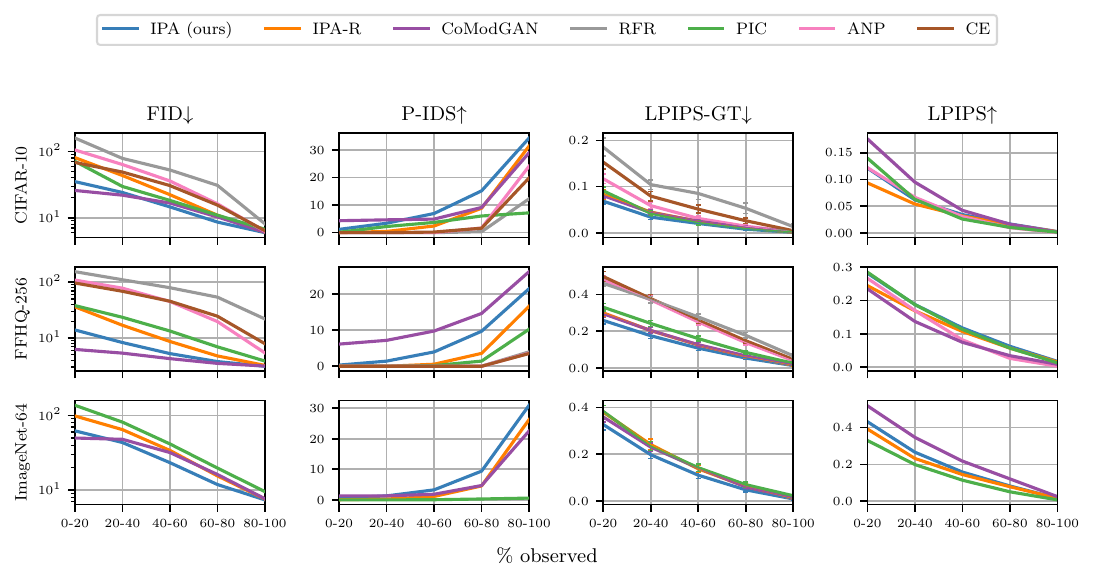}
  \caption{Expanded \cref{fig:metrics} including results on ImageNet-64 and the
    LPIPS diversity score.}
  \label{fig:extra-metrics}
\end{figure}

\begin{figure}
  \includegraphics[scale=1]{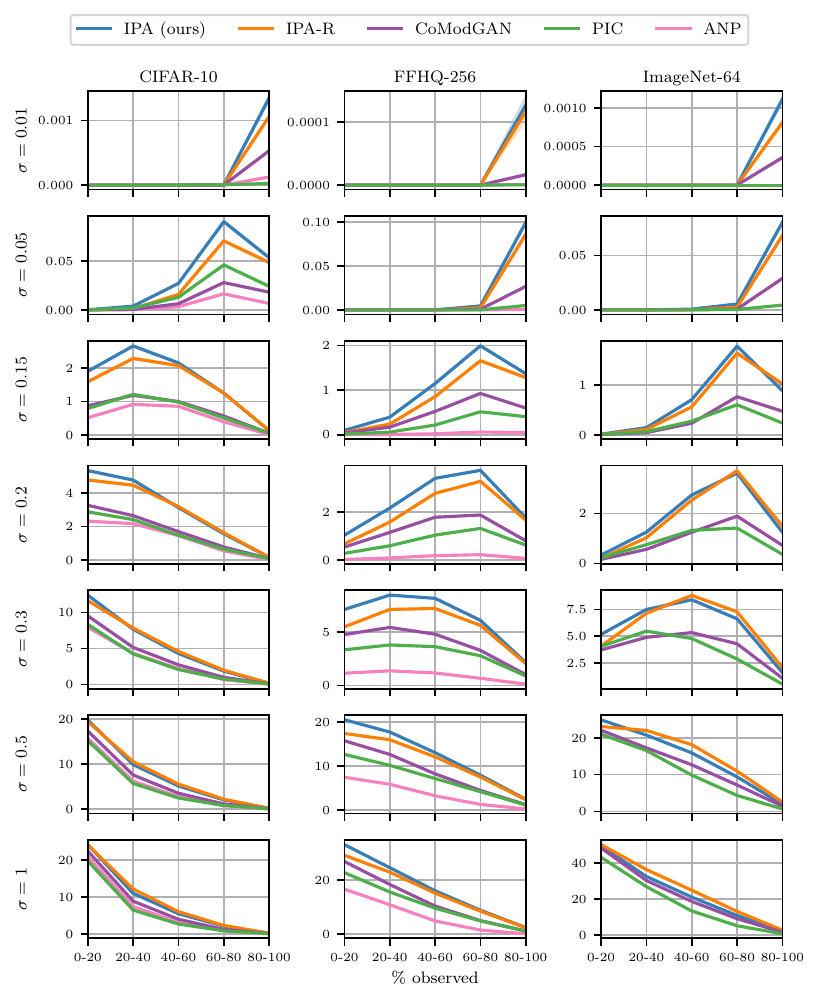}
  \caption{The faithfulness weighted variance for inpainting on CIFAR-10,
    FFHQ-256, and ImageNet-64, for various values of $\sigma$. IPA obtains the
    best performance on almost all datasets and values of $\sigma$, only being
    outperformed by IPA-R on CIFAR-10 with high values of $\sigma$. This
    indicates that IPA both assigns high probability density to the ground truth
    (and so performs well for small $\sigma$) and generates diverse samples (and
    so performs well for larger $\sigma$).}
  \label{fig:fwv}
\end{figure}

\Cref{fig:extra-metrics} shows a breakdown of various metric for different sizes
of mask. This is similar to \cref{fig:metrics} in the main paper, but also shows
the LPIPS diversity score and scores on ImageNet-64. In \cref{fig:fwv}, we plot
the faithfulness weighted variance described in \cref{supp:metrics}. We compute
it for the values of $\sigma$ reported by \citet{li2020multimodal}, as well as
two higher values and two lower values.

\section{Bayesian optimal experimental design} \label{sec:supp-boed}

\subsection{EIG estimators}  \label{sec:eig-estimator}

In this section, we expand on how our estimator for the expected information
gain differs from prior work~\citep{harvey2019near}. Repeating some relevant
definitions from the main text, $f_{\coord{}_1,\ldots,\coord{}_t}$ is a function which maps
from a full image to a sequence of observations of the image at locations
$\coord{}_1,\ldots,\coord{}_t$. When estimating the EIG at time step $t$, we will have already
observed a sequence of observations denoted $\partI{}_{\coord{}_1,\ldots,\coord{}_{t-1}}$,
extracted from some latent image $\I{}$ as $f_{\coord{}_1,\ldots,\coord{}_{t-1}}(\I{})$. We
use a CNN which outputs $g(v|f_{\coord{}_1,\ldots,\coord{}_t}(\I{}))$, an approximation of the
posterior over the latent variable of interest given a sequence of observations.
Following \citet{harvey2019near}, we will from now on refer to this as the
AVP-CNN (attentional variational posterior CNN).
Both our method and that of \citet{harvey2019near} sample image completions,
which we denote $\I{}^{(1)},\ldots,\I{}^{(N)}$, conditioned on
$\partI{}_{\coord{}_1,\ldots,\coord{}_{t-1}}$. \citet{harvey2019near} do so by retrieving
images which roughly match the observed pixel values from a database. Although
completions from this are diverse, they can match the observed values poorly. We
therefore replace this stage with an image completion network trained using
IPA.

Given these components, \citet{harvey2019near} approximate the expected
information gain with
\begin{align}
  \label{eq:old-eig}
  \text{EIG}(\coord{}_t; \partI{}_{\coord{}_1,\ldots,\coord{}_{t-1}}) &\approx \overbrace{\mathcal{H} \left[ g(\cdot|\partI{}_{\coord{}_{1},\ldots,\coord{}_{t-1}}) \right]}^{\text{entropy after $t-1$ scans}} - \overbrace{\frac{1}{N} \sum_{n=1}^N  \mathcal{H} \left[ g(\cdot|f_{\coord{}_1,\ldots,\coord{}_t}(\I{}^{(n)})) \right]}^{\text{expected entropy after $t$ scans}}.
\end{align}
Since the prior entropy term of \cref{eq:old-eig} does not depend on $\coord{}_t$, it
can be neglected when choosing $\coord{}_t$ with BOED. As only the expected posterior
entropy then needs to be estimated and compared for various $\coord{}_t$, we will from
now on refer to this estimator with the acronym EPE (for `expected posterior
entropy').

\begin{figure*}[t]
  \centering
  \includegraphics[width=\textwidth]{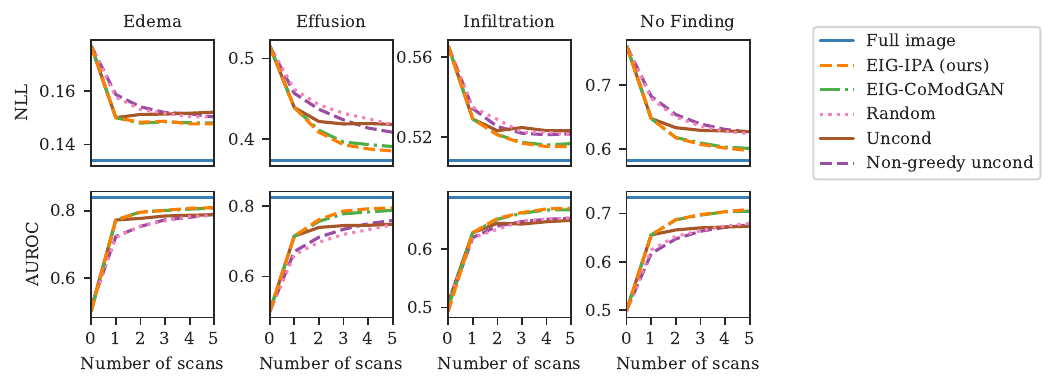}
  \caption{We extend on \cref{fig:boed} by additionally reporting negative
    log-likelihoods (top row) for the classification tasks. We also include
    results from the `EPE' estimator used by \citet{harvey2019near}. We see that
    our BOED estimator provides significant improvements; resulting performance
    is always superior to the other baselines. This is only sometimes the case
    for the `EPE' estimator. }
  \label{fig:boed-auroc-supp}
\end{figure*}

We use a different estimator for the prior entropy. It becomes equivalent as $N
\rightarrow \infty$ if $g$ and the image completion method are perfect but, when
this is not the case, produces an estimate of the prior entropy with a
dependence on $\coord{}_t$. To repeat \cref{eq:new-eig}, this results in the following
estimator for the EIG:
\begin{align}
  \label{eq:new-eig-supp}
  \text{EIG}(\coord{}_t;& \partI{}_{\coord{}_1,\ldots,\coord{}_{t-1}}) \approx \overbrace{\mathcal{H} \left[ \frac{1}{N} \sum_{n=1}^N g(\cdot|f_{\coord{}_1,\ldots,\coord{}_t}(\I{}^{(n)})) \right]}^{\text{entropy after $t-1$ scans}} - \overbrace{\frac{1}{N} \sum_{n=1}^N  \mathcal{H} \left[ g(\cdot|f_{\coord{}_1,\ldots,\coord{}_t}(\I{}^{(n)})) \right]}^{\text{expected entropy after $t$ scans}}.
\end{align}
While it is not immediately clear that this estimator is better than that in
\cref{eq:old-eig}, one useful property is that, like the true expected
information gain, it is guaranteed to be non-negative. This can be seen as
follows. First, note that the prior entropy term is the entropy of an
expectation (over $n \sim \text{Uniform}(1, N)$), and the expected posterior
entropy term swaps the order of the entropy and expectation. Since the mapping
from a distribution $g(\cdot)$ to its entropy $\mathcal{H}[g(\cdot)]$ is
strictly concave, Jensen's inequality is applicable. Invoking Jensen's
inequality shows that the prior entropy term must be greater than or equal to
the expected posterior entropy, and so the estimate of the EIG will be
non-negative. In \cref{fig:boed-auroc-supp}, we compare the performance of this
estimator (denoted EIG) and that in \cref{eq:old-eig} (EPE). We find that the
estimator denoted EIG leads to significantly better classification performance.

Another view of our EIG estimator is as a variant of the inception
score~\citep{salimans2016improved} for the sampled images
$\I{}^{(1)},\ldots,\I{}^{(N)}$. The standard inception score is computed using
an image classifier trained on ImageNet~\citep{salimans2016improved}.
\Cref{eq:new-eig} is the inception score if this classifier is replaced with the
AVP-CNN acting on observations of each image at locations $\coord{}_1,\ldots,\coord{}_t$.

\begin{figure}[t]
  \centering
  \includegraphics[scale=1]{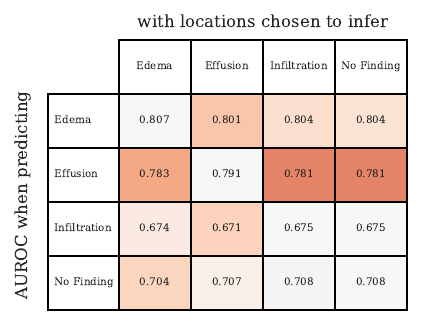}
  \caption{AUROC scores when performing one classification task with scan
    locations chosen for a different classification task. The intensity of the
    colour of each cell is proportional to the difference between its value and
    the greatest value in its row.}
  \label{fig:boed-correlations}
\end{figure}

\subsection{BOED baselines}
\paragraph{Random}
This baseline simply samples the scan location independently at each time step
from a uniform distribution over all valid locations.

\paragraph{Uncond}
This baseline ablates our estimator for the EIG (\cref{eq:new-eig}) by sampling
images $\I{}^{(1)},\ldots,\I{}^{(N)}$ from the training dataset (without
conditioning on $\partI{}_{\coord{}_1,\ldots,\coord{}_{t-1}}$) instead of from IPA
(conditioned on $\partI{}_{\coord{}_1,\ldots,\coord{}_{t-1}}$). That is, we use:
\begin{align}
  \label{eq:uncond}
  \text{U}^{\text{uncond}}(\coord{}_t; \coord{}_1,&\ldots,\coord{}_{t-1}) = \mathcal{H} \left[ \frac{1}{N} \sum_{n=1}^N g(\cdot|f_{\coord{}_1,\ldots,\coord{}_t}(\I{}^{(n)})) \right] - \frac{1}{N} \sum_{n=1}^N  \mathcal{H} \left[ g(\cdot|f_{\coord{}_1,\ldots,\coord{}_t}(\I{}^{(n)})) \right]
\end{align}
with $\I{}^{(1)},\ldots,\I{}^{(N)} \sim p(\I{})$. The resulting choice of scan
location $\coord{}_t$ at each time step $t$ has no dependence on previous
observations $\partI{}_{\coord{}_1,\ldots,\coord{}_{t-1}}$.

\paragraph{Non-greedy uncond}
When selecting scan locations that optimise the EIG, we select the scan location
at each time step greedily to optimise the EIG from the next scan. To do
otherwise (i.e.~maximise the EIG from multiple next scans) is intractable
because of the dependence of the EIG at each $t$ on previous observations
$\partI{}_{\coord{}_1,\ldots,\coord{}_{t-1}}$. However, since the above `Uncond' baseline
selects each $\coord{}_t$ independently of what is observed at previous scans, we can
remove this greedy property. We do so by selecting all $\coord{}_1,\ldots,\coord{}_T$
simultaneously to maximise
\begin{align}
  \label{eq:uncond}
  \text{U}^{\text{non-greedy}}(\coord{}_1,&\ldots,\coord{}_T) = \mathcal{H} \left[ \frac{1}{N} \sum_{n=1}^N g(\cdot|f_{\emptyset}(\I{}^{(n)})) \right] - \frac{1}{N} \sum_{n=1}^N  \mathcal{H} \left[ g(\cdot|f_{\coord{}_1,\ldots,\coord{}_T}(\I{}^{(n)})) \right]
\end{align}
with $\I{}^{(1)},\ldots,\I{}^{(N)} \sim p(\I{})$. Since the number of possible
values of $\coord{}_1,\ldots,\coord{}_T$ increases exponentially with $T$, it is not feasible
to search over all of them. We therefore optimise this sequence by
randomly sampling a set of such sequences, and choosing the best from this set.

\begin{figure*}[p]
  \centering
  \includegraphics[width=0.98\textwidth]{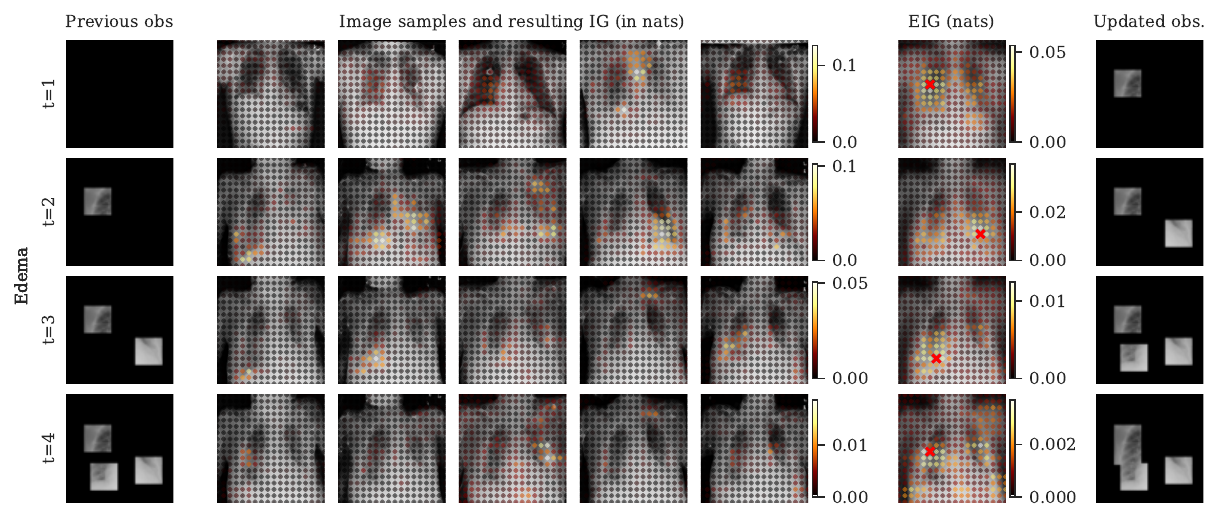}
  \includegraphics[width=0.98\textwidth]{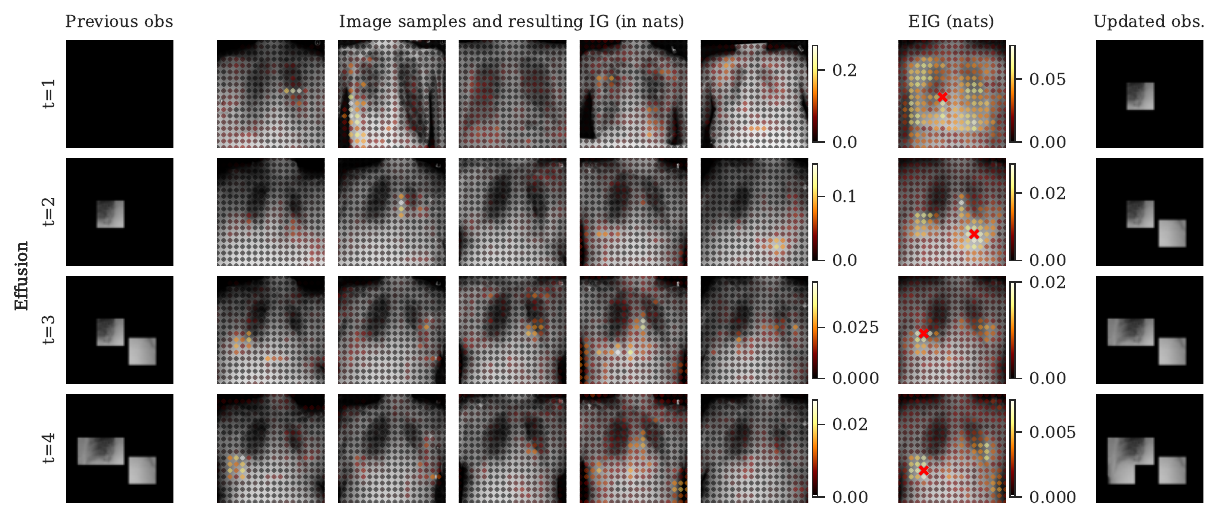}
  \includegraphics[width=\textwidth]{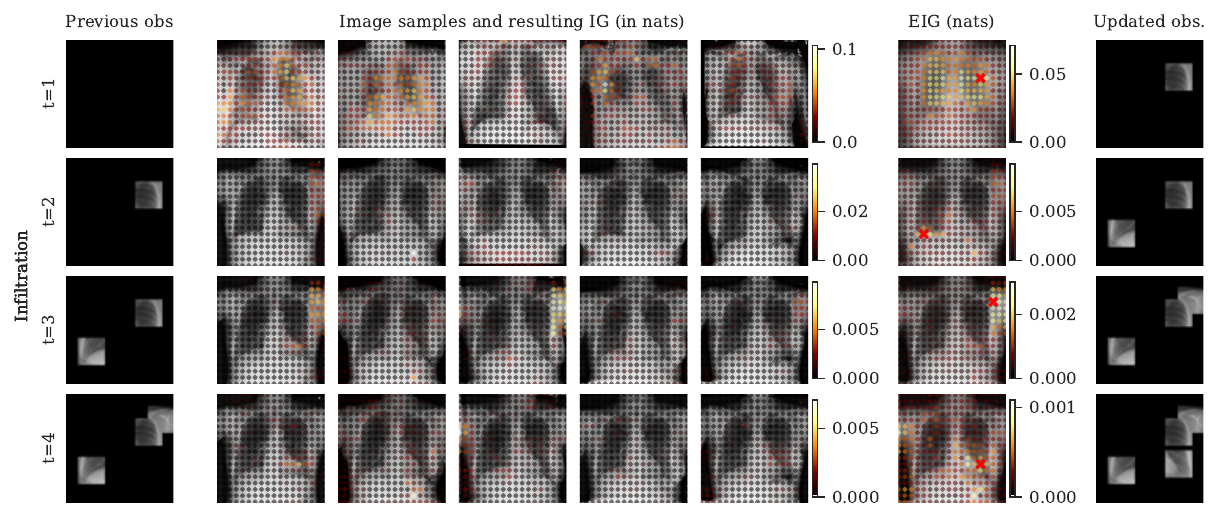}
  \caption{Visualisations of BOED processes. On the top, the task is to detect
    `Edema'. In the middle, we aim to detect `Effusion' in a different patient.
    The task at the bottom is to detect `Infiltration'. Four scan locations are
    selected for each. The left column shows the observations made before
    selecting each scan location. The next five columns show five of the $N=10$
    sampled image completions conditioned on these observations. Each is
    overlaid with the information gain predicted after scanning any location.
    The second column from the left shows the expected information gain at each
    location, given by averaging the information gains arising from each sampled
    image completion. A red cross marks the maximum. The final column shows the
    updated observations after scanning the location which maximises the
    expected information gain.}
  \label{fig:extra-xray-boed}
\end{figure*}

\subsection{Additional BOED results} \label{supp:boed-results}
\Cref{fig:boed-auroc-supp} shows additional results from classification with
sequences chosen by BOED and our baselines. In particular, we report negative
log-likelihoods for the class labels as well as the AUROC scores. We see that
the negative log-likelihood sometimes increases as more scans are taken,
indicating that the classifier used is not perfectly trained. Interestingly,
this only happens when glimpses are chosen with the EPE estimator of
\citet{harvey2019near}. This may be because the EPE estimator has a tendency to
select locations which cause the classifier to be overconfident (and thus
produce artificially low posterior entropies). With the EIG estimator, and the
baselines, this issue is not observed.

To illustrate that the scan locations chosen by BOED are task-dependent,
\cref{fig:boed-correlations} shows the results from using scan locations chosen
for one classification task (i.e.~diagnosing a particular disease) to perform a
different classification task. Each row shows the results for a particular task,
with each column showing results when scan locations are chosen based on a
particular task. As expected, the highest (or at least joint-highest) AUROC
score in each row occurs on the diagonal, when the classification task and
choice of scan locations are aligned. Away from the diagonal, the scores are
only slightly lower, perhaps reflecting that there is a large overlap between
the relevant areas for these tasks.

\Cref{fig:extra-xray-boed} shows additional visualisations of the Bayesian
experimental design process for different test images, adding to the one shown
in the main paper. The `information gains' shown for a particular $\I{}^*$
(sampled conditioned on observations $\partI{}_{\coord{}_1,\ldots,\coord{}_{t-1}}$) are
computed as the following KL divergence:
% EIG but the `expected
% entropy after $t$ scans' term is replaced by the entropy given a particular
% sampled $\I{}^*$. That is,
\begin{align}
  \label{eq:ig}
%  \text{IG}(\coord{}_t; \partI{}_{\coord{}_1,\ldots,\coord{}_{t-1}}) \approx \overbrace{\mathcal{H} \left[ \frac{1}{N} \sum_{n=1}^N g(\cdot|f_{\coord{}_1,\ldots,\coord{}_t}(\I{}^{(n)})) \right]}^{\text{entropy after $t-1$ scans}} - \overbrace{\mathcal{H} \left[ g(\cdot|f_{\coord{}_1,\ldots,\coord{}_t}(\I{}^*)) \right]}^{\text{entropy after $t$ scans given $\I{}^*$}}. \\
                                                   \text{IG}(\I{}^*, \coord{}_t; \partI{}_{\coord{}_1,\ldots,\coord{}_{t-1}}) \approx \kl[\Bigg]{g(\cdot|f_{\coord{}_1,\ldots,\coord{}_t}(\I{}^*))}{\frac{1}{N} \sum_{n=1}^N g(\cdot|f_{\coord{}_1,\ldots,\coord{}_t}(\I{}^{(n)}))}.
\end{align}
Crucially, averaging information gains computed with $\I{}^*=\I{}^{(1)}$ to
$\I{}^{(N)}$ gives the estimate of the EIG presented in \cref{eq:new-eig}.

\subsection{BOED experimental details}
The AVP-CNN, $g$, is trained to map from masked images, $\partI{}$, to
distributions over the class labels. Each image in the Chest X-ray 14 dataset
has labels indicating the presence or absence of each of 14 pathologies. We
train $g$ to produce 15 outputs: an estimated probability of the presence or
absence of each of the 14 conditions individually; and an additional estimate of
the probability that any (one or more) of these conditions is present. We train
$g$ to estimate these using a cross-entropy loss. Masked images are sampled
using almost the same mask distribution as for training the image completion
networks (described in \cref{sec:experiments}); the only difference is that
patches now have 25\% rather than 35\% of the image width, to match the
observations we use in the experiments with BOED. We use an AVP-CNN pretrained
on ImageNet and then trained on Chest X-ray 14 for 32\,000 iterations with a
batch size of 32 and learning rate $1\times10^{-5}$. We train IPA as described
in \cref{supp:exp-details}.

We select each scan location by evaluating \cref{eq:new-eig} at every point in
an evenly-spaced $17\times17$ grid over the image, and choosing the maximum. We
evaluate the EIG with $N=10$ sampled image completions. This is repeated to
select the scan location for each $t=1,\ldots,5$ (with the sampled images
conditioned on observations up to $t-1$ at each stage). Where applicable, we use
the same trained networks and hyperparameters for the baselines. For the
`non-greedy uncond' baseline, we select a scan location sequence by choosing the
best of 289 sampled sequences. This number was chosen to match the number of
locations in the $17\times17$ grid that the other methods search over at each
time step. The results reported in \cref{fig:boed,fig:boed-auroc-supp} were
computed on a randomly-sampled 5000 image subset of the Chest X-ray 14 test set.

\section{Out-of-distribution detection} \label{supp:ood} A major advantage of
likelihood-based models is the possibility to use them to detect when an input
is dissimilar to the training data (out-of-distribution, or OOD). Since a
learned model is likely to perform poorly on such inputs, it is important for
deployed systems to be able to recognise them and this has been the focus of a
substantial body of
work~\citep{ren2019likelihood,hendrycks2016baseline,xiao2020likelihood,havtorn2021hierarchical}.
The task we consider is detecting whether a partially observed image (e.g.~the
input to an image completion system) is OOD.

\begin{figure}[t]
  \centering
  \begin{minipage}{0.41\textwidth}

    \centering
    \includegraphics[scale=.8]{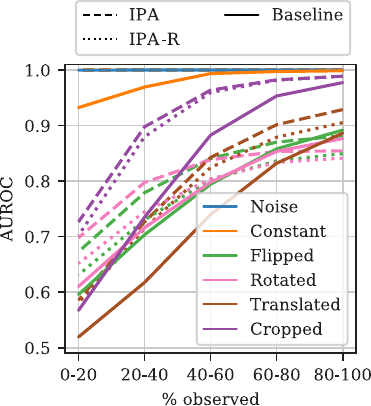}

  \end{minipage}
  \begin{minipage}{.58\textwidth}

%  \caption{AUROC scores for OOD detection. Higher is better.}
  \begin{tabular}{lccccc}
    \toprule
    ID & \multicolumn{5}{c}{CIFAR-10}  \\
    \cmidrule(r){2-6}
    OOD & Noise & Const. & Flip & Rotate & SVHN \\
    \midrule
    IPA      & \textbf{.995} & \textbf{.949} & \underline{.552} & \underline{.594} & \textbf{.753}    \\
    IPA-R    & .987         & \underline{.921}                  & \textbf{.635}    & \textbf{.619}    & \underline{.686} \\
    Baseline  & \underline{.993} & .909 & .543             & .536             & .556             \\
    \bottomrule
  \end{tabular}
  \newline
  \vspace*{.1cm}
  \newline
  \begin{tabular}{lcccccc}
    \toprule
    ID & \multicolumn{5}{c}{FFHQ-256}  \\
    \cmidrule(r){2-6}
    OOD         & Noise & Const. & Flip & Rotate & ~Crop~ \\
    \midrule
    IPA         & \textbf{1.00} & \textbf{1.00} & \textbf{.809}    & \textbf{.808}    & \textbf{.912}    \\
    IPA-R       & \textbf{1.00} & \textbf{1.00} & \underline{.770} & \underline{.775} & \underline{.902} \\
    Baseline     & \textbf{1.00} & 0.98  & .769 & .771             & .823             \\
    \bottomrule
  \end{tabular}

\end{minipage}%
\caption{OOD-detection on incomplete images. On the left, we show AUROC scores
  on FFHQ-256 computed separately for masks from varying distributions. On the
  right we report, for both CIFAR-10 and FFHQ-256, the average of these scores
  over all mask distributions.}
\label{fig:ood-results}
% \vspace{-.5cm}
\end{figure}

OOD-detection using probabilistic models is theoretically appealing and robust,
since no assumptions need to be made about the OOD
data~\citep{xiao2020likelihood,havtorn2021hierarchical}. Such OOD-detection
metrics commonly have the following interface: they take as input a
probabilistic model (e.g.~a VAE) and a data point, and output a scalar value.
The greater the value of this scalar, the more likely it is that the data point
is within the distribution of the generative model. One intuitive metric is
simply the log-likelihood of the data point under the model. This can work
poorly in practice, however, and it has been found that learned models sometimes
assign higher likelihood to out-of-distribution data than to in-distribution
data~\citep{nalisnick2018deep}. Several alternative OOD-detection metrics have
been proposed specifically for
VAEs~\citep{xiao2020likelihood,havtorn2021hierarchical}. To use them, we note
that we can use networks trained with either IPA or IPA-R to lower-bound the
likelihood of a partially-observed image. Specifically, this lower-bound is the
reverse KL objective in \cref{eq:reverse-obj}. After conducting preliminary
experiments using the VD-VAE with several OOD-detection
metrics~\citep{xiao2020likelihood,havtorn2021hierarchical}, we found that
\begin{wrapfigure}[18]{R}{.4\textwidth}
  \vspace{-.4cm}
  \centering
  \includegraphics[scale=.8]{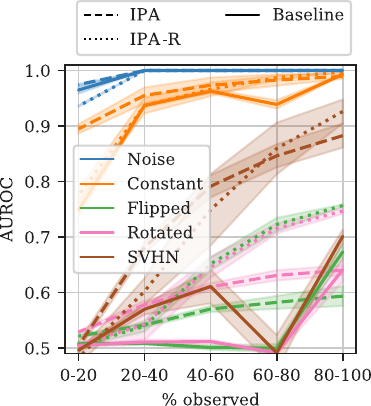}
  \caption{Breakdown of AUROC for each mask distribution on CIFAR-10. }
  \label{fig:ood-supplementary}
\end{wrapfigure}
improved performance could be obtained (for OOD-detection of both partially- and
fully-observed images) using the ``temperature gradient'' metric described
further in \cref{supp:temp-grad} below.
It is efficient to compute, requiring the VAE to be run just once and gradients
computed.

As a baseline, we complete the partially observed image with our best performing
baseline image completion method, CoModGAN, and perform OOD-detection on the
result (using the ``temperature gradient'' metric on an unconditional VD-VAE).
\Cref{fig:ood-results} shows our results on CIFAR-10 and FFHQ-256. We test
various types of OOD data: \textbf{Noise} samples each pixel value independently
from a uniform distribution; \textbf{Constant} uniformly samples a single colour
for the entire image; \textbf{Flip} vertically flips an in-distribution (ID)
image; \textbf{Rotate} rotates an ID image by 90$^{\circ}$; \textbf{Crop}
upsamples an ID image by a factor of 1.5 and then crops to the original size.
\textbf{SVHN} is the Street View House Number dataset~\citep{netzer2011reading}.
With 80-100\% of the image observed, we obtain better AUROCs than similar
methods applied to the full image. In particular, on CIFAR-10 vs SVHN, we
outperform both \citet{xiao2020likelihood} and \citet{havtorn2021hierarchical},
albeit with a larger VAE architecture. Using IPA for OOD-detection results in
performance that degrades gracefully as less of the image is observed. There are
therefore particularly large benefits over the baseline when only 0-20\% or
20-40\% of the image is observed.

\Cref{fig:ood-supplementary} shows a breakdown of the AUROC scores for each mask
distribution on CIFAR-10, similar to that shown for FFHQ-256 in
\cref{fig:ood-results}. Error bars are standard deviations computed with 3 runs,
with the unconditional VD-VAE and IPA/IPA-R retrained each time.

\subsection{Out-of-distribution detection with temperature gradients} \label{supp:temp-grad}

We conducted preliminary experiments with several techniques for OOD-detection
with VAEs, applying each to the VD-VAE architecture. Using the likelihood-regret
technique~\citep{xiao2020likelihood} with the VD-VAE is computationally costly
(as it involves optimising neural network parameters), and we obtained results
on CIFAR-10 considerably worse than those produced by the smaller VAE
architecture with which it was introduced. We also implemented the
log-likelihood ratio metric \citep{havtorn2021hierarchical} for the VD-VAE, but
did not find any hyperparameter configurations with which it was consistently
better than random guessing across different tasks.

We therefore introduce a metric which we found to work well with the VD-VAE. It
is based on a common heuristic to make a VAE produce more regular samples:
reducing the variance of the prior over its latent
variables~\citep{vahdat2020nvae}. If $\mathbf{t}$ is a vector containing the
temperature $\mathbf{t}_l$ for each layer $l$, then doing so corresponds to
modifying the prior $\pmodel{}(\z{})$ so that the Gaussian distribution over each
group of latent variables is scaled by the corresponding temperature:
\begin{equation}
  \label{eq:hierarchical-prior}
  \pmodel{}(\z{}; \mathbf{t}) = \prod_{l=1}^L  \frac{1}{C(\mathbf{t}_l, \z{}_{<l})} \pmodel{}(\z{}_l|\z{}_{<l})^{\frac{1}{\mathbf{t}^2_l}}
\end{equation}
where $C(\mathbf{t}_l, \z{}_{<l})$ is a normalisation constant. In practice, since
$\pmodel{}(\z{}_l|\z{}_{<l})$ is a Gaussian distribution, this scaling corresponds to
simply multiplying the standard deviation by $\mathbf{t}_l$.

We can lower-bound the corresponding marginal likelihood $\pmodel{}(\partI{};
\mathbf{t})$ using the partial encoder similarly to the training objective in
\cref{eq:reverse-obj}:
\begin{align}
  \EX_{\pdata{}(\partI{})} \left[ \log \pmodel(\partI{}; \mathbf{t}) \right] \geq \EX_{\pdata{}(\partI{})}\EX_{\partq{}(\z{}|\partI{})} \left[ \log\frac{\pmodel{}(\z{}; \mathbf{t})\pmodel{}(\partI{}|\z{})}{\partq{}(\z{}|\partI{})} \right] .
\end{align}
We hypothesise that OOD examples will usually either be excessively regular, and
therefore have a higher marginal likelihood for $t<1$, or excessively irregular,
and therefore have a higher likelihood for $t>1$. In-distribution (ID) examples,
on the other hand, should have the highest marginal likelihood for $t\approx 1$.
A useful feature to distinguish between ID and OOD examples is therefore likely
to be the gradient $\frac{\partial}{\partial\mathbf{t}}p(\partI{};
\mathbf{t})\lvert_{\mathbf{t}=\mathbf{1}}$, which we can approximate with the
gradient w.r.t. $\mathbf{t}$ of the lower-bound above. We fit a multivariate
Gaussian to 5000 samples of
$\frac{\partial}{\partial\mathbf{t}}p(\partI{}^{(i)};
\mathbf{t})\lvert_{\mathbf{t}=\mathbf{1}}$ with $\partI{}^{(i)}$ sampled from
the training data with the training mask distribution. To obtain a score for how
in-distribution a new example $\partI{}$ is, we compute the probability density
of $\frac{\partial}{\partial\mathbf{t}}p(\partI{};
\mathbf{t})\lvert_{\mathbf{t}=\mathbf{1}}$ under this multivariate Gaussian. The
greater this score, the more likely it is that an example is in-distribution. We
call this the ``temperature-gradient'' metric.

\subsection{OOD detection details} We compute the AUROC scores presented in
\cref{fig:ood-results} using the 5000 image test set for CIFAR-10 and its
OOD-transformations; a 5000 images subset of the SVHN test set; and the 7000
image test set for FFHQ-256. We use 5000 samples of $32\times32$ `Noise' and
`Constant' images when comparing against CIFAR-10, and 200 samples at
$256\times256$ resolution when comparing against FFHQ-256.

We also present results for the `Translated' transformation in
\cref{fig:ood-results} for FFHQ-256. This transformation takes a crop with 90\%
of the image size from one of the four corners and then rescales it to the
original size. The result is thus slightly translated so that e.g.~the eyes of
an FFHQ image are no longer aligned. It is not shown in our results table due to
space constraints.

\section{Lack of semantic diversity in CoModGAN} \label{supp:comodgan-failure}
In this section we demonstrate the some failure cases of CoModGAN in which it
fails to produce semantically diverse images. We suspect that this behaviour of
CoModGAN contributes to its poor performance on the LPIPS-GT metric.
\Cref{fig:comodgan-failure} shows a few examples of partial images for which
CoModGAN struggles to generate semantically diverse images. Compare these
results with IPA's results reported in \cref{fig:comodgan-failure-aipo},
showing a wide range of completions for all the partial images.

\newcommand{\cmgfailureimgheight}{1cm}

\begin{figure*}[t]
  \centering
  \begin{subfigure}[t]{0.14\textwidth}
    \centering
    \includegraphics[trim=0px 0px 2560px 0px, clip, height=\cmgfailureimgheight]{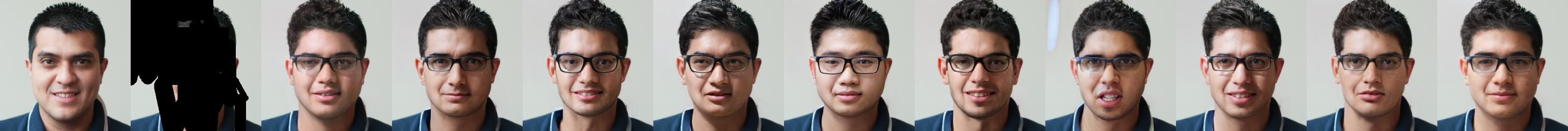}
    \includegraphics[trim=0px 0px 2560px 0px, clip, height=\cmgfailureimgheight]{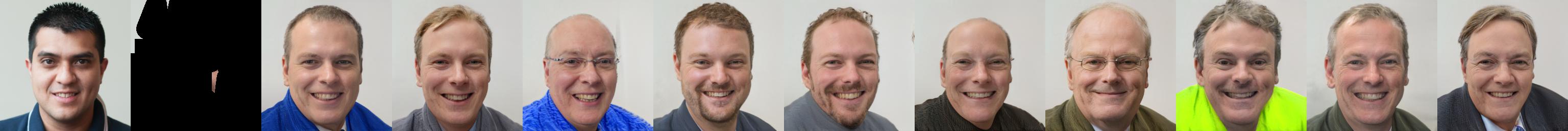}
    \includegraphics[trim=0px 0px 2560px 0px, clip, height=\cmgfailureimgheight]{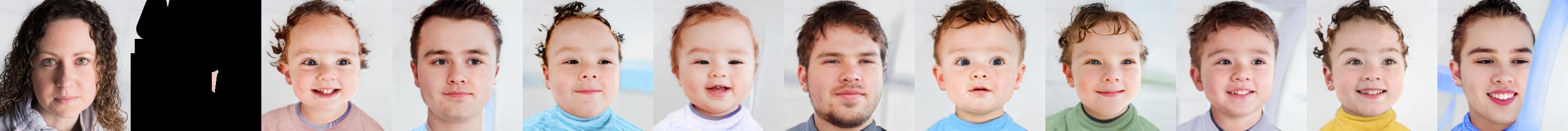}
    \includegraphics[trim=0px 0px 2560px 0px, clip, height=\cmgfailureimgheight]{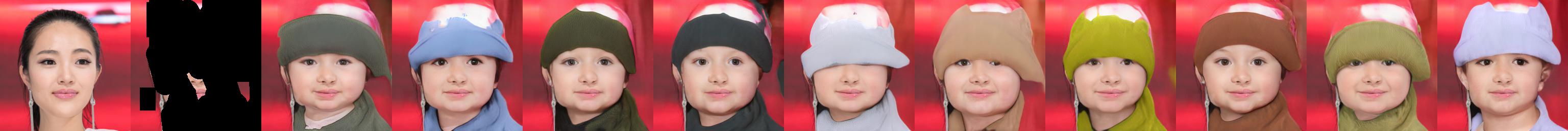}
    \caption{$\I, \partI$}
  \end{subfigure}
  \begin{subfigure}[t]{0.73\textwidth}
    \centering
    \includegraphics[trim=512px 0px 0px 0px, clip, height=\cmgfailureimgheight]{figs/co_mod_gan_failure/co_mod_gan_0_3_2.jpg}
    \includegraphics[trim=512px 0px 0px 0px, clip, height=\cmgfailureimgheight]{figs/co_mod_gan_failure/co_mod_gan_0_4_2.jpg}
    \includegraphics[trim=512px 0px 0px 0px, clip, height=\cmgfailureimgheight]{figs/co_mod_gan_failure/co_mod_gan_1_4_2.jpg}
    \includegraphics[trim=512px 0px 0px 0px, clip, height=\cmgfailureimgheight]{figs/co_mod_gan_failure/co_mod_gan_56_4_12.jpg}
    \caption{Sampled completions}
  \end{subfigure}
  \begin{subfigure}[t]{0.1\textwidth}
    \centering
    \includegraphics[height=\cmgfailureimgheight]{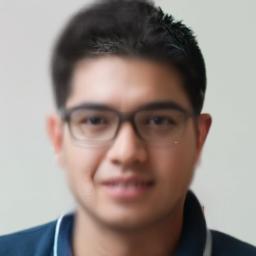}
    \includegraphics[height=\cmgfailureimgheight]{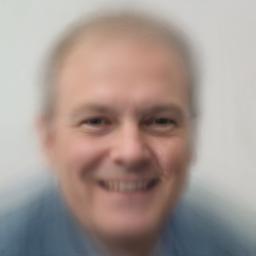}
    \includegraphics[height=\cmgfailureimgheight]{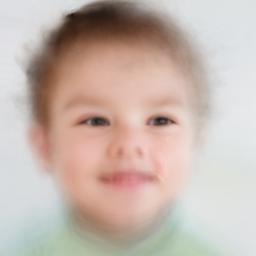}
    \includegraphics[height=\cmgfailureimgheight]{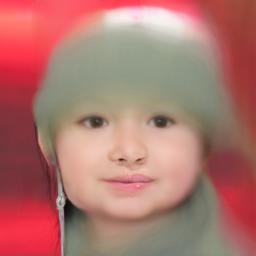}
    \caption{Mean}
  \end{subfigure}
  %%%%%%%%%%%%%%%%%%%%%%
  \caption{Examples of completions lacking semantic diversity in CoModGAN. Panel
    (a) shows the true image and the masked version on which the completions are
    conditioned. Panel (b) shows 10 completions sampled randomly from the
    CoModGAN model. Panel (c) shows the mean image computed from 100 sampled
    completions. On each row, the completions are mostly semantically similar to
    eachother yet different from the ground truth image, indicating that they
    are not faithfully representing the true posterior. This behaviour can be
    contrasted with that of IPA in \cref{fig:comodgan-failure-aipo}.}
  \label{fig:comodgan-failure}
  \vspace{.5cm}
  \begin{subfigure}[t]{0.14\textwidth}
    \centering
    \includegraphics[trim=0px 0px 2560px 0px, clip, height=\cmgfailureimgheight]{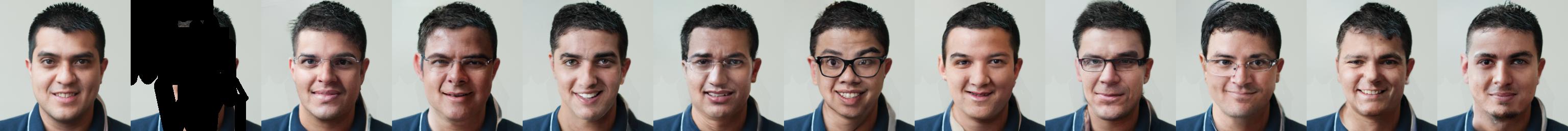}
    \includegraphics[trim=0px 0px 2560px 0px, clip, height=\cmgfailureimgheight]{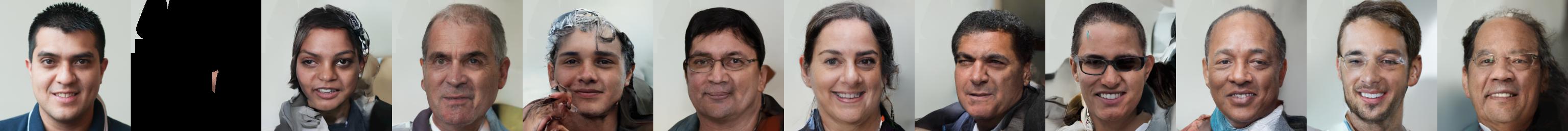}
    \includegraphics[trim=0px 0px 2560px 0px, clip, height=\cmgfailureimgheight]{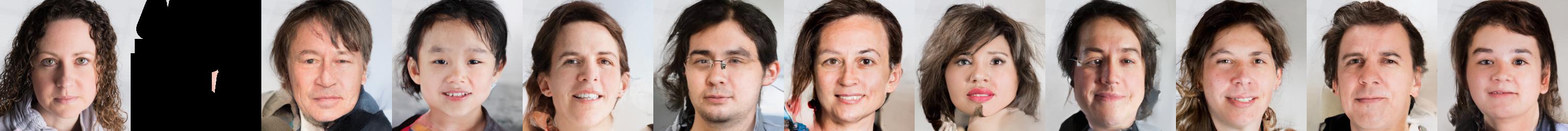}
    \includegraphics[trim=0px 0px 2560px 0px, clip, height=\cmgfailureimgheight]{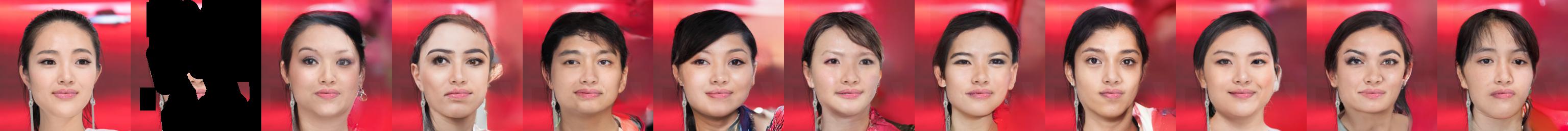}
    \caption{$\I, \partI$}
  \end{subfigure}
  \begin{subfigure}[t]{0.73\textwidth}
    \centering
    \includegraphics[trim=512px 0px 0px 0px, clip, height=\cmgfailureimgheight]{figs/co_mod_gan_failure/aipo_0_3_2.jpg}
    \includegraphics[trim=512px 0px 0px 0px, clip, height=\cmgfailureimgheight]{figs/co_mod_gan_failure/aipo_0_4_2.jpg}
    \includegraphics[trim=512px 0px 0px 0px, clip, height=\cmgfailureimgheight]{figs/co_mod_gan_failure/aipo_1_4_2.jpg}
    \includegraphics[trim=512px 0px 0px 0px, clip, height=\cmgfailureimgheight]{figs/co_mod_gan_failure/aipo_56_4_12.jpg}
    \caption{Sampled completions}
  \end{subfigure}
  \begin{subfigure}[t]{0.1\textwidth}
    \centering
    \includegraphics[height=\cmgfailureimgheight]{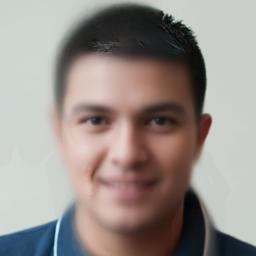}
    \includegraphics[height=\cmgfailureimgheight]{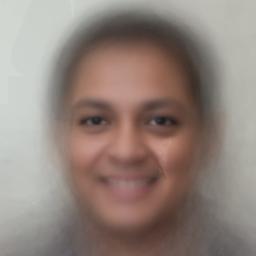}
    \includegraphics[height=\cmgfailureimgheight]{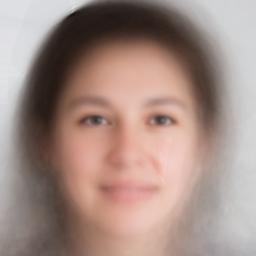}
    \includegraphics[height=\cmgfailureimgheight]{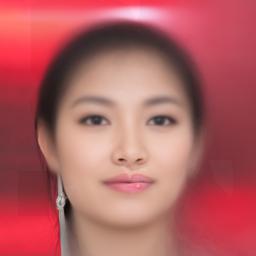}
    \caption{Mean}
  \end{subfigure}
  %%%%%%%%%%%%%%%%%%%%%%
  \caption{Examples completions from our IPA model, to compare with
    \cref{fig:comodgan-failure}. These samples from IPA provide a much better
    representation of the inherent uncertainty in the image given the
    observations. We did not find any $(\I{}, \partI{})$ pairs on which the IPA
    model failed in the same way as CoModGAN in \cref{fig:comodgan-failure}. }
  \label{fig:comodgan-failure-aipo}
\end{figure*}

\section{Image samples}\label{supp:image-samples}

\Cref{fig:cifar-samples-0,fig:imagenet-samples,fig:ffhq-samples,fig:bags-samples,fig:shoes-samples,fig:xray-samples}
on the following pages show conditionally generated images for all datasets we have
experimented with, from both IPA and our baselines. It is apparent from the
image samples, particularly when the number of observed pixels is small, that
IPA-R has less sample diversity than IPA, which can be explained by the
mode-seeking behaviour of the reverse-KL and subsequent under-estimation of
posterior uncertainty \citep{minka2005divergence}. Finally, we note a flaw with
some of the completions produced by IPA: a lack of bilateral symmetry in
FFHQ-256 samples. Again, this is most apparent when few pixels are observed.
This issue can be seen in unconditional samples from the underlying VAEs as well
(see the samples reported by \citet{child2020very}). Therefore, it is likely
that any future advances in image modelling with VAEs could be integrated to
improve this aspect of the results.

% \section{Licences}
% The Chest X-ray 14 dataset~\citep{wang2017chestx} was released under the CC0:
% Public Domain Licence. The FFHQ dataset was released under the Creative Commons
% BY-NC-SA 4.0 Licence. CIFAR-10~\citep{krizhevsky2009learning} was released under
% the MIT licence. SVHN~\citep{netzer2011reading} was released under the CC0 1.0
% Universal (CC0 1.0) Licence. IPA and IPA-R make use of code and pretrained
% weights released by \citet{child2020very} under the MIT Licence. Our
% implementations of our baselines are based on code (linked to in \cref{supp:exp-details})
% licensed as follows: \textbf{PIC:} Creative Commons Attribution-NonCommercial 4.0
% International License; \textbf{ANP:} MIT Licence; \textbf{CE:} MIT Licence; \textbf{CoModGAN:}
% Nvidia Source Code License-NC; \textbf{RFR:} MIT Licence; \textbf{VQ-VAE:} MIT License.

\newcommand{\qualimgheight}{4cm} \newcommand{\qualcaption}[1]{Sampled
  completions from each method for #1 image. Panels (a), (d) and (h) all show
  the true image and the masked image on which the samples in each row are
  conditioned. Panels (b) and (c) show completions from IPA and IPA-R. (g) and
  (k) show the single deterministic completion produced by CE and RFR
  respectively. Each of the remaining panels show five completions for the
  stochastic baselines.} \newcommand{\freeformimgheight}{1.25cm}

% CIFAR samples -------------------------------------------------------------------------------------------------------------
\newcommand{\cifarimgheight}{2.2cm}
\newcommand{\clipby}{20}
  \begin{figure*}[t]
    \centering
    \begin{subfigure}[t]{0.11\textwidth}
      \centering
      \includegraphics[height=\cifarimgheight]{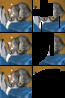}
      \includegraphics[height=\cifarimgheight]{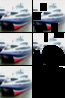}
      \includegraphics[height=\cifarimgheight]{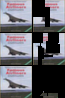}
      \caption{$\I, \partI$}
    \end{subfigure}
    \begin{subfigure}[t]{0.17\textwidth}
      \centering
      \includegraphics[height=\cifarimgheight]{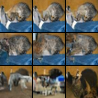}
      \includegraphics[height=\cifarimgheight]{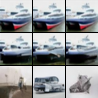}
      \includegraphics[height=\cifarimgheight]{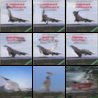}
      \caption{IPA, $t=0.85$}
    \end{subfigure}
    \begin{subfigure}[t]{0.17\textwidth}
      \centering
      \includegraphics[height=\cifarimgheight]{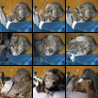}
      \includegraphics[height=\cifarimgheight]{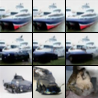}
      \includegraphics[height=\cifarimgheight]{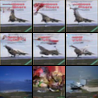}
      \caption{IPA}
    \end{subfigure}
    \begin{subfigure}[t]{0.17\textwidth}
      \centering
      \includegraphics[height=\cifarimgheight]{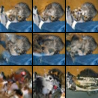}
      \includegraphics[height=\cifarimgheight]{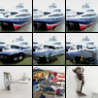}
      \includegraphics[height=\cifarimgheight]{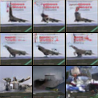}
      \caption{\scriptsize IPA from ImageNet}
    \end{subfigure}
    \begin{subfigure}[t]{0.17\textwidth}
      \centering
      \includegraphics[height=\cifarimgheight]{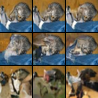}
      \includegraphics[height=\cifarimgheight]{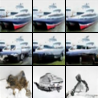}
      \includegraphics[height=\cifarimgheight]{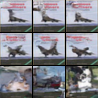}
      \caption{\scriptsize IPA from scratch}
    \end{subfigure}
    \begin{subfigure}[t]{0.17\textwidth}
      \centering
      \includegraphics[height=\cifarimgheight]{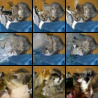}
      \includegraphics[height=\cifarimgheight]{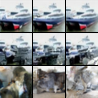}
      \includegraphics[height=\cifarimgheight]{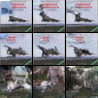}
      \caption{IPA-R}
    \end{subfigure}
    %%%%%%%%%%%%%%%%%%%%%%
    \begin{subfigure}[t]{0.11\textwidth}
      \centering
      \includegraphics[height=\cifarimgheight]{figs/image-samples/cifar10/freeform_aipo_0_gt_masked}
      \includegraphics[height=\cifarimgheight]{figs/image-samples/cifar10/freeform_aipo_1_gt_masked}
      \includegraphics[height=\cifarimgheight]{figs/image-samples/cifar10/freeform_aipo_3_gt_masked}
      \caption*{$\I, \partI$}
    \end{subfigure}
    \begin{subfigure}[t]{0.17\textwidth}
      \centering
      \includegraphics[height=\cifarimgheight]{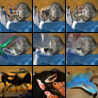}
      \includegraphics[height=\cifarimgheight]{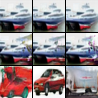}
      \includegraphics[height=\cifarimgheight]{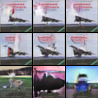}
      \caption{CoModGAN}
    \end{subfigure}
    \begin{subfigure}[t]{0.17\textwidth}
      \centering
      \includegraphics[height=\cifarimgheight]{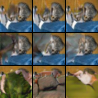}
      \includegraphics[height=\cifarimgheight]{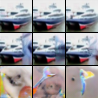}
      \includegraphics[height=\cifarimgheight]{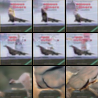}
      \caption{PIC}
    \end{subfigure}
    \begin{subfigure}[t]{0.17\textwidth}
      \centering
      \includegraphics[height=\cifarimgheight]{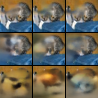}
      \includegraphics[height=\cifarimgheight]{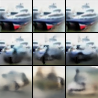}
      \includegraphics[height=\cifarimgheight]{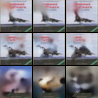}
      \caption{ANP}
    \end{subfigure}
    \begin{subfigure}[t]{0.17\textwidth}
      \centering
      \includegraphics[height=\cifarimgheight]{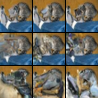}
      \includegraphics[height=\cifarimgheight]{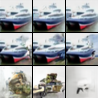}
      \includegraphics[height=\cifarimgheight]{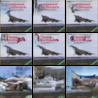}
      \caption{VQ-VAE}
    \end{subfigure}
    \begin{subfigure}[t]{0.08\textwidth}
      \centering
      \includegraphics[height=\cifarimgheight]{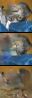}
      \includegraphics[height=\cifarimgheight]{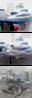}
      \includegraphics[height=\cifarimgheight]{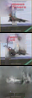}
      \caption{CE}
    \end{subfigure}
    \begin{subfigure}[t]{0.08\textwidth}
      \centering
      \includegraphics[height=\cifarimgheight]{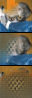}
      \includegraphics[height=\cifarimgheight]{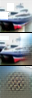}
      \includegraphics[height=\cifarimgheight]{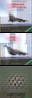}
      \caption{RFR}
    \end{subfigure}
    \caption{Sampled completions on CIFAR-10. Panel (a) shows test images along
      with masked versions on which samples in each row are conditioned. The
      remaining panels in the top row show samples from IPA (with an artifact
      pretrained on the same dataset) with and without a reduced temperature,
      IPA with an artifact pretrained on ImageNet, a conditional VAE trained
      from scratch, and finally IPA-R (with an artifact pretrained on the same
      dataset). The bottom row shows samples from our baselines. Three samples
      per masked image are shown for each stochastic method, while the single
      deterministic completion is shown for CE and RFR. All samples are taken
      with temperature 1 where not stated otherwise.}
    \label{fig:cifar-samples-0}
  \end{figure*}

  % ImageNet64 samples -------------------------------------------------------------------------------------------------------
% show samples for 0, 1, 2, 3, 4
\newcommand{\imagenetimgheight}{1.05cm}
  \begin{figure*}[t]
    \centering
    \begin{subfigure}[t]{0.15\textwidth}
      \centering
      \includegraphics[height=\imagenetimgheight]{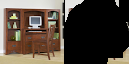}
      \includegraphics[height=\imagenetimgheight]{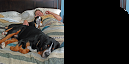}
      \includegraphics[height=\imagenetimgheight]{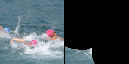}
      \includegraphics[height=\imagenetimgheight]{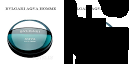}
      \includegraphics[height=\imagenetimgheight]{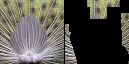}
      \caption{$\I, \partI$}
    \end{subfigure}
    \begin{subfigure}[t]{0.2\textwidth}
      \centering
      \includegraphics[height=\imagenetimgheight]{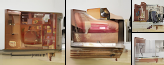}
      \includegraphics[height=\imagenetimgheight]{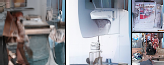}
      \includegraphics[height=\imagenetimgheight]{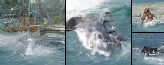}
      \includegraphics[height=\imagenetimgheight]{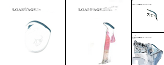}
      \includegraphics[height=\imagenetimgheight]{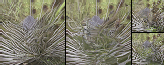}
      \caption{IPA, $t=0.85$}
    \end{subfigure}
    \begin{subfigure}[t]{0.2\textwidth}
      \centering
      \includegraphics[height=\imagenetimgheight]{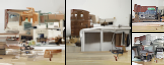}
      \includegraphics[height=\imagenetimgheight]{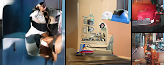}
      \includegraphics[height=\imagenetimgheight]{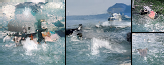}
      \includegraphics[height=\imagenetimgheight]{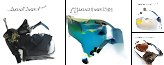}
      \includegraphics[height=\imagenetimgheight]{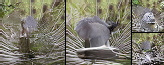}
      \caption{IPA}
    \end{subfigure}
    \begin{subfigure}[t]{0.2\textwidth}
      \centering
      \includegraphics[height=\imagenetimgheight]{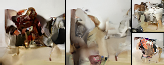}
      \includegraphics[height=\imagenetimgheight]{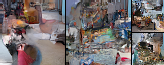}
      \includegraphics[height=\imagenetimgheight]{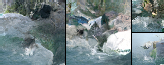}
      \includegraphics[height=\imagenetimgheight]{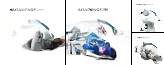}
      \includegraphics[height=\imagenetimgheight]{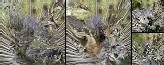}
      \caption{IPA from scratch}
    \end{subfigure}
    \begin{subfigure}[t]{0.2\textwidth}
      \centering
      \includegraphics[height=\imagenetimgheight]{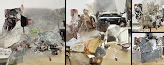}
      \includegraphics[height=\imagenetimgheight]{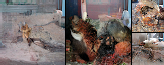}
      \includegraphics[height=\imagenetimgheight]{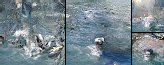}
      \includegraphics[height=\imagenetimgheight]{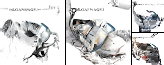}
      \includegraphics[height=\imagenetimgheight]{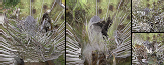}
      \caption{IPA-R}
    \end{subfigure}
    \begin{subfigure}[t]{0.15\textwidth}
      \centering
      \includegraphics[height=\imagenetimgheight]{figs/image-samples/imagenet64/freeform_aipo_0_gt_masked.png}
      \includegraphics[height=\imagenetimgheight]{figs/image-samples/imagenet64/freeform_aipo_1_gt_masked.png}
      \includegraphics[height=\imagenetimgheight]{figs/image-samples/imagenet64/freeform_aipo_2_gt_masked.png}
      \includegraphics[height=\imagenetimgheight]{figs/image-samples/imagenet64/freeform_aipo_3_gt_masked.png}
      \includegraphics[height=\imagenetimgheight]{figs/image-samples/imagenet64/freeform_aipo_4_gt_masked.png}
      \caption*{$\I, \partI$}
    \end{subfigure}
    \begin{subfigure}[t]{0.2\textwidth}
      \centering
      \includegraphics[height=\imagenetimgheight]{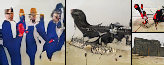}
      \includegraphics[height=\imagenetimgheight]{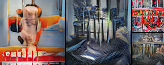}
      \includegraphics[height=\imagenetimgheight]{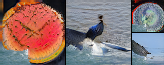}
      \includegraphics[height=\imagenetimgheight]{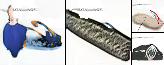}
      \includegraphics[height=\imagenetimgheight]{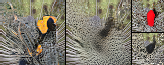}
      \caption{CoModGAN}
    \end{subfigure}
    \begin{subfigure}[t]{0.2\textwidth}
      \centering
      \includegraphics[height=\imagenetimgheight]{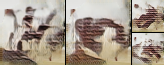}
      \includegraphics[height=\imagenetimgheight]{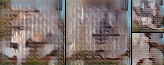}
      \includegraphics[height=\imagenetimgheight]{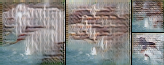}
      \includegraphics[height=\imagenetimgheight]{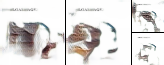}
      \includegraphics[height=\imagenetimgheight]{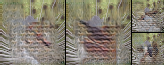}
      \caption{PIC}
    \end{subfigure}
    \begin{subfigure}[t]{0.4\textwidth}
      \setlength{\fboxrule}{0pt}
      \fbox{
        \begin{minipage}{5cm}
          \hfill\hspace{5cm}
        \end{minipage}
      }
    \end{subfigure}
    \caption{Sampled completions on ImageNet-64. Panel (a) shows a test image
      and the masked version on which samples in each row are conditioned. The
      remaining panels in the top row show samples from IPA with and without a
      reduced temperature, from an IPA-style conditional VAE trained from
      scratch, and from IPA-R. The bottom row shows samples CoModGAN.}
    \vspace{-.5cm}
    \label{fig:imagenet-samples}
  \end{figure*}

  % FFHQ samples -------------------------------------------------------------------------------------------------------------
% show samples for 0, 13, 32
\newcommand{\ffhqimgheight}{1.4cm}
  \begin{figure*}[t]
    \centering
    \begin{subfigure}[t]{0.22\textwidth}
      \centering
      \includegraphics[height=\ffhqimgheight]{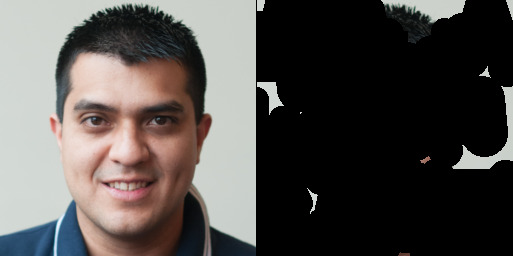}
      \includegraphics[height=\ffhqimgheight]{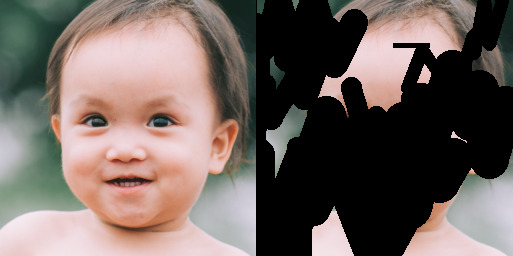}
      \includegraphics[height=\ffhqimgheight]{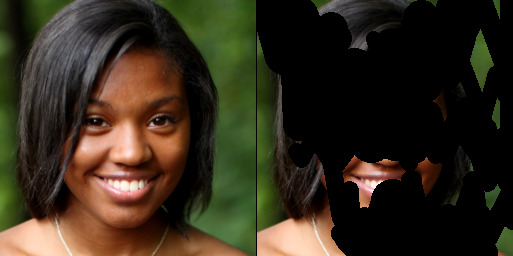}
      \caption{$\I, \partI$}
    \end{subfigure}
    \begin{subfigure}[t]{0.25\textwidth}
      \centering
      \includegraphics[height=\ffhqimgheight]{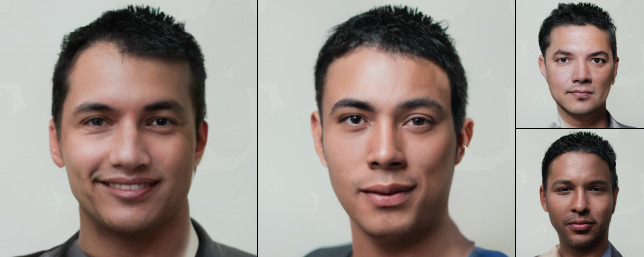}
      \includegraphics[height=\ffhqimgheight]{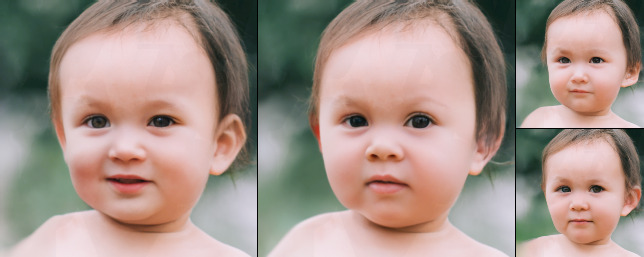}
      \includegraphics[height=\ffhqimgheight]{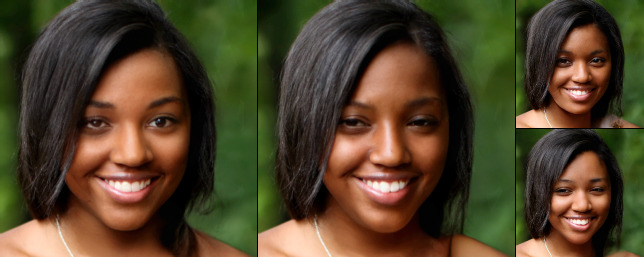}
      \caption{IPA, $t=0.85$}
    \end{subfigure}
    \begin{subfigure}[t]{0.25\textwidth}
      \centering
      \includegraphics[height=\ffhqimgheight]{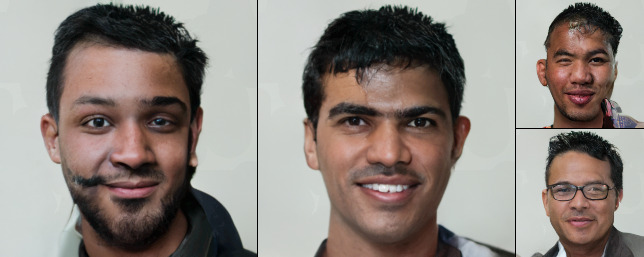}
      \includegraphics[height=\ffhqimgheight]{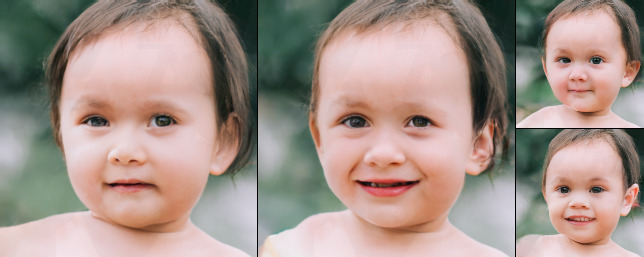}
      \includegraphics[height=\ffhqimgheight]{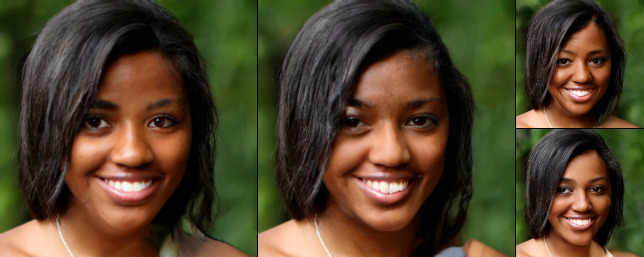}
      \caption{IPA}
    \end{subfigure}
    \begin{subfigure}[t]{0.25\textwidth}
      \centering
      \includegraphics[height=\ffhqimgheight]{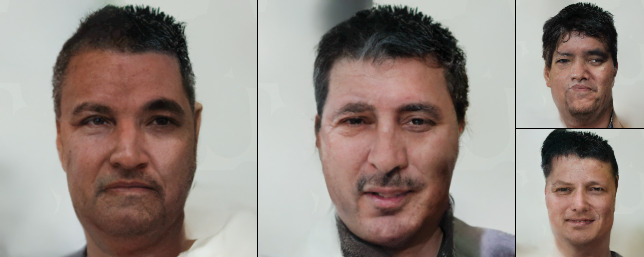}
      \includegraphics[height=\ffhqimgheight]{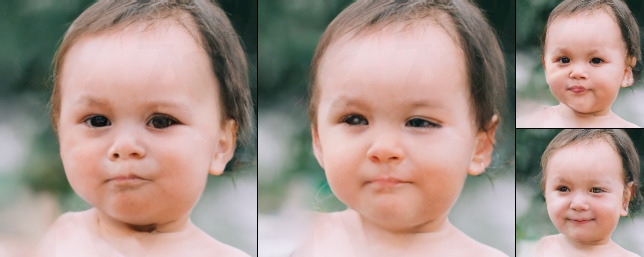}
      \includegraphics[height=\ffhqimgheight]{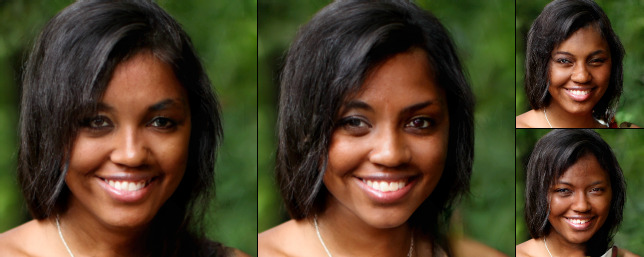}
      \caption{IPA-R}
    \end{subfigure}
    %%%%%%%%%%%%%%%%%%%%%%
    \begin{subfigure}[t]{0.22\textwidth}
      \centering
      \includegraphics[height=\ffhqimgheight]{figs/image-samples/ffhq256/freeform_aipo_0_gt_masked.jpg}
      \includegraphics[height=\ffhqimgheight]{figs/image-samples/ffhq256/freeform_aipo_13_gt_masked.jpg}
      \includegraphics[height=\ffhqimgheight]{figs/image-samples/ffhq256/freeform_aipo_32_gt_masked.jpg}
      \caption*{$\I, \partI$}
    \end{subfigure}
    \begin{subfigure}[t]{0.25\textwidth}
      \centering
      \includegraphics[height=\ffhqimgheight]{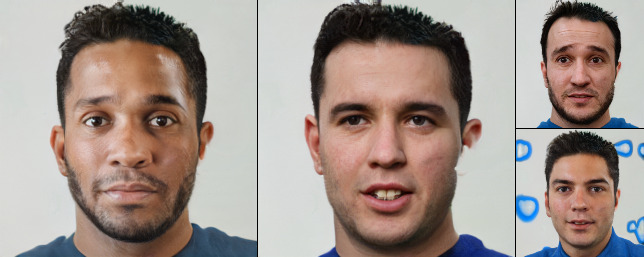}
      \includegraphics[height=\ffhqimgheight]{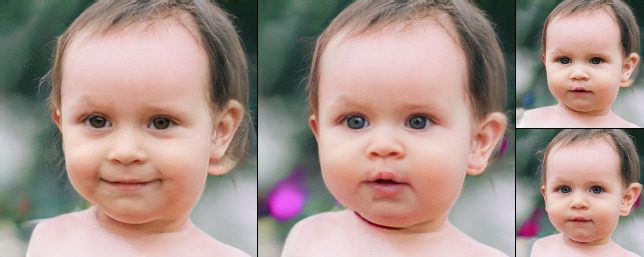}
      \includegraphics[height=\ffhqimgheight]{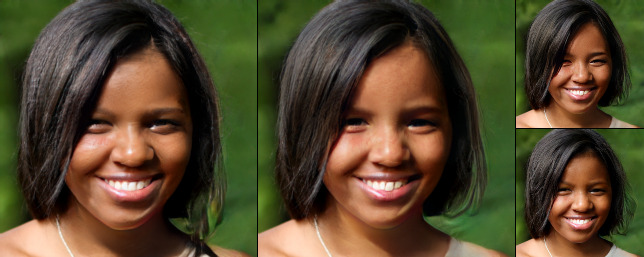}
      \caption{CoModGAN}
    \end{subfigure}
    \begin{subfigure}[t]{0.25\textwidth}
      \centering
      \includegraphics[height=\ffhqimgheight]{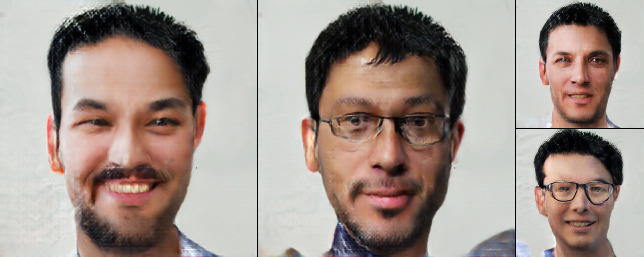}
      \includegraphics[height=\ffhqimgheight]{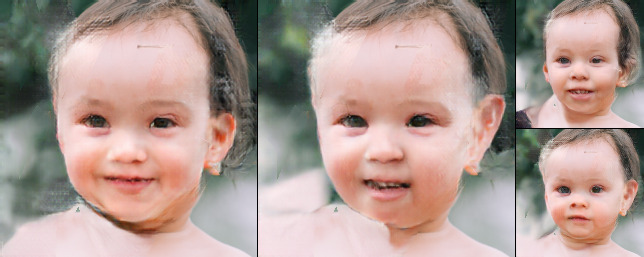}
      \includegraphics[height=\ffhqimgheight]{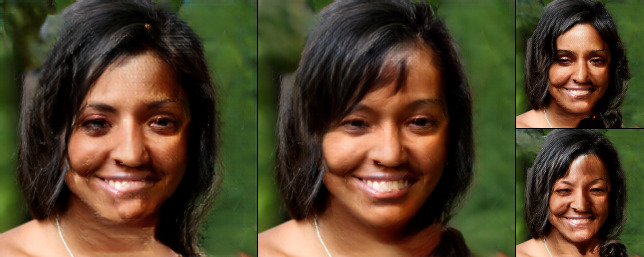}
      \caption{PIC}
    \end{subfigure}
    \begin{subfigure}[t]{0.25\textwidth}
      \centering
      \includegraphics[height=\ffhqimgheight]{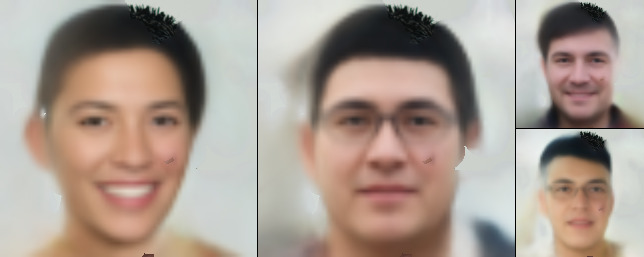}
      \includegraphics[height=\ffhqimgheight]{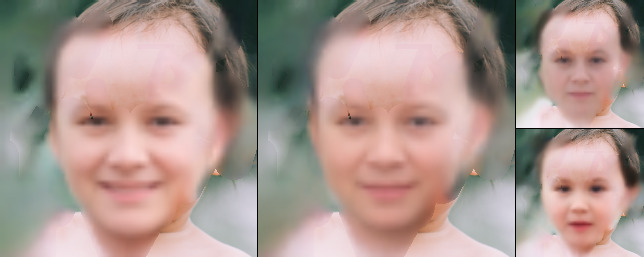}
      \includegraphics[height=\ffhqimgheight]{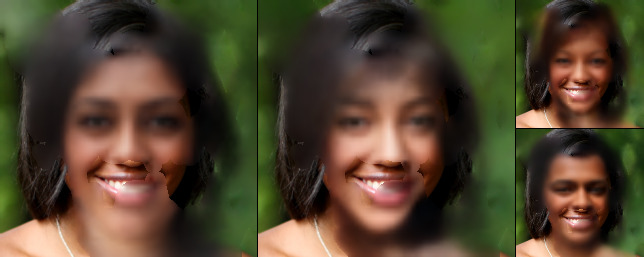}
      \caption{ANP}
    \end{subfigure}
    %%%%%%%%%%%%%%%%%%%%%%
    \begin{subfigure}[t]{0.22\textwidth}
      \centering
      \includegraphics[height=\ffhqimgheight]{figs/image-samples/ffhq256/freeform_aipo_0_gt_masked.jpg}
      \includegraphics[height=\ffhqimgheight]{figs/image-samples/ffhq256/freeform_aipo_13_gt_masked.jpg}
      \includegraphics[height=\ffhqimgheight]{figs/image-samples/ffhq256/freeform_aipo_32_gt_masked.jpg}
      \caption*{$\I, \partI$}
    \end{subfigure}
    \begin{subfigure}[t]{0.25\textwidth}
      \centering
      \includegraphics[height=\ffhqimgheight]{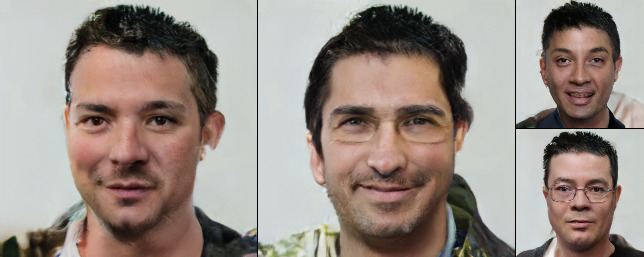}
      \includegraphics[height=\ffhqimgheight]{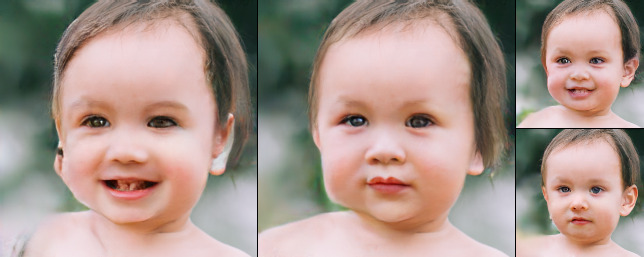}
      \includegraphics[height=\ffhqimgheight]{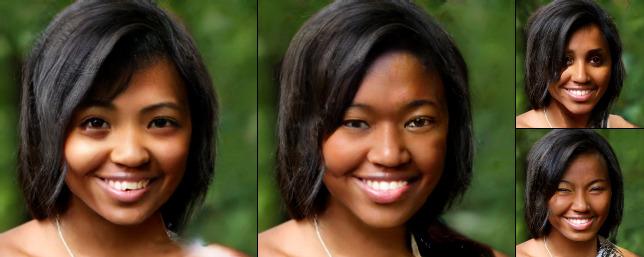}
      \caption{VQ-VAE}
    \end{subfigure}
    \begin{subfigure}[t]{0.25\textwidth}
      \centering
      \includegraphics[height=\ffhqimgheight]{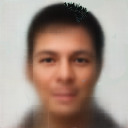}\\
      \includegraphics[height=\ffhqimgheight]{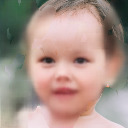}\\
      \includegraphics[height=\ffhqimgheight]{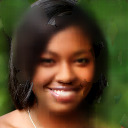}
      \caption{CE}
    \end{subfigure}
    \begin{subfigure}[t]{0.25\textwidth}
      \centering
      \includegraphics[height=\ffhqimgheight]{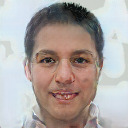}\\
      \includegraphics[height=\ffhqimgheight]{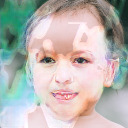}\\
      \includegraphics[height=\ffhqimgheight]{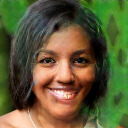}
      \caption{RFR}
    \end{subfigure}
    \begin{subfigure}[t]{0.16\textwidth}
      \setlength{\fboxrule}{0pt}
      \fbox{
        \begin{minipage}{1in}
          \hfill\hspace{1in}
        \end{minipage}
      }
    \end{subfigure}
    \vspace{-.5cm}
    \caption{Sampled completions on FFHQ-256. Panel (a) shows test images along with
      masked versions on which samples in each row are conditioned. The
      remaining panels in the top row show samples from IPA with and without a
      reduced temperature and from IPA-R. The other rows show samples from our
      baselines. Four samples per masked image are shown for each stochastic
      method, while the single deterministic completion is shown for CE and
      RFR.}
    \vspace{-.5cm}
    \label{fig:ffhq-samples}
  \end{figure*}

  \clearpage
  % Edges2Stuff samples ------------------------------------------------------------------------------------------------------
% show samples for ...
\newcommand{\edgesstuffimgheight}{1.05cm}
  \begin{figure*}[t]
    \centering
    \begin{subfigure}[t]{0.15\textwidth}
      \centering
      \includegraphics[height=\edgesstuffimgheight]{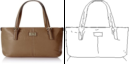}
      \includegraphics[height=\edgesstuffimgheight]{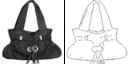}
      \includegraphics[height=\edgesstuffimgheight]{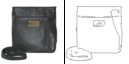}
      \includegraphics[height=\edgesstuffimgheight]{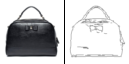}
      \includegraphics[height=\edgesstuffimgheight]{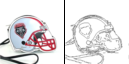}
      \caption{$\I, \partI$}
    \end{subfigure}
    \begin{subfigure}[t]{0.27\textwidth}
      \centering
      \includegraphics[height=\edgesstuffimgheight]{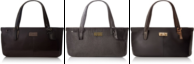}
      \includegraphics[height=\edgesstuffimgheight]{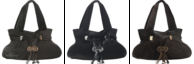}
      \includegraphics[height=\edgesstuffimgheight]{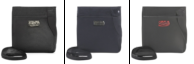}
      \includegraphics[height=\edgesstuffimgheight]{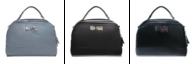}
      \includegraphics[height=\edgesstuffimgheight]{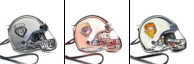}
      \caption{\scriptsize IPA from ImageNet, $t$=$0.85$}
    \end{subfigure}
    \begin{subfigure}[t]{0.27\textwidth}
      \centering
      \includegraphics[height=\edgesstuffimgheight]{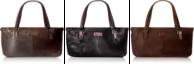}
      \includegraphics[height=\edgesstuffimgheight]{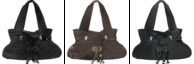}
      \includegraphics[height=\edgesstuffimgheight]{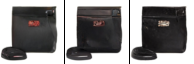}
      \includegraphics[height=\edgesstuffimgheight]{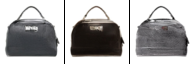}
      \includegraphics[height=\edgesstuffimgheight]{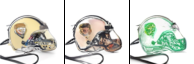}
      \caption{IPA from ImageNet}
    \end{subfigure}
    \begin{subfigure}[t]{0.27\textwidth}
      \centering
      \includegraphics[height=\edgesstuffimgheight]{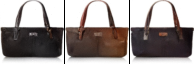}
      \includegraphics[height=\edgesstuffimgheight]{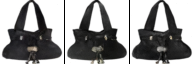}
      \includegraphics[height=\edgesstuffimgheight]{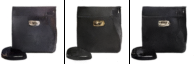}
      \includegraphics[height=\edgesstuffimgheight]{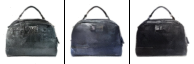}
      \includegraphics[height=\edgesstuffimgheight]{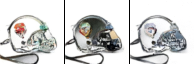}
      \caption{IPA from scratch}
    \end{subfigure}
    \caption{Comparison of conditional generations on Edges2Bags.}
    \vspace{-.5cm}
    \label{fig:bags-samples}
  \end{figure*}

  \begin{figure*}[t]
    \centering
    \begin{subfigure}[t]{0.2\textwidth}
      \centering
      \includegraphics[height=\edgesstuffimgheight]{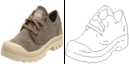}
      \includegraphics[height=\edgesstuffimgheight]{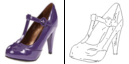}
      \includegraphics[height=\edgesstuffimgheight]{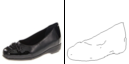}
      \includegraphics[height=\edgesstuffimgheight]{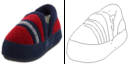}
      \includegraphics[height=\edgesstuffimgheight]{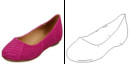}
      \caption{$\I, \partI$}
    \end{subfigure}
    \begin{subfigure}[t]{0.25\textwidth}
      \centering
      \includegraphics[height=\edgesstuffimgheight]{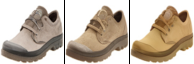}
      \includegraphics[height=\edgesstuffimgheight]{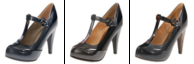}
      \includegraphics[height=\edgesstuffimgheight]{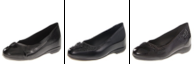}
      \includegraphics[height=\edgesstuffimgheight]{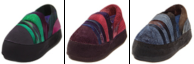}
      \includegraphics[height=\edgesstuffimgheight]{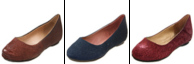}
      \caption{\scriptsize IPA from ImageNet, $t$=$0.85$}
    \end{subfigure}
    \begin{subfigure}[t]{0.25\textwidth}
      \centering
      \includegraphics[height=\edgesstuffimgheight]{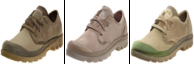}
      \includegraphics[height=\edgesstuffimgheight]{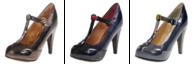}
      \includegraphics[height=\edgesstuffimgheight]{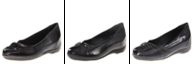}
      \includegraphics[height=\edgesstuffimgheight]{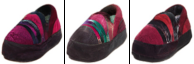}
      \includegraphics[height=\edgesstuffimgheight]{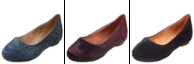}
      \caption{IPA from ImageNet}
    \end{subfigure}
    \begin{subfigure}[t]{0.25\textwidth}
      \centering
      \includegraphics[height=\edgesstuffimgheight]{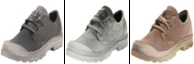}
      \includegraphics[height=\edgesstuffimgheight]{figs/image-samples/shoes/image_aipo_1_scratch_samples.png}
      \includegraphics[height=\edgesstuffimgheight]{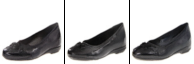}
      \includegraphics[height=\edgesstuffimgheight]{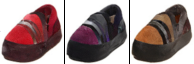}
      \includegraphics[height=\edgesstuffimgheight]{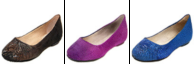}
      \caption{IPA from scratch}
    \end{subfigure}
    \caption{Comparison of conditional generations on Edges2Shoes.}
    \vspace{-.5cm}
    \label{fig:shoes-samples}
  \end{figure*}

  % x-ray samples ---------------------------------------------------------------------------------------------------------
\newcommand{\xrayimgheight}{1.05cm}
\begin{figure*}[t]
  \centering
  \begin{subfigure}[t]{0.16\textwidth}
    \centering
    \includegraphics[height=\xrayimgheight]{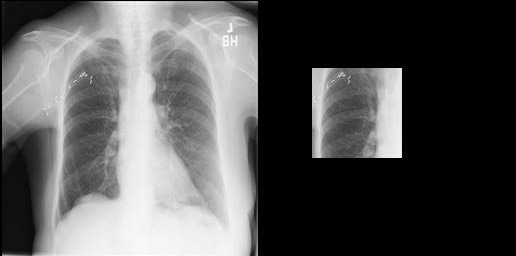}
    \includegraphics[height=\xrayimgheight]{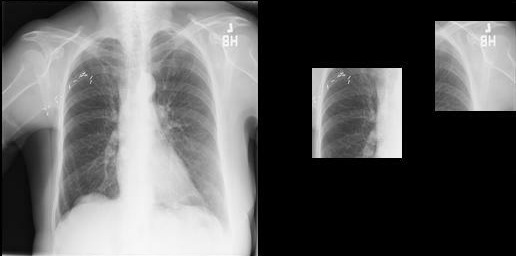}
    \includegraphics[height=\xrayimgheight]{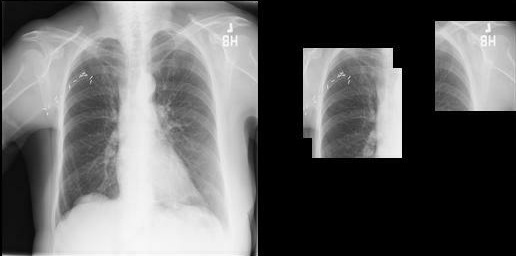}
    \includegraphics[height=\xrayimgheight]{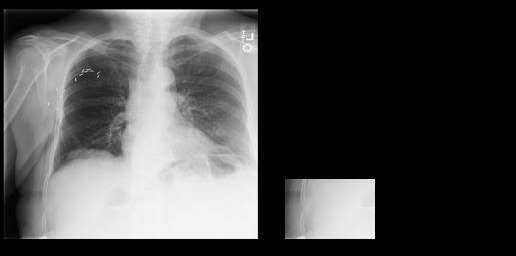}
    \includegraphics[height=\xrayimgheight]{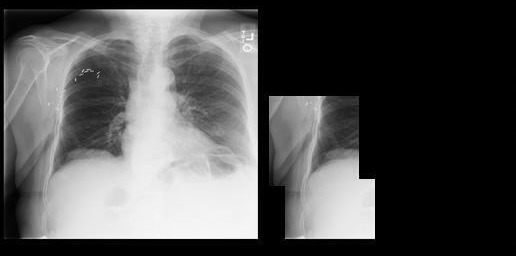}
    \includegraphics[height=\xrayimgheight]{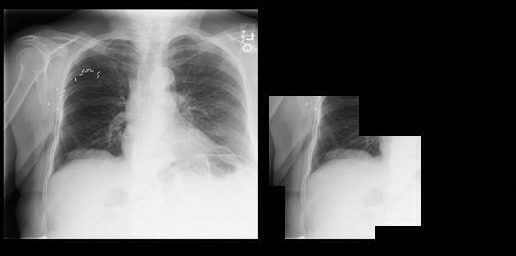}
    \caption{$\I, \partI$}
  \end{subfigure}
  \begin{subfigure}[t]{0.4\textwidth}
    \centering
    \includegraphics[height=\xrayimgheight]{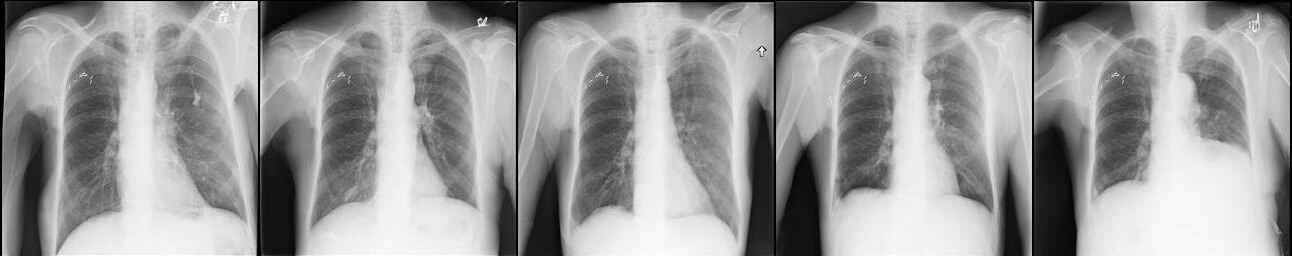}
    \includegraphics[height=\xrayimgheight]{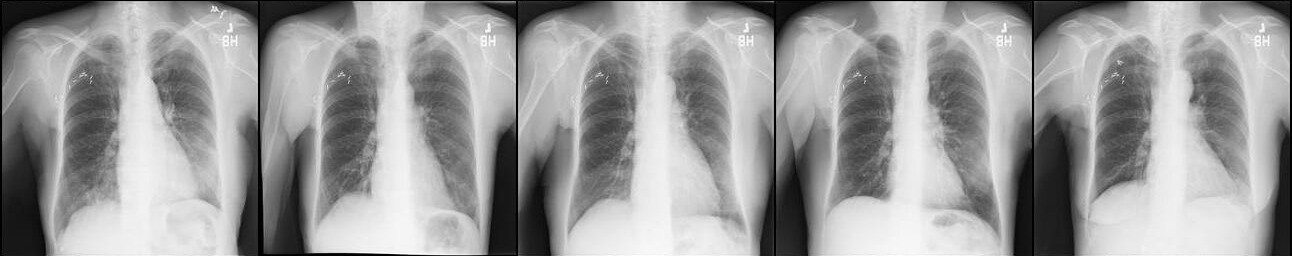}
    \includegraphics[height=\xrayimgheight]{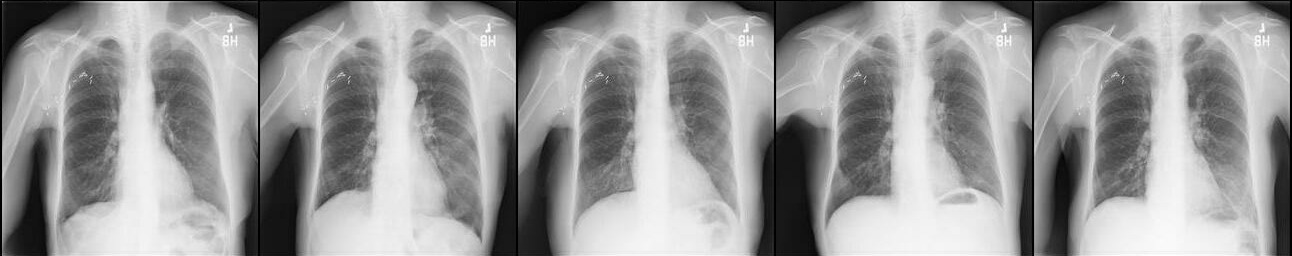}
    \includegraphics[height=\xrayimgheight]{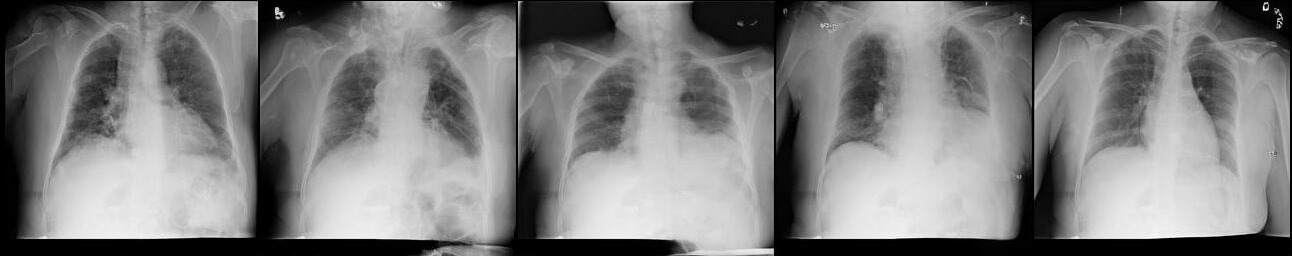}
    \includegraphics[height=\xrayimgheight]{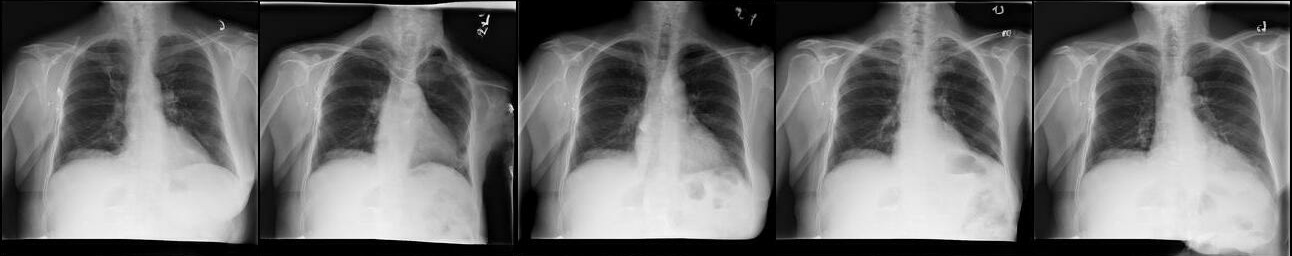}
    \includegraphics[height=\xrayimgheight]{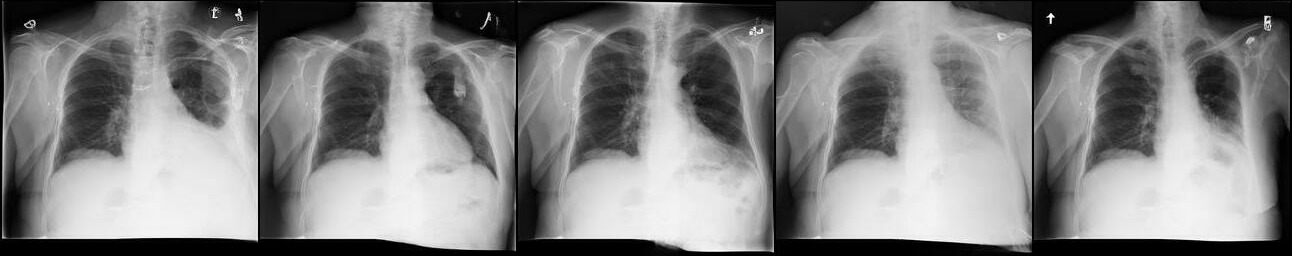}
    \caption{IPA}
  \end{subfigure}
  \begin{subfigure}[t]{0.4\textwidth}
    \centering
    \includegraphics[height=\xrayimgheight]{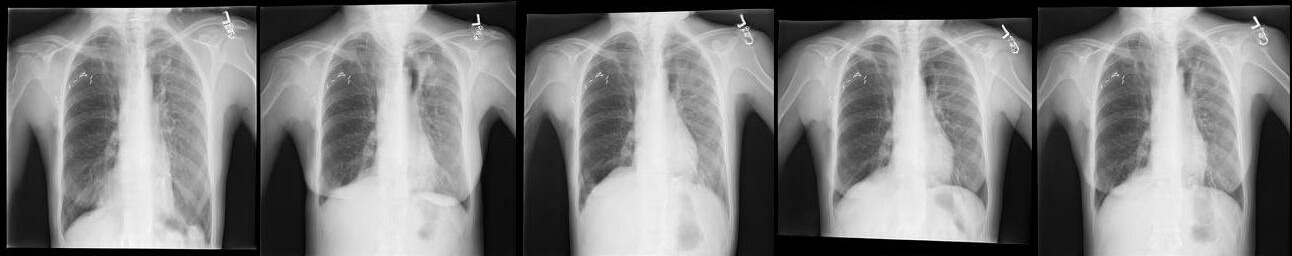}
    \includegraphics[height=\xrayimgheight]{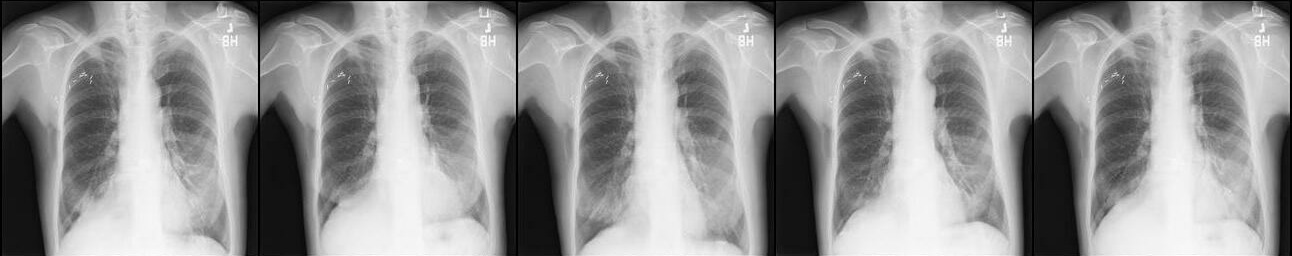}
    \includegraphics[height=\xrayimgheight]{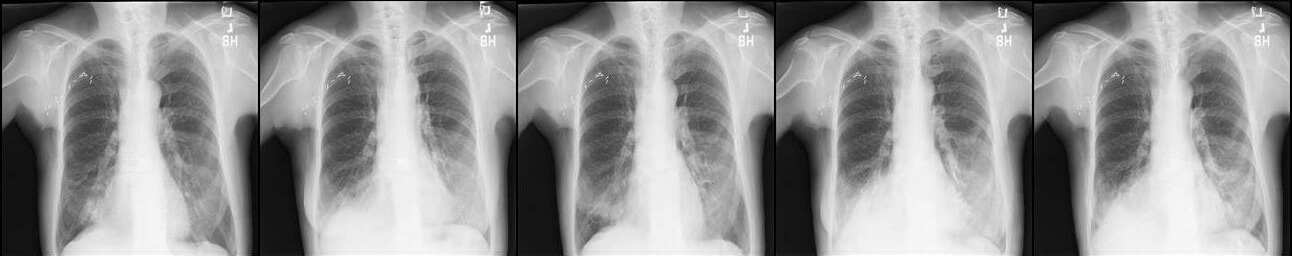}
    \includegraphics[height=\xrayimgheight]{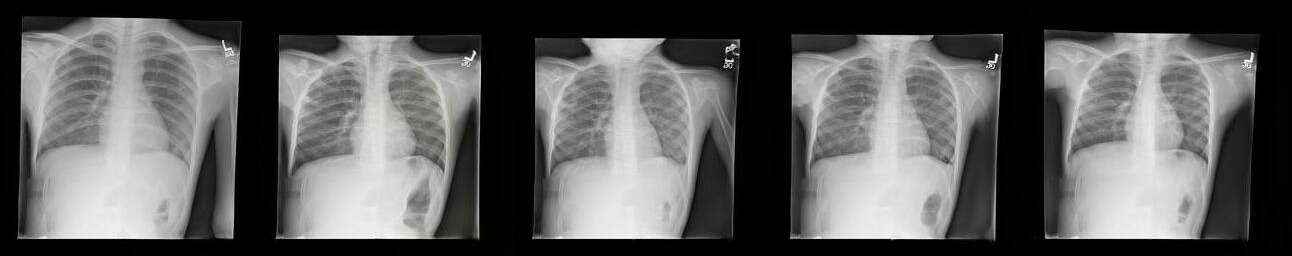}
    \includegraphics[height=\xrayimgheight]{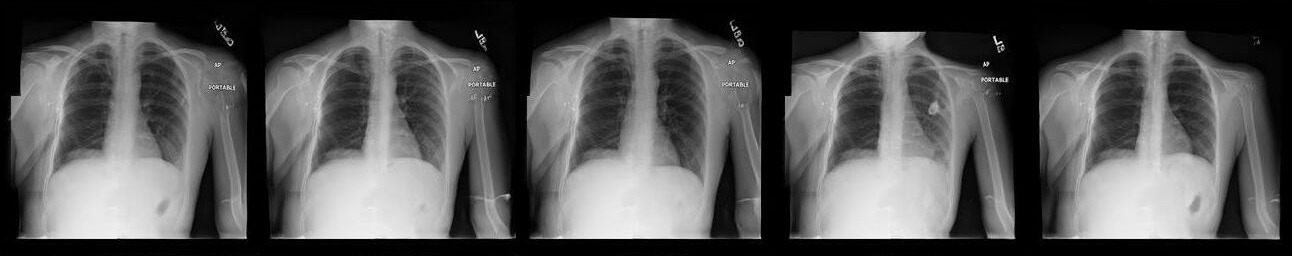}
    \includegraphics[height=\xrayimgheight]{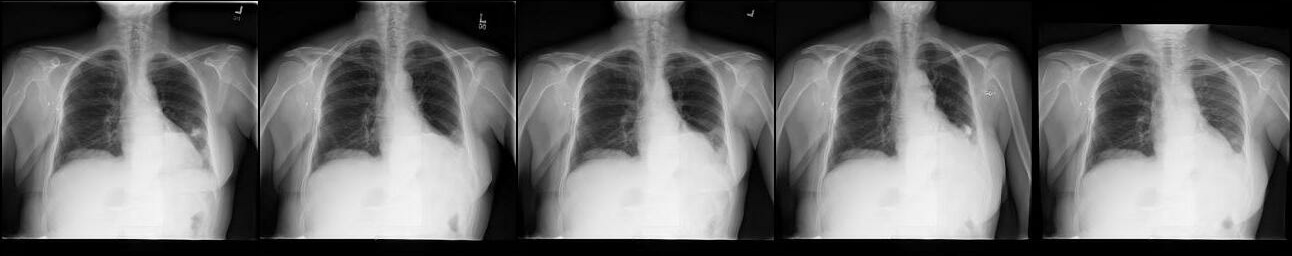}
    \caption{CoModGAN}
  \end{subfigure}
  %%%%%%%%%%%%%%%%%%%%%%
  \caption{Sampled completions given three different masks for each of two
    ground truth x-ray images. For the first ground truth image (top three
    rows), there is little obvious difference between the completions produced
    by IPA and by CoModGAN. However, for each of the masks applied to the second
    ground truth image (bottom three rows), CoModGAN produces a posterior with
    minimal diversity, and which does not appear to include the ground truth. We
    inspected completion panels for CoModGAN on 20 different ground truth
    images, and found that this type of mode collapse occurred in 5 of them.
    IPA, in contrast, produces reliably diverse image completions for any
    $\partI{}$ or ground truth image inspected. We do not show samples with a
    reduced temperature, as we did not use them for BOED on the basis that this
    could significantly reduce coverage of the posterior. }
    \label{fig:xray-samples}
  \end{figure*}

\end{document}